\documentclass[11pt]{article}
\pdfoutput=1
\usepackage[final]{acl}
\usepackage{siunitx} 
\usepackage{subcaption}
\usepackage{color,soul}
\usepackage{times}
\usepackage{latexsym}
\usepackage[T1]{fontenc}
\usepackage[utf8]{inputenc}
\usepackage{microtype}
\usepackage{inconsolata} 
\usepackage{csquotes}
\usepackage{enumitem}
\usepackage{pifont}

\usepackage{amsmath}
\usepackage{amssymb}  
\usepackage{amsthm}
\usepackage{mathtools}

\usepackage[ruled,vlined,linesnumbered]{algorithm2e}
\SetKwComment{tcp}{\small\textcolor{gray}{$\triangleright$\,}}{}
\DontPrintSemicolon

\usepackage{booktabs}
\usepackage{multirow}
\usepackage{multicol}
\usepackage{tabularx}
\usepackage{tabulary}
\usepackage{adjustbox}
\usepackage{siunitx}
\usepackage{makecell}

\usepackage{graphicx}
\usepackage{bm}
\usepackage{xcolor}
\usepackage{xcolor,colortbl}

\definecolor{darkgreen}{rgb}{0.0, 0.5, 0.0}
\definecolor{darkred}{rgb}{0.6, 0.0, 0.0}

\usepackage[most]{tcolorbox} 
\usepackage{cleveref}
\usepackage{makecell}

\newtcbtheorem[auto counter, number within=section, 
crefname={example}{Example},
Crefname={Example}{Example}]
{exmp}{Exam\smash{p}le} 
{colback=pale_red!5, colframe=DarkPurple, left=.02in, right=.02in,bottom=.02in, top=.02in}{exmp}  

\definecolor{navy}{rgb}{0.0, 0.0, 0.5}

\newtcbtheorem[auto counter, 
crefname={prompt}{Prompt},
Crefname={Prompt}{Prompt}]
{prompt}{Prom\smash{p}t} 
{colback=blue!0, colframe=navy, left=.02in, right=.02in,bottom=.02in, top=.02in}{prompt}  

\usepackage{listings}
\definecolor{Violet}{RGB}{148,0,211}    
\definecolor{DarkPurple}{RGB}{75,0,130} 
\definecolor{lightergray}{RGB}{230,230,230}
\definecolor{DarkRed}{RGB}{130,25,0}
\definecolor{PurpleRed}{RGB}{204,0,102}
\definecolor{DarkGreen}{RGB}{30,130,30}
\definecolor{DarkBlue}{RGB}{0,0,250}
\definecolor{DarkYellow}{RGB}{255,128,0}
\definecolor{light-gray}{gray}{0.95}
\definecolor{lightgreen}{RGB}{231,255,219}
\definecolor{lightred}{RGB}{252,231,234}
\definecolor{lightyellow}{RGB}{250,253,191}
\definecolor{lightpurple}{RGB}{229,204,255}
\definecolor{lightblue}{RGB}{229,246,254}
\definecolor{value-modification}{RGB}{250, 217, 86}
\definecolor{digit-expansion}{RGB}{216, 194, 104}
\definecolor{integer-decimal-fraction}{RGB}{240, 133, 51}
\definecolor{semantic-paraphrasing}{RGB}{85, 157, 63}
\definecolor{complexity-increasing}{RGB}{58, 120, 175}
\definecolor{question-transformation}{RGB}{174, 205, 225}
\definecolor{interference-injection}{RGB}{255,204,229}
\definecolor{remove-constrain}{RGB}{204,204,255}

\definecolor{pale_green}{rgb}{0.55,0.75,0.60}
\definecolor{pale_red}{rgb}{0.90,0.61,0.58}
\definecolor{pale_yellow}{rgb}{0.95,0.92,0.72}

\tcbset{
  aibox/.style={
    width=\textwidth,
    top=0pt, bottom=0pt, left=5pt, right=5pt,
    colback=white,
    colframe=black,
    colbacktitle=black,
    enhanced,
    center,
    attach boxed title to top left={yshift=-0.1in,xshift=0.15in},
    boxed title style={boxrule=0pt,colframe=white,},
  }
}
\newtcolorbox{AIbox}[2][]{aibox,title=#2,#1}
\addtocontents{toc}{\protect\setcounter{tocdepth}{-1}} 

\usepackage{amsmath,amssymb,bbm}   

\usepackage{graphicx}    
\usepackage{booktabs}    

\usepackage{tcolorbox}
\usepackage{xcolor}

\definecolor{defaultcolor}{HTML}{DAE8FC} 
\definecolor{deterministiccolor}{HTML}{D5E8D4} 
\definecolor{exploratorycolor}{HTML}{FFF2CC} 
\definecolor{detexploratorycolor}{HTML}{FADBD8} 
\definecolor{darkblue}{RGB}{25,25,112} 

\title{\textsc{Debate, Train, Evolve:} Self-Evolution of Language Model Reasoning}

\author{
 \textbf{Gaurav Srivastava\textsuperscript{$\heartsuit ^*$}},
 \textbf{Zhenyu Bi\textsuperscript{$\heartsuit$}},
 \textbf{Meng Lu\textsuperscript{$\heartsuit$}},
 \textbf{Xuan Wang\textsuperscript{$\heartsuit$\thanks{Corresponding Author}}}
\\
 \textsuperscript{$\heartsuit$}Department of Computer Science, Virginia Tech, USA
\\
 \small{
        \texttt{(\href{gks@vt.edu}{gks}, \href{zhenyub@vt.edu}{zhenyub}, \href{menglu@vt.edu}{menglu}, \href{xuanw@vt.edu}{xuanw})@vt.edu}
 }
 \\
 \small{
   \textbf{Website: }\href{https://ctrl-gaurav.github.io/debate-train-evolve.github.io/}{\textcolor{darkblue}{\texttt{debate-train-evolve.github.io/}}}
}
}

\begin{document}

\maketitle

\begin{abstract}
Large language models (LLMs) have improved significantly in their reasoning through extensive training on massive datasets. However, relying solely on additional data for improvement is becoming increasingly impractical, highlighting the need for models to autonomously enhance their reasoning without external supervision. In this paper, we propose \textsc{Debate, Train, Evolve (DTE)}, a novel ground truth-free training framework that uses multi-agent debate traces to evolve a single language model. We also introduce a new prompting strategy \textsc{Reflect-Critique-Refine}, to improve debate quality by explicitly instructing agents to critique and refine their reasoning. Extensive evaluations on \textbf{seven} reasoning benchmarks with \textbf{six} open-weight models show that our DTE framework achieve substantial improvements, with an average accuracy gain of \textbf{8.92\%} on the GSM-PLUS dataset. Furthermore, we observe strong cross-domain generalization, with an average accuracy gain of \textbf{5.8\%} on all other benchmarks, suggesting that our method captures general reasoning capabilities. \textit{Our framework code and trained models are publicly available at} \href{https://github.com/ctrl-gaurav/Debate-Train-Evolve}{\textit{https://github.com/ctrl-gaurav/Debate-Train-Evolve}}. 
\footnote{%
\textbf{GitHub:} \href{https://github.com/ctrl-gaurav/Debate-Train-Evolve}{github.com/ctrl-gaurav/Debate-Train-Evolve}

\hspace{0.8em}\textbf{DTE Website:} \href{https://ctrl-gaurav.github.io/debate-train-evolve.github.io/}{https://ctrl-gaurav.github.io/debate-train-evolve.github.io/}
}
\end{abstract}

\section{Introduction}

Over the past few years, the advancements in large language models (LLMs) have largely depended on training over massive datasets \cite{abdin2024phi, abdin2025phi}. However, eventually, we will approach a saturation point where feeding more data into these models may not further improve their reasoning capabilities \cite{costello2025think}. This motivates a new research question: \emph{How can language models continue to improve without relying on additional external supervision?} 

Recent approaches attempt to overcome the data bottleneck by enabling models to generate and learn from synthetic data, which is generated by automatically expanding a small set of seed tasks into large synthetic instruction datasets \cite{wang2022self, zeng2024automaticinstructionevolvinglarge}. Other methods \cite{madaan2023self, Jiang2023SelfEvolveAC, Gou2023CRITICLL, zelikman2024star, costello2025think} refine model-generated outputs through iterative self-feedback or preference optimization. Despite their effectiveness, these self-evolution strategies predominantly rely on judgments from a single model or a teacher-student configuration, often leading to confirmation bias and insufficient reasoning diversity. 

To address these limitations, one promising direction emerged is multi-agent debate (MAD) \cite{du2023improving}. It involves multiple models independently generating and critically analyzing each other's answers, helping to reveal subtle reasoning errors often overlooked by individual models \cite{liang2023encouraging,Wang2024RethinkingTB}. Although MAD shows improved reasoning accuracy, current works predominantly use MAD as an inference-time technique \cite{smit2023should}, requiring multiple models to be run simultaneously for each query. This substantially increases computational overhead and latency \cite{subramaniam2025multiagent}, making MAD impractical for large-scale deployments. This motivates our research question: \emph{Can we \textbf{evolve} a single model reasoning by fine-tuning on these debate traces?} 

\begin{figure*}[ht]
    \centering
    \includegraphics[width=1\linewidth]{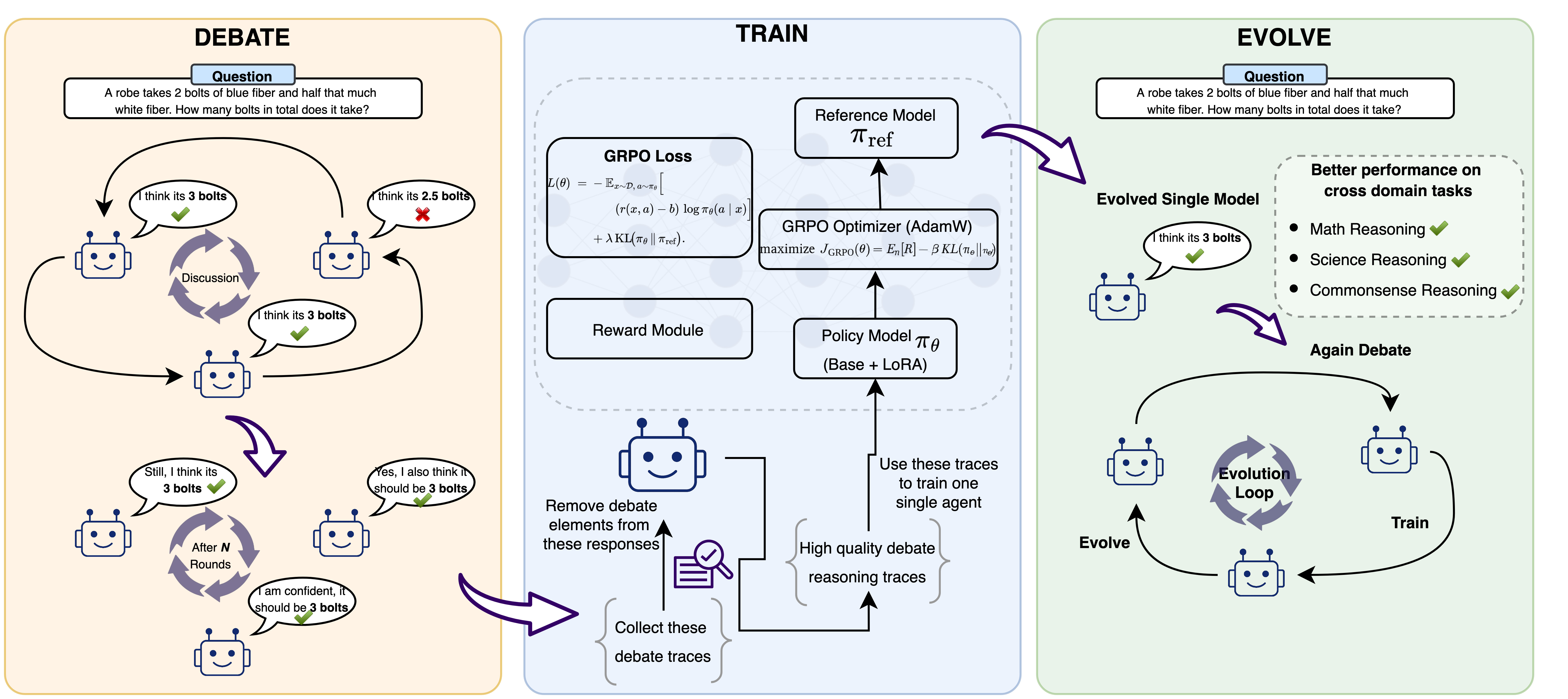}
            \caption{Overview of the proposed \textbf{\textsc{Debate–Train–Evolve}} framework. \textit{Left}-\textbf{Debate}: Several agents debate until they converge on a consensus \textcolor{darkgreen}{(green \ding{51})} or expose a wrong path \textcolor{red}{(red \ding{55})}. \textit{Centre}-\textbf{Train}: we remove pure debate elements, keep the high-quality reasoning traces and consensus answer, and use them to fine-tune a single policy with GRPO. \textit{Right}-\textbf{Evolve}: the evolved agent replaces its earlier self, so future inference require just one forward pass yet they outperform the committee on maths, science, and commonsense benchmarks.}
    \label{fig:main}
\end{figure*}

Building upon this intuition, we propose \textsc{Debate, Train, Evolve (DTE)}, a novel framework that combines the strengths of MAD with efficient single-model inference. Specifically, we introduce a ground-truth-free training approach in which a model learns from its own debate traces generated during MAD, thereby evolving autonomously over iterative training cycles. Our framework addresses key challenges of existing methods by extracting high-quality reasoning insights from diverse multi-agent interactions, thus avoiding single-model biases and computational inefficiencies.

\textbf{First}, we conduct a large-scale empirical analysis of MAD using open-source models, where we identify limitations of the original MAD prompting approach, particularly in smaller models \cite{du2023improving}. To address this, we propose a \textsc{Reflect-Critique-Refine (RCR)} prompting strategy, which explicitly forces agents to identify, critique, and correct reasoning errors in both their own and peers' answers. \textbf{Second,} using this prompting strategy, we build our \textsc{DTE} framework (Figure \ref{fig:main}). \textbf{Finally}, we find that models with $<3B$ parameters suffer accuracy loss \cite{srivastava2025towards, srivastava2025llmthinkbenchbasicmathreasoning, srivastava2025beyondbenchbenchmarkfreeevaluationreasoning} after second evolution round; our controlled study shows that the problem correlates with large temperature-induced variance and high KL divergence from the base policy. Lowering the sampling temperature from 0.7 to 0.3 cuts the KL drift by $1/3rd$ and recovers up to 76\% of the lost performance, preventing catastrophic forgetting in smaller models without extra supervision.

Our experiments show significant gains in reasoning performance across multiple datasets. Specifically, our evolved models show an average accuracy improvement of \textbf{8.92\%} on GSM-PLUS dataset compared to their original versions. Moreover, our framework achieves notable cross-domain generalization, enhancing model performance across datasets not seen during training. These results confirm that our method successfully distills multi-agent debate's insights into efficient single-model inference, bridging the gap between computational efficiency and improved reasoning.

\section{Related Work}
\label{sec:related_work}
\paragraph{Multi-Agent Debate Approaches} \citet{du2023improving} first showed that letting several large models debate improves accuracy on maths, strategy, and factual QA without any new parameters. Later, \citet{liang2023encouraging} highlighted the risk of \emph{degeneration‑of‑thought}: a single agent quickly converges on one path, whereas a two‑debater plus judge setup maintains diversity and outperforms GPT‑4 on tricky arithmetic. \textsc{RECONCILE} \citep{chen2023reconcile} mixes agents from different model families, reaches consensus through confidence‑weighted votes, and adds up to eleven points on seven reasoning benchmarks. \citet{smit2023should} shows that MAD beats sampling ensembles only after careful tuning. Finally, works like PREDICT~\citep{park2024predict} apply multi-agent debate to tasks beyond QA, such as hate-speech classification, where agents reason under different guidelines. Recent advances further incorporate explicit reinforcement learning into the debate process. For example, the ACC-Collab framework \cite{estornell2024acc} utilized an actor-critic approach to explicitly optimize agent collaboration, yielding superior performance on reasoning tasks.

\paragraph{Self-Evolution in Language Models} \textsc{SELF‑INSTRUCT} \citep{wang2022self} prompts GPT‑3 to write 52000 novel instructions plus answers and then fine‑tunes on its own output, reducing the gap to InstructGPT by thirty‑three points on Super‑Natural‑Instructions without extra human labels. \textsc{STaR} \citep{zelikman2024star} augments a few chain‑of‑thought exemplars by letting the model explain wrong answers in reverse, doubling CommonsenseQA accuracy for a 350M model. \textsc{SELF‑REFINE} \citep{madaan2023self} and the broader \textsc{SELF} framework \citep{lu2023self} turn one model into writer, critic and re‑writer, looping feedback at inference or during fine‑tuning to improve on GSM8K by around seven points. Instruction‑tuning variants refine the idea: \textsc{Self‑Refine Instruction‑Tuning} \citep{ranaldi2024self} pairs Llama‑2 and Mistral students with large teacher rationales and then lets each student prefer its own better reasoning, closing the size gap on commonsense and math tasks. More recently, \textsc{Think, Prune, Train, Improve} \citep{costello2025think} shows that careful filtering of self‑generated traces can raise Gemma‑2B to 58\% on GSM8K and push Llama‑3‑70B beyond GPT‑4o. These studies confirm that single‑agent loops, with or without ground truth, can expand a model’s ability. \smallskip

Despite these works, two things remain unexplored: \textbf{\textit{1)}} Fully autonomous, ground-truth-free self-evolution; \textbf{\textit{2)}} Integration of MAD into model evolution. Our work addresses this by the \textsc{Debate, Train, Evolve} framework, which combines MAD with self-supervised reinforcement learning (GRPO) to enable models to autonomously evolve their reasoning capabilities.

\section{\textsc{Debate, Train, Evolve} Framework}
\label{sec:dte_framework}

In this section, we first analyze limitations of existing multi-agent debate approaches (§\ref{sec:mad_analysis}), introduce our improved prompting strategy (§\ref{sec:rcr_prompt}), and then detail the mathematical framework for training models using debate-derived rewards (§\ref{sec:grpo_training}). Our DTE framework uses multi-agent debate to generate high-quality reasoning traces, then distills these traces into a single model through group-relative policy optimization \cite{shao2024deepseekmathpushinglimitsmathematical}.

\subsection{Preliminary Analysis of Multi-Agent Debate}
\label{sec:mad_analysis}

Let $\mathcal{A} = \{a_1, \ldots, a_N\}$ denote a set of $N$ language model agents, and let $q$ represent an input query. In the standard multi-agent debate framework, each agent $a_i$ independently generates an initial response $(y_i^{(0)}, r_i^{(0)})$ consisting of an answer $y_i^{(0)}$ and rationale $r_i^{(0)}$. Agents then engage in $T$ rounds of debate, where in round $t$, each agent observes peer responses $\{(y_j^{(t-1)}, r_j^{(t-1)})\}_{j \neq i}$ and produces an updated response $(y_i^{(t)}, r_i^{(t)})$.

Our empirical analysis of this standard approach revealed two critical failure modes. First, we observed high rates of \textbf{\emph{sycophancy}}, where agents abandon correct answers in favor of incorrect but confidently-stated peer solutions. Second, we identified a \textbf{\emph{verbosity bias}} where agents preferentially adopt longer rationales regardless of logical validity \cite{saito2023verbosity}. These effects resulted in degraded debate quality (substantial fraction of [correct $\rightarrow$ incorrect] transitions during debate), particularly for smaller models where sycophancy rates exceeded 28\% on average.

\subsection{\textsc{Reflect-Critique-Refine} Prompting Strategy}
\label{sec:rcr_prompt}

To address these limitations, we introduce the \textsc{RCR} prompting strategy. Unlike standard debate prompts that simply request answer revision \cite{madaan2023self,Gou2023CRITICLL,Peng2023CheckYF}, RCR structures agent responses through three explicit phases: \textbf{\textit{1) Reflect}}: Each agent $a_i$ must identify potential errors in its current reasoning $r_i^{(t-1)}$ by generating a self-critique $c_i^{\text{self}}$. \textbf{\textit{2) Critique}}: The agent then evaluates exactly two peer rationales, producing critiques $\{c_i^{j}\}_{j \in \mathcal{P}_i}$ where $|\mathcal{P}_i| = 2$ and $\mathcal{P}_i \subset \mathcal{A} \setminus \{a_i\}$. \textbf{\textit{3) Refine}}: Finally, the agent updates its response to $(y_i^{(t)}, r_i^{(t)})$ subject to the constraint that if $y_i^{(t)} \neq y_i^{(t-1)}$, then $r_i^{(t)}$ must contain at least one novel reasoning step not present in $\bigcup_{j, s<t} r_j^{(s)}$.\smallskip

Phrases like \textit{“identify any errors”} reliably trigger negative tokens \textbf{(“error”, “mistake”, “step X is wrong”)} which LLMs have learned during supervised finetuning. By specifying valid next moves (defend/correct/adopt), we implicitly shape the log‑probability mass toward useful trajectories, shrinking the space of rambling answers. The single‑step explanation requirement forces agents to think before copying and reduces sycophancy by requiring agents to justify answer changes with novel reasoning, while the fixed critique quota prevents unbounded verbosity. Algorithm \ref{alg:mad} presents the complete debate protocol, where the debate terminates when either consensus is reached (all $y_i^{(t)}$ identical) or after $T$ rounds, with the final answer determined by majority vote.

\begin{algorithm}[t]
\small
\caption{Multi-Agent Debate with \textsc{RCR} Prompting}
\label{alg:mad}
\SetAlgoNoLine
\KwIn{query $q$, agents $\mathcal{A} = \{a_1, \ldots, a_N\}$, max rounds $T$}
\KwOut{consensus answer $y^*$ and reasoning traces $\mathcal{R}$}
\textbf{Round 0:} Each $a_i \in \mathcal{A}$ generates $(y_i^{(0)}, r_i^{(0)}) \sim \pi_{a_i}(\cdot | q)$\;
\If{all $y_i^{(0)}$ are identical}{\Return $(y_i^{(0)}, \{r_i^{(0)}\}_{i=1}^N)$}
\For{$t = 1$ \KwTo $T$}{
  \ForEach{agent $a_i \in \mathcal{A}$}{
    Receive peer responses: $\mathcal{P}_i^{(t-1)} = \{(y_j^{(t-1)}, r_j^{(t-1)})\}_{j \neq i}$\;
    \textbf{Reflect:} Generate self-critique $c_i^{\text{self}}$ identifying errors in $r_i^{(t-1)}$\;
    \textbf{Critique:} Select two peers and generate critiques $\{c_i^j\}_{j \in S_i}$ where $|S_i| = 2$\;
    \textbf{Refine:} Update response $(y_i^{(t)}, r_i^{(t)})$ with novel reasoning if $y_i^{(t)} \neq y_i^{(t-1)}$\;
  }
  \If{all $y_i^{(t)}$ are identical}{\Return $(y_i^{(t)}, \bigcup_{i,s \leq t} r_i^{(s)})$}
}
\Return $(\text{majority\_vote}(\{y_i^{(T)}\}), \bigcup_{i,t} r_i^{(t)})$
\end{algorithm}

\subsection{Training via Group Relative Policy Optimization}
\label{sec:grpo_training}

We now formalize how debate traces are used to train a single language model. Let $\pi_\theta$ denote a language model policy parameterized by $\theta$, which models the conditional distribution over token sequences: $\pi_\theta(a|s) = \prod_{t=1}^{|a|} \pi_\theta(a_t | s, a_{<t})$, where $s$ is the input state (query) and $a = (a_1, \ldots, a_{|a|})$ is the generated token sequence.

\paragraph{Debate Trace Extraction and Reward Design} Given a query $q$, we run multi-agent debate using Algorithm \ref{alg:mad} to obtain a consensus answer $y^*$ and a set of reasoning traces $\mathcal{R} = \{r_i^{(t)}\}_{i,t}$. From these traces, we extract a consolidated rationale $R$ by identifying reasoning steps that either (i) appear in multiple agents' responses or (ii) introduce novel symbolic manipulations. This yields a training instance $(q, y^*, R)$. For each generated response $y$ to query $q$, we define a shaped reward function:

\vspace{-1.5em}

\begin{align*}
r(q, y) =\;& 
    w_{\text{ans}} \cdot \mathbb{1}[y = y^*] 
    + w_{\text{fmt}} \cdot f_{\text{format}}(y) \nonumber \\
    &+ w_{\text{len}} \cdot \exp(-|y|/\tau)
\end{align*}

where $\mathbb{1}[y = y^*]$ indicates answer correctness (verified via exact string match after normalization), $f_{\text{format}}$ checks adherence to the XML template structure, $|y|$ denotes token length, and $(w_{\text{ans}}, w_{\text{fmt}}, w_{\text{len}}) = (2.0, 0.5, 0.5)$ with $\tau = 120$.

\paragraph{Group Relative Advantage Estimation} For training, we use Group Relative Policy Optimization (GRPO), which eliminates the need for a separate value function by estimating advantages through group-wise comparisons. For each query $q$ in our training batch, we sample $G$ responses $\{o_1, \ldots, o_G\}$ from the current policy $\pi_{\theta_{\text{old}}}$. Each response $o_i$ receives a scalar reward $r_i = r(q, o_i)$.

Instead of learning a value function $V(s)$ to estimate expected returns, GRPO computes advantages using the group statistics. The advantage for response $o_i$ at token position $t$ is:
$$\hat{A}_{i,t} = \frac{r_i - \bar{r}}{\sigma_r + \epsilon}$$
where $\bar{r} = \frac{1}{G}\sum_{j=1}^{G} r_j$ is the mean reward, $\sigma_r = \sqrt{\frac{1}{G}\sum_{j=1}^{G}(r_j - \bar{r})^2}$ is the standard deviation, and $\epsilon = 10^{-8}$ prevents division by zero.

This formulation provides several key benefits. \textbf{\textit{First,}} responses with above-average rewards receive positive advantages, encouraging the model to increase their likelihood. \textbf{\textit{Second,}} normalization by standard deviation ensures that advantages remain stable across different reward scales. \textbf{\textit{Third,}} using group statistics rather than a learned baseline reduces memory requirements by eliminating the value network.

\paragraph{Policy Optimization Objective} Given the group-relative advantages, we optimize the policy using a clipped surrogate objective with KL regularization. The GRPO loss for a single query is:

\vspace{-0.5em}

{\small
\[
\mathcal{L}_{\text{GRPO}}(\theta) =
\frac{1}{G}\sum_{i=1}^{G} \frac{1}{|o_i|}
\sum_{t=1}^{|o_i|}
\left[ \ell_{\text{clip}}(i,t) - \beta \cdot D_{\text{KL}}^{(i,t)} \right]
\]
}

where the clipped policy gradient loss is:

\vspace{-1.5em}

\begin{align*}
\ell_{\text{clip}}(i,t) = -\min\Big( 
    & \rho_{i,t} \cdot \hat{A}_{i,t}, \\
    & \text{clip}(\rho_{i,t}, 1-\epsilon, 1+\epsilon) \cdot \hat{A}_{i,t} 
\Big)
\end{align*}

Here, $\rho_{i,t} = \frac{\pi_\theta(a_{i,t} | q, o_{i,<t})}{\pi_{\theta_{\text{old}}}(a_{i,t} | q, o_{i,<t})}$ is the importance ratio between the new and old policies, and $\epsilon = 0.2$ is the clipping threshold. The clipping mechanism prevents destructively large policy updates: when $\rho_{i,t}$ exceeds $1+\epsilon$ or falls below $1-\epsilon$, the gradient contribution is capped.

The KL divergence term $D_{\text{KL}}^{(i,t)}$ regularizes the policy to prevent excessive deviation from a reference model $\pi_{\text{ref}}$ (typically the initial supervised fine-tuned model):
$$D_{\text{KL}}^{(i,t)} = \log\frac{\pi_\theta(a_{i,t} | q, o_{i,<t})}{\pi_{\text{ref}}(a_{i,t} | q, o_{i,<t})}$$

with regularization strength $\beta = 0.02$. This KL penalty serves a different purpose than the clipping: while clipping prevents large single-step updates, the KL term anchors the policy to maintain linguistic coherence and prevent catastrophic forgetting.

\paragraph{Gradient Estimation and Optimization} Gradient of $\mathcal{L}_{\text{GRPO}}$ with respect to $\theta$ is estimated using the REINFORCE algorithm. For each token $a_{i,t}$ in response $o_i$, the gradient contribution is:

\vspace{-1.5em}

{\small
\begin{equation*}
\nabla_\theta \mathcal{L}_{\text{GRPO}} 
= -\mathbb{E}_{o_i \sim \pi_{\theta_{\text{old}}}}
\left[ \sum_{t=1}^{|o_i|} \nabla_\theta 
\log \pi_\theta(a_{i,t}|q, o_{i,<t}) \cdot g(i,t) \right]
\end{equation*}
}

where $g(i,t)$ is the effective advantage after clipping and KL regularization. This expectation is approximated through Monte Carlo sampling using the $G$ generated responses. We optimize using AdamW with learning rate $\eta = 2 \times 10^{-5}$, weight decay $\lambda = 0.01$, and a 50-step linear warmup. To enhance training efficiency, we use LoRA (Low-Rank Adaptation) with rank $r = 128$ and dropout probability $p = 0.05$, applying adaptations to attention and MLP projection matrices while keeping embeddings and layer normalizations frozen.

\subsection{Evolution through Iterative Training}
\label{sec:evolution}

The complete \textsc{DTE} framework operates as an iterative process, formalized in Algorithm \ref{alg:dte}. Starting with a base policy $\pi_{\theta_0}$, we perform evolution rounds where each round $k$ consists of: \textbf{\textit{1) Debate Generation}}: Sample a batch of queries $\mathcal{Q}_k$ and generate debate traces using RCR-prompted multi-agent debate (Algorithm \ref{alg:mad}), producing dataset $\mathcal{D}_k = \{(q, y^*, R)\}$. \textbf{\textit{2) Policy Update}}: Fine-tune $\pi_{\theta_{k-1}}$ on $\mathcal{D}_k$ using GRPO to obtain $\pi_{\theta_k}$. \textbf{\textit{3) Agent Replacement}}: Replace the previous version in the debate ensemble with the evolved policy.

The process continues until validation performance plateaus or a maximum number of iterations is reached. For smaller models ($<3$B parameters), we implement temperature annealing from $T=0.7$ to $T=0.3$ across rounds to mitigate KL divergence growth and prevent catastrophic forgetting, as high-temperature sampling in later rounds can cause excessive policy drift.

This framework achieves autonomous reasoning improvement by combining the exploration benefits of multi-agent debate with the efficiency of single-model deployment, while GRPO's group-relative formulation provides stable training without requiring auxiliary value networks.

\begin{algorithm}[t]
\small
\caption{\textsc{Debate, Train, Evolve}}
\label{alg:dte}
\SetAlgoNoLine
\KwIn{base policy $\pi_{\theta_0}$, agent pool $\mathcal{A}_0 = \{\pi_{\theta_0}\} \cup \mathcal{B}$, query dataset $\mathcal{Q}$, max iterations $K$}
\KwOut{evolved policy $\pi_{\theta_K}$}
Initialize: $\theta \leftarrow \theta_0$\;
\For{$k = 1$ \KwTo $K$}{
  Sample batch $\mathcal{Q}_k \subset \mathcal{Q}$ of size $B$\;
  $\mathcal{D}_k \leftarrow \emptyset$\;
  \ForEach{query $q \in \mathcal{Q}_k$}{
    $(y^*, \mathcal{R}) \leftarrow$ Algorithm \ref{alg:mad} with agents $\mathcal{A}_{k-1}$ on query $q$\;
    $R \leftarrow$ ExtractRationale($\mathcal{R}$) \tcp{Extract consolidated reasoning}
    $\mathcal{D}_k \leftarrow \mathcal{D}_k \cup \{(q, y^*, R)\}$\;
  }
  \For{epoch $e = 1$ \KwTo $E$}{
    \ForEach{$(q, y^*, R) \in \mathcal{D}_k$}{
      Sample $G$ responses: $\{o_i\}_{i=1}^G \sim \pi_{\theta}(\cdot | q)$\;
      Compute rewards: $r_i = r(q, o_i)$ for each $o_i$\;
      Compute advantages: $\hat{A}_i = \frac{r_i - \bar{r}}{\sigma_r + \epsilon}$\;
      Update $\theta$ via gradient step on $\mathcal{L}_{\text{GRPO}}(\theta)$\;
    }
  }
  Update agent pool: $\mathcal{A}_k \leftarrow (\mathcal{A}_{k-1} \setminus \{\pi_{\theta_{k-1}}\}) \cup \{\pi_{\theta}\}$\;
  \If{validation improvement $< \delta$}{\textbf{break}}
}
\Return $\pi_{\theta}$
\end{algorithm}

\begin{table*}[t]
\centering
\resizebox{\textwidth}{!}{
\begin{tabular}{@{}l cc c cc c cc c cc c cc c @{}}
\toprule
\multirow{3}{*}{\textbf{Model}} & \multicolumn{3}{c}{\textbf{GSM8K}} & \multicolumn{3}{c}{\textbf{GSM-Plus}} & \multicolumn{3}{c}{\textbf{MATH}} & \multicolumn{3}{c}{\textbf{ARC-Challenge}} & \multicolumn{3}{c}{\textbf{GPQA Main}} \\
\cmidrule(lr){2-4} \cmidrule(lr){5-7} \cmidrule(lr){8-10} \cmidrule(lr){11-13} \cmidrule(lr){14-16}
 & Original & 3 Agent & Evolved Single & Original & 3 Agent & Evolved Single & Original & 3 Agent & Evolved Single & Original & 3 Agent & Evolved Single & Original & 3 Agent & Evolved Single \\
 & Model & MAD & Model (DTE) & Model & MAD & Model (DTE) & Model & MAD & Model (DTE) & Model & MAD & Model (DTE) & Model & MAD & Model (DTE) \\
\midrule
Qwen-2.5-1.5B & 62.77 & 72.33 & 73.09 (\textcolor{darkgreen}{+10.32 $\uparrow$}) & 42.00 & 53.33 & 55.92 (\textcolor{darkgreen}{+13.92 $\uparrow$}) & 45.08 & 50.68 & 52.20 (\textcolor{darkgreen}{+7.12 $\uparrow$}) & 69.21 & 68.52 & 68.36 (\textcolor{darkred}{-0.85 $\downarrow$}) & 19.42 & 18.75 & 20.10 (\textcolor{darkgreen}{+0.68 $\uparrow$}) \\
Qwen-2.5-3B   & 84.08 & 85.14 & 86.05 (\textcolor{darkgreen}{+1.97 $\uparrow$})  & 61.75 & 68.00 & 69.50 (\textcolor{darkgreen}{+7.75 $\uparrow$})  & 61.36 & 65.72 & 67.10 (\textcolor{darkgreen}{+5.74 $\uparrow$}) & 83.53 & 84.64 & 83.95 (\textcolor{darkred}{-0.42 $\downarrow$}) & 28.12 & 29.24 & 30.50 (\textcolor{darkgreen}{+2.38 $\uparrow$}) \\
Qwen-2.5-7B   & 90.67 & 91.21 & 88.32 (\textcolor{darkred}{-2.35 $\downarrow$})  & 68.62 & 74.17 & 74.71 (\textcolor{darkgreen}{+6.09 $\uparrow$})  & 73.08 & 75.58 & 77.20 (\textcolor{darkgreen}{+4.12 $\uparrow$}) & 87.22 & 91.64 & 90.89 (\textcolor{darkgreen}{+3.67 $\uparrow$}) & 32.81 & 33.71 & 35.20 (\textcolor{darkgreen}{+2.39 $\uparrow$}) \\
Qwen-2.5-14B  & 92.80 & 93.33 & 93.74 (\textcolor{darkgreen}{+0.94 $\uparrow$})  & 71.79 & 77.25 & 78.88 (\textcolor{darkgreen}{+7.09 $\uparrow$})  & 76.18 & 78.62 & 80.10 (\textcolor{darkgreen}{+3.92 $\uparrow$}) & 90.27 & 93.77 & 93.13 (\textcolor{darkgreen}{+2.86 $\uparrow$}) & 41.29 & 42.19 & 43.60 (\textcolor{darkgreen}{+2.31 $\uparrow$}) \\
Llama-3.2-3B      & 72.55 & 73.84 & 75.06 (\textcolor{darkgreen}{+2.51 $\uparrow$})  & 45.67 & 51.12 & 53.79 (\textcolor{darkgreen}{+8.12 $\uparrow$})  & 39.76 & 41.90 & 43.80 (\textcolor{darkgreen}{+4.04 $\uparrow$}) & 73.12 & 76.19 & 77.23 (\textcolor{darkgreen}{+4.11 $\uparrow$}) & 26.12 & 29.24 & 30.80 (\textcolor{darkgreen}{+4.68 $\uparrow$}) \\
Llama-3.1-8B      & 81.73 & 82.18 & 86.81 (\textcolor{darkgreen}{+5.08 $\uparrow$})  & 55.62 & 60.79 & 66.17 (\textcolor{darkgreen}{+10.55 $\uparrow$}) & 46.66 & 47.90 & 49.40 (\textcolor{darkgreen}{+2.74 $\uparrow$}) & 77.65 & 85.07 & 86.53 (\textcolor{darkgreen}{+8.88 $\uparrow$}) & 27.46 & 32.37 & 34.10 (\textcolor{darkgreen}{+6.64 $\uparrow$}) \\
\bottomrule
\end{tabular}
} 
\caption{\textbf{Performance of one \textsc{Debate–Train–Evolve} round.}  
For six open-weight models we report test accuracy on five reasoning benchmarks in three settings: the single \emph{base} model (``Original''), a \textbf{3-agent} debate using our \textsc{RCR} prompt (``MAD''), and the \emph{evolved single} student obtained after one DTE round. \textcolor{darkgreen}{\textbf{Green}} numbers denote the absolute gain of the evolved model over its Original Model, \textcolor{darkred}{\textbf{red}} numbers a decrease in performance.}  
\label{tab:dte_main}
\end{table*}

\section{Experiments}
\label{sec:exp}

\subsection{Experimental Setup}

\paragraph{Datasets.} We conduct experiments on \textbf{seven} public reasoning benchmarks: \textbf{\textit{1)} GSM8K} \cite{cobbe2021trainingverifierssolvemath}, \textbf{\textit{2)} GSM-Plus} \cite{li2024gsmpluscomprehensivebenchmarkevaluating} (adversarial math problems), \textbf{\textit{3)} MATH} \cite{hendrycks2021measuring} (competition-level mathematics), \textbf{\textit{4) ARC-Easy}}, \textbf{\textit{5) ARC-Challenge}} \citep{clark2018thinksolvedquestionanswering} (science reasoning), \textbf{\textit{6) GPQA Main}} \cite{rein2024gpqa} (graduate-level STEM questions), and \textbf{\textit{7) CommonsenseQA}} \citep{talmor-etal-2019-commonsenseqa}.

\paragraph{Baselines and models.}
We conduct of RCR prompting study on \textbf{ten} open-weight models, Qwen (0.5-32B), Llama-3/8B, Mistral-7B, Phi-mini, and \textbf{two} proprietary models, GPT-4o and GPT-4o-mini. We study our DTE framework with 6 models (Qwen 1.5B-14B, Llama-3B and Llama-8B). \textbf{Baselines are:} (i) the single \emph{original} model; (ii) \emph{vanilla MAD} with the original MAD prompt.

\paragraph{Parameter settings.} During debate we sample each agent once per query at temperature $T\!=\!1.0$ (exploratory) or $0.0$ (deterministic); mixed-teams use one exploratory and two deterministic agents. For GRPO training, we adopt LoRA fine-tuning (rank 128, dropout 0.05) on attention and MLP projections, freezing embeddings and layer norms. GRPO is optimized with AdamW (learning rate $\eta = 5 \times 10^{-6}$, weight decay $\lambda = 0.1$, and momentum coefficients $\beta_1 = 0.9$, $\beta_2 = 0.99$). We set the GRPO-specific hyperparameters as: clipping threshold $\epsilon = 0.2$, KL coefficient $\beta = 0.02$, and group size $G = 8$ responses per query. Each evolution epoch processes 8k debate traces ($\sim$2 M tokens) and runs on A100-80 GB GPUs for a 7B model; larger models scale near-linearly.

\paragraph{Evaluation metrics.} Task performance is \emph{exact match} for GSM-style datasets and \emph{accuracy} for MC-QA. For RCR evaluation, we also track \textbf{Sycophancy-Rate} and [incorrect $\rightarrow$ correct] instances.

\subsection{Main Results}
Our main results are organized into three main parts: \textbf{\textit{(1)}} First, we evaluate the effectiveness of \textsc{DTE} framework, \textbf{\textit{(2)}} Next, we test its generalization across different reasoning tasks, and \textbf{\textit{(3)}} Finally, we analyze the extent of model self-evolution through iterative rounds. 

\paragraph{\textsc{1) Overall DTE performance.}} \textbf{Evolved model using \textsc{DTE} shows an average gain of \textsc{8.92\% accuracy} on GSM-PLUS compared to its vanilla performance.} Table \ref{tab:dte_main} contrasts three settings: the single base model (``Original’’), a three-agent debate with our RCR prompt (``MAD’’), and the \emph{evolved single model} produced by one \textsc{Debate–Train–Evolve} pass. On \textbf{GSM-Plus}-the hard math dataset-DTE improves every model, with an average gain of \textbf{+2.38 points} over three-agent MAD. Qwen-1.5B shows the largest jump (+13.92 pts), confirming that \textbf{\textit{evolution is most helpful when the base model has head-room and the debate provides diverse traces.}} On \textbf{GSM8K} the average gain is smaller (\,+0.84 pts) because several models were already near their ceiling after debate. \textbf{ARC-Challenge} sees a mixed results: large models benefit (+3.67 pts for Qwen-7B, +8.88 pts for Llama-8B) while small models drift by $<1$ pt. Overall, DTE shows a mean improvement of \textbf{3.06 pts} over single model and \textbf{+1.09 pts} over MAD while restoring single-pass inference.

\begin{table*}[t]
\centering
\resizebox{\textwidth}{!}{
\begin{tabular}{@{}lcccccccc@{}}
\toprule
\multirow{3}{*}{\textbf{Model}} & \multicolumn{4}{c}{\textbf{Fine-tuned on GSM8K}} & \multicolumn{4}{c}{\textbf{Fine-tuned on GSM-Plus}} \\
\cmidrule(lr){2-5} \cmidrule(lr){6-9}
& GSM-Plus & ARC-Easy & ARC-Challenge & CommonsenseQA & GSM8K & ARC-Easy & ARC-Challenge & CommonsenseQA \\
& ($\Delta$) & ($\Delta$) & ($\Delta$) & ($\Delta$) & ($\Delta$) & ($\Delta$) & ($\Delta$) & ($\Delta$) \\
\midrule
Qwen-2.5-1.5B
& \textcolor{darkgreen}{+9.21 $\uparrow$}   
& \textcolor{darkred}{-1.60 $\downarrow$}    
& \textcolor{darkgreen}{+0.67 $\uparrow$}   
& \textcolor{darkred}{-2.23 $\downarrow$}    
& \textcolor{darkgreen}{+10.32 $\uparrow$}  
& \textcolor{darkred}{-1.52 $\downarrow$}    
& \textcolor{darkgreen}{+0.24 $\uparrow$}   
& \textcolor{darkred}{-2.31 $\downarrow$}    
\\
Qwen-2.5-3B
& \textcolor{darkgreen}{+3.79 $\uparrow$}   
& \textcolor{darkgreen}{+1.27 $\uparrow$}   
& \textcolor{darkgreen}{+0.83 $\uparrow$}  
& \textcolor{darkgreen}{+3.26 $\uparrow$}   
& \textcolor{darkgreen}{+1.36 $\uparrow$}   
& \textcolor{darkgreen}{+1.09 $\uparrow$}   
& \textcolor{darkgreen}{+0.60 $\uparrow$}   
& \textcolor{darkgreen}{+3.26 $\uparrow$}   
\\
Qwen-2.5-7B
& \textcolor{darkgreen}{+1.01 $\uparrow$}   
& \textcolor{darkgreen}{+1.73 $\uparrow$}   
& \textcolor{darkgreen}{+4.50 $\uparrow$}   
& \textcolor{darkgreen}{+3.40 $\uparrow$}   
& \textcolor{darkgreen}{+1.14 $\uparrow$}   
& \textcolor{darkgreen}{+1.69 $\uparrow$}   
& \textcolor{darkgreen}{+3.65 $\uparrow$}   
& \textcolor{darkgreen}{+3.32 $\uparrow$}   
\\
Qwen-2.5-14B
& \textcolor{darkgreen}{+1.67 $\uparrow$}   
& \textcolor{darkgreen}{+2.53 $\uparrow$}   
& \textcolor{darkgreen}{+3.42 $\uparrow$}   
& \textcolor{darkgreen}{+1.33 $\uparrow$}   
& \textcolor{darkgreen}{+0.53 $\uparrow$}   
& \textcolor{darkgreen}{+2.32 $\uparrow$}   
& \textcolor{darkgreen}{+4.01 $\uparrow$}   
& \textcolor{darkred}{-0.14 $\downarrow$}    
\\ 
Llama-3.2-3B
& \textcolor{darkgreen}{+6.71 $\uparrow$}   
& \textcolor{darkgreen}{+2.48 $\uparrow$}   
& \textcolor{darkred}{-1.11 $\downarrow$}    
& \textcolor{darkgreen}{+3.10 $\uparrow$}   
& \textcolor{darkgreen}{+3.80 $\uparrow$}   
& \textcolor{darkgreen}{+1.93 $\uparrow$}   
& \textcolor{darkred}{-3.92 $\downarrow$}    
& \textcolor{darkgreen}{+3.51 $\uparrow$}   
\\
Llama-3.1-8B
& \textcolor{darkgreen}{+8.13 $\uparrow$}   
& \textcolor{darkgreen}{+3.91 $\uparrow$}   
& \textcolor{darkgreen}{+6.74 $\uparrow$}   
& \textcolor{darkgreen}{+1.10 $\uparrow$}   
& \textcolor{darkgreen}{+5.15 $\uparrow$}   
& \textcolor{darkgreen}{+4.88 $\uparrow$}   
& \textcolor{darkgreen}{+7.84 $\uparrow$}   
& \textcolor{darkgreen}{+0.85 $\uparrow$}   
\\
\bottomrule
\end{tabular}
} 
\caption{\textbf{Cross-domain generalisation of evolved models.}  Each cell shows the change in test accuracy (\(\Delta\), in points) after one DTE pass, relative to the same model before evolution. The table is split by the dataset used for fine-tuning-GSM8K (left block) or GSM-Plus (right block)-and reports transfer to four unseen targets. \textcolor{darkgreen}{\textbf{Green}} numbers signal gains, \textcolor{darkred}{\textbf{red}} numbers losses.}
\label{tab:finetune_delta_performance}
\end{table*}

\vspace{-1em}
\paragraph{\textsc{2) Cross-domain generalization.}} \textbf{Our results suggests that DTE improves reasoning that travels beyond the source data, with larger models showing the most stable improvements.} Table \ref{tab:finetune_delta_performance} reports how well the evolved models generalize on other datasets. We test two scenarios: evolve using \textbf{\textit{(i)}} GSM8K; \textbf{\textit{(ii)}} GSM-Plus and test on four unseen datasets. When trained on GSM8K, every model gains on GSM-Plus (average \textbf{+5.8} pts) and on ARC-Challenge (+2.5 pts on average). ARC-Easy also sees small but consistent gains except for the 1.5B model, which drops 1.6 pts. CommonsenseQA improves for 5/6 models, indicating that the reward shaped from mathematical traces still helps improve on commonsense reasoning. Negative deltas are confined to the smallest model (Qwen-1.5B) and to a lesser degree Qwen-3B, suggesting that small models struggles to reconcile new skills with prior knowledge. In contrast, models $\geq7$B never lose more than 0.2 pts on any transfer task. Training on GSM-Plus and testing on GSM8K yields similar behaviour: large gains on the GSM8K (+3.7 pts on average) and moderate gains on others. \textbf{\textit{The symmetry suggests that DTE learns general reasoning heuristics (e.g. numeric decomposition, unit tracking) rather than memorising dataset-specific patterns.}}

\vspace{-1em}

\begin{figure}[t]
    \centering
    \includegraphics[width=\linewidth]{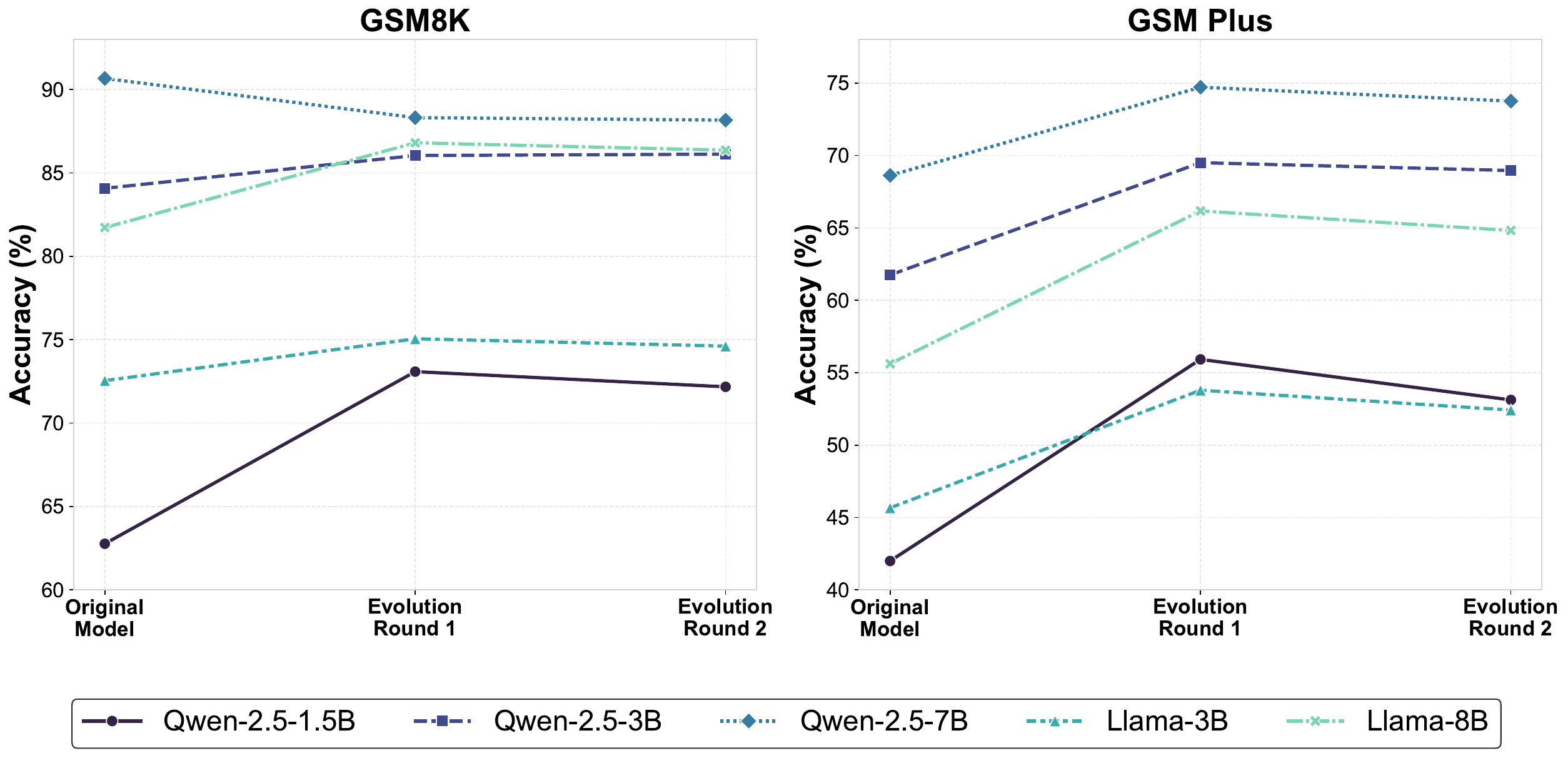}
    \caption{\textbf{Accuracy vs.\ evolution round.} }
    \label{fig:evo_curve}
\end{figure}

\paragraph{\textsc{3) How far can a model evolve?}} \textbf{Results show that one evolution round captures nearly all of the available gains.} Figure \ref{fig:evo_curve} reports accuracy over two evolution rounds for five models on GSM8K and GSM-Plus. Round 1 almost always helps: the smallest model (Qwen-1.5B) jumps from 42.0 $\rightarrow$ 55.9 on GSM-Plus and 62.8 $\rightarrow$ 73.1 on GSM8K, while Llama-8B gains 10.6 and 5.1 points on the same datasets. The only counter-example is Qwen-7B, which drops 2.4 points on GSM8K despite improving 6.1 on GSM-Plus; upon manual inspection we see that its Round-1 traces over-emphasise shortcut heuristics that hurt easier questions. \textbf{In Round 2, we observe  little improvement and sometimes the performance even drops.} Large models ($\geq7$ B) add at most +0.8 points, for Qwen-3B on GSM8K, and more often lose 0.4–1.4 points. The 1.5B model gives back 0.9 points on GSM8K and 2.8 on GSM-Plus, but still ends well above its starting point. Across all runs the mean forgetting $\text{Fgt}_2=\max_{t<2}(\text{Acc}_t-\text{Acc}_2)$ is 0.92 pts for models $\geq7$ B and 1.6 pts for smaller ones, confirming that smaller models suffers from catastrophic forgetting.

\begin{figure*}[ht]
    \centering
    \includegraphics[width=\linewidth]{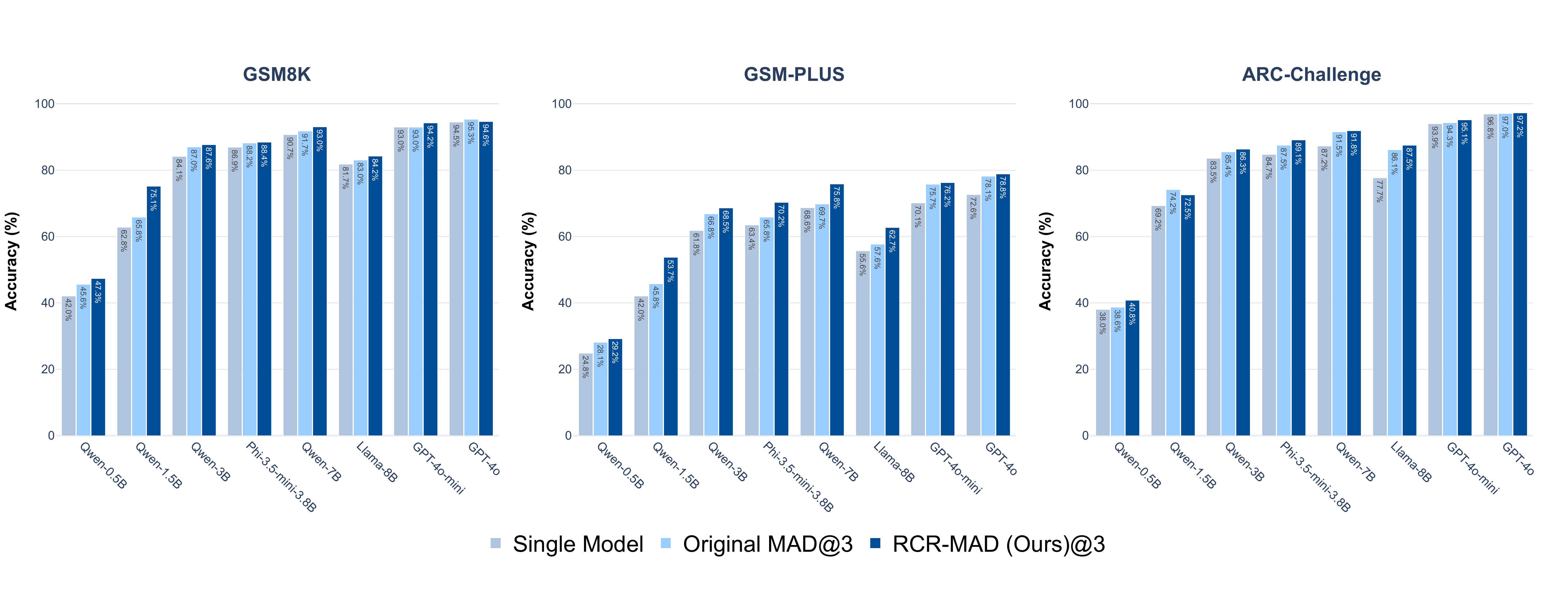}
    \caption{Results (\%) on: GSM8K, GSM-PLUS, and ARC-Challenge datasets. Performance is compared across three evaluation settings: single model inference, the Original Multi-Agent Debate (MAD@3) prompt, and our proposed RCR (RCR-MAD (Ours)@3) prompting.}
    \label{fig:mad_results}
\end{figure*}

\subsection{Ablation Studies}
\label{sec:ablation}

\paragraph{\textsc{1) Effectiveness of the RCR prompt in MAD.}} \textbf{RCR prompting substantially boost performance over original MAD prompting} \cite{du2023improving}. Figure~\ref{fig:mad_results} compares single-model inference, the original debate prompt (MAD@3), and our \textsc{Reflect–Critique–Refine} (RCR-MAD@3) prompt. Across eight diverse models the RCR prompting raises three-agent accuracy by an average of \textbf{+1.9 pts} on GSM8K, \textbf{+3.7 pts} on GSM-Plus, and \textbf{+0.7 pts} on ARC-Challenge. The gain scales with task difficulty: GSM-Plus, which contains harder adversarial questions, benefits the most (up to +7.9 pts for Qwen-1.5B and +6.1 pts for Qwen-7B). On ARC-Challenge improvements are smaller but still positive for 6/8 models. \textbf{RCR prompting also significantly reduces sycophancy.} It \emph{halves} the mean sycophancy rate (from 0.28 to 0.13 on GSM-Plus) and narrows the verbosity gap by 43 \%, indicating that agents now switch answers only when they can articulate a new reasoning step. \textbf{\textit{These observations confirm that RCR is a necessary pre-step for producing high-quality traces later utilized by the DTE training loop.}}

\paragraph{\textsc{2) How many agents are enough?}} 

\begin{figure}[t]
    \centering
    \includegraphics[width=\linewidth]{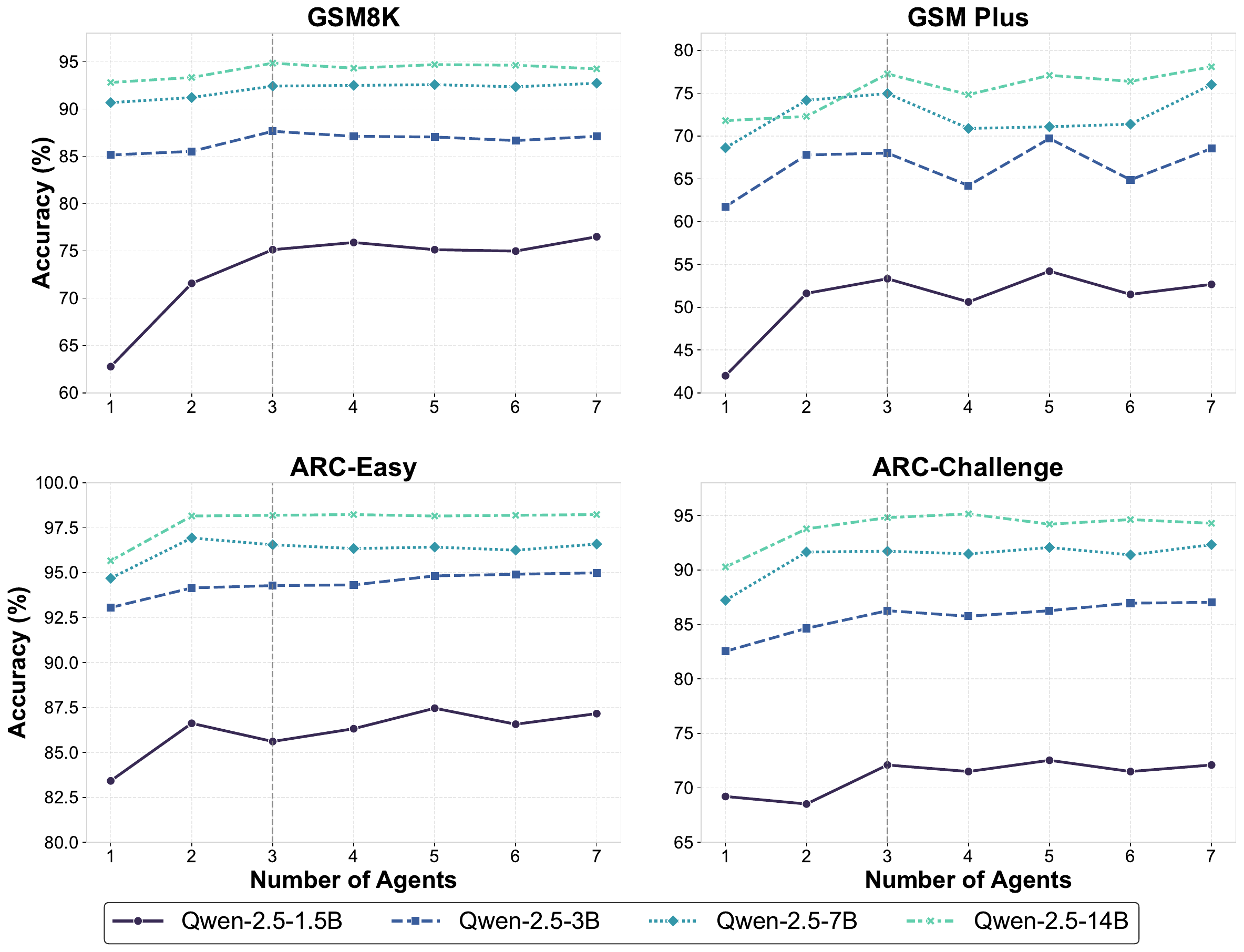}
    \caption{\textbf{Scaling up agents}  
    Accuracy of four Qwen model sizes as the number of agents grows from 1-7. }
    \label{fig:scaling_mad}
\end{figure}

\textbf{Results shows that \emph{three} agents MAD captures 85-95 \% of the maximum gains.} Figure \ref{fig:scaling_mad} sweeps the agents size from $1 - 7$ and reports trends on four benchmark. We observe three clear patterns here: \textbf{\textit{1)} Beyond 3-agent the curve plateaus and even oscillates}, suggesting the marginal information added by the 4th or 5th agent. \textbf{\textit{2)} Small models benefit most from extra agents.} Already strong single-agent (Qwen-14B) adds minimal improvement upon scaling up after three. \textbf{\textit{3)} Harder tasks need (slightly) more agents.} On GSM-Plus the optimum often shifts to four or five agents: Qwen-7B reaches its peak accuracy (76.0\%) at 7 agents, 1.04 pts above the three-agent setting. ARC-Easy, a much easier dataset, saturates at 2 agents for every model; extra debaters add noise rather than insight.

\paragraph{\textsc{3) Does agent diversity matter?}} We observe two consistent trends here: \textbf{First,} when the individual agents have comparable standalone accuracy, cross-family mixtures beat homogeneous agents team, supporting the idea that architectural diversity yields complementary reasoning paths. \textbf{Second,} when the pool mixes a strong and a weaker model, the debate result gravitates toward the stronger member-adding the weaker agent neither helps nor seriously harms, suggesting that diversity only helps when all agents can contribute novel insights. Complete results for every dataset and roster is available in Appendix B.

\paragraph{\textsc{4) Why GRPO over other fine-tuning methods?}} \textbf{GRPO consistently outperforms the alternatives,} indicating that its relative-advantage reward balances exploration and policy stability better than plain maximum-likelihood (SFT) or preference-only (DPO/PPO) updates. Table \ref{tab:rl_compare} compare three update rules under a fixed compute budget: (1) classical supervised fine-tuning on debate answers (SFT); (2) Direct Preference Optimisation using the majority vote as the preferred sample; (3) Group Relative Policy Optimisation (GRPO). GRPO delivers the largest accuracy jump on GSM-Plus for every model size. Both SFT and DPO give smaller gains and even slight regressions on the 3 B model, highlighting the risk of over-fitting when the reward ignores policy shift. We also observe that GRPO keeps $\text{KL}<0.24$ across sizes, whereas DPO averages 0.43. The relative-advantage term in GRPO therefore not only boosts reward but also constrains drift, reducing catastrophic forgetting.

\begin{table}[t]
\centering
\small
\begin{tabular}{@{}l S[table-format=2.2] S[table-format=2.2] S[table-format=2.2] S[table-format=2.2]@{}}
\toprule
\textbf{Model} & {\shortstack{\textbf{Original} \\ \textbf{(GSM-Plus)}}} & {\textbf{SFT}} & {\textbf{DPO}} & {\textbf{GRPO}} \\
\midrule
Qwen-2.5-1.5B & 42.00 & 47.31 & 51.34 & \textbf{55.92} \\
Qwen-2.5-3B   & 61.75 & 58.33 & 64.32 & \textbf{69.50} \\
Qwen-2.5-7B   & 68.62 & 67.89 & 69.88 & \textbf{74.71} \\
\bottomrule
\end{tabular}
\caption{Accuracy on GSM-Plus after \textbf{10K }training steps using three optimization objectives.}
\label{tab:rl_compare}
\end{table}

\paragraph{\textsc{5) Data selection strategy.}} We test three data sampling schemes on GSM-Plus: \emph{Random-2K} selects 2000 examples uniformly from the full pool (10552); \emph{Debate-Only} keeps only data points where agents entered at least one critique round (\(t\ge1\)); \emph{All-Traces} trains on the entire cleaned set. Table \ref{tab:data_selection} shows that accuracy rises monotonically with coverage: the full corpus beats Debate-Only by \textbf{4.43 pts} (avg) and Random-2K by \textbf{9.17 pts }(avg). The gap is largest for Qwen-1.5B, suggesting that smaller models benefit from easier “round-0’’ examples that Random-2K may miss and Debate-Only discards. We therefore use the full trace set in all other experiments.

\begin{table}[t]
\centering
\small
\begin{tabular}{@{}lccc@{}}
\toprule
\textbf{Model} & \textbf{Random-2K} & \textbf{Debate-Only} & \textbf{All-Traces} \\
\midrule
Qwen-1.5B & 44.82 & 51.61 & \textbf{55.92} \\
Qwen-3B   & 58.10 & 62.70 & \textbf{69.50} \\
Qwen-7B   & 69.71 & 72.53 & \textbf{74.71} \\
\bottomrule
\end{tabular}
\caption{\textbf{Effect of training-set size and composition.} GSM-Plus accuracy after one evolution round using three trace-selection schemes.}
\label{tab:data_selection}
\end{table}

\begin{figure}[t]
    \centering
    \includegraphics[width=0.8\linewidth]{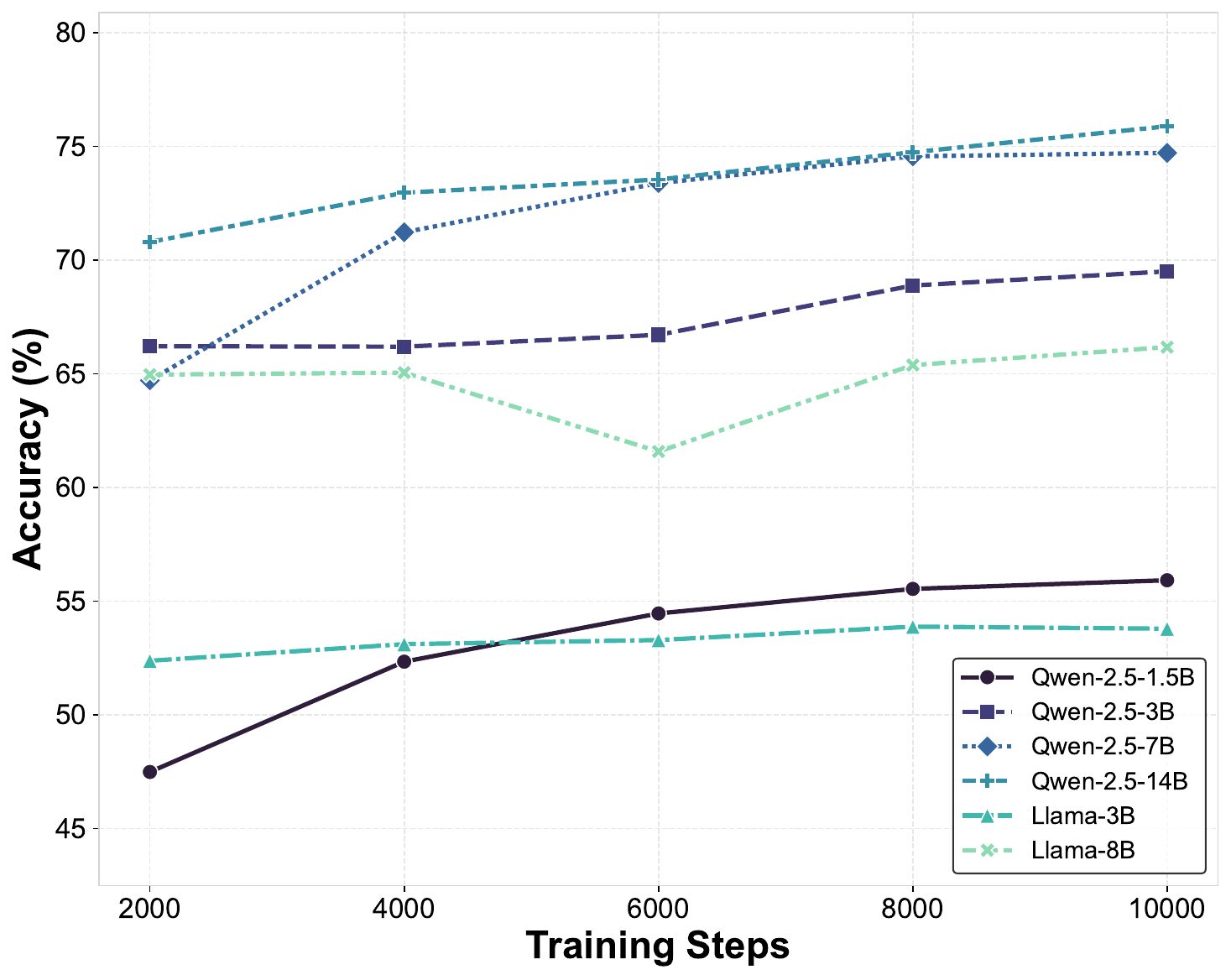}
    \caption{\textbf{Diminishing returns in GRPO updates after 8K steps.}  
    GSM-Plus accuracy for five models as a function of the number of training steps during GRPO.}

    \label{fig:steps}
\end{figure}

\paragraph{\textsc{6) How long do we train?}} Figure \ref{fig:steps} plots GSM-Plus accuracy as we grow the number of GRPO training steps from 2K to 10K. All models share the similiar trend: rapid gains up to about 8K steps followed by saturation. Small and mid-size models profit the most from the early updates-Qwen-1.5B climbs 8.0 pts between 2K and 6K samples-whereas larger models such as Qwen-14B rise more slowly but steady. Beyond 8K the curve flattens: the average improvement from 8K $→$ 10 k is only +0.32 pts while wall-clock time grows by 25\%.  

\begin{figure}[t]
    \centering
    \includegraphics[width=0.8\linewidth]{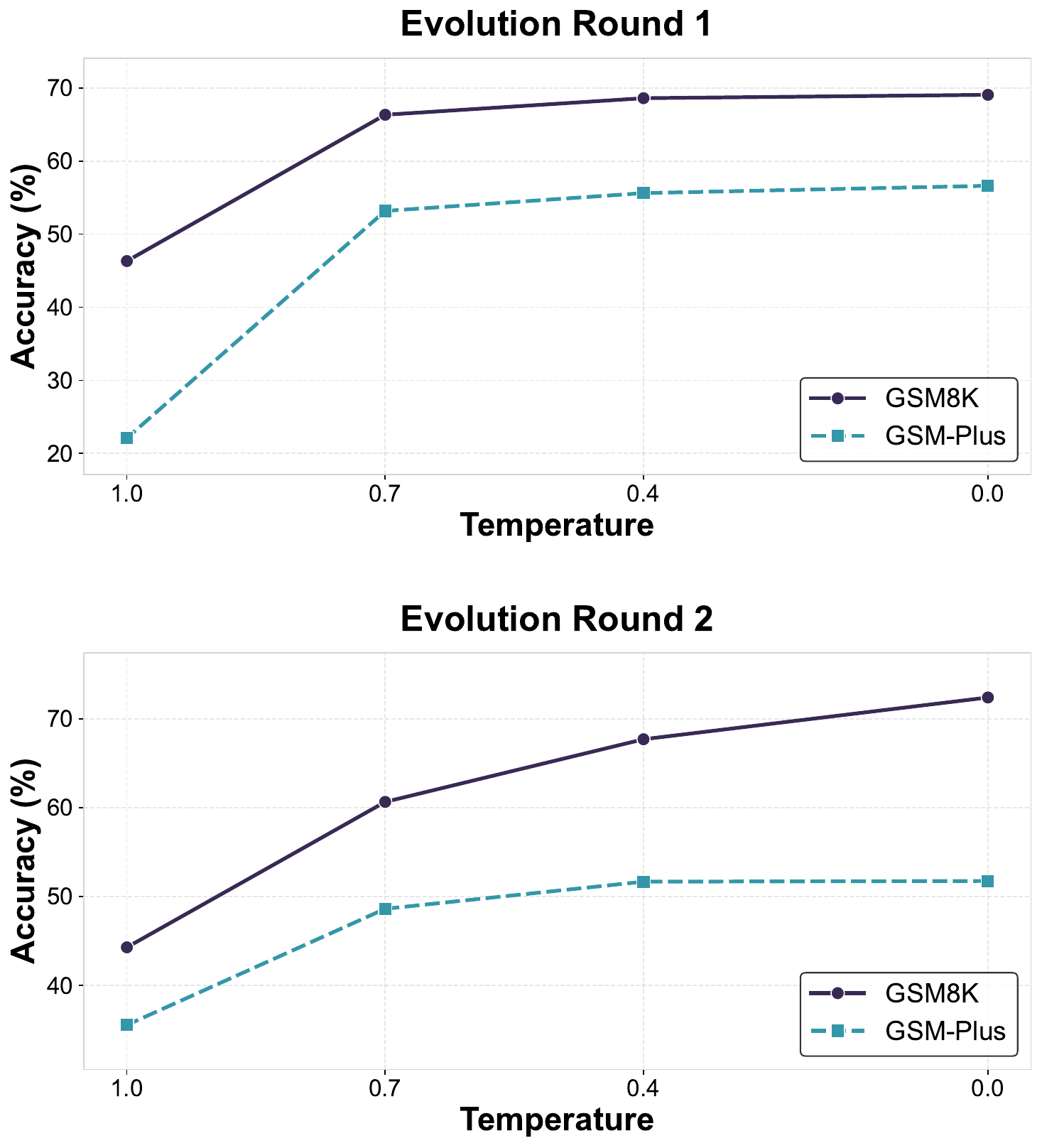}
    \caption{\textbf{Iterative fine-tuning and forgetting.}  
    Accuracy of Qwen-1.5 B after the first and second evolution rounds at four sampling temperatures.}
    \label{fig:forget}
\end{figure}

\paragraph{\textsc{7) Does iterative fine-tuning hurt?}} Figure \ref{fig:forget} plots GSM8K and GSM-Plus accuracy for Qwen-1.5B after the first and second evolution rounds under four sampling temperatures. When we keep the original exploratory setting (\(T=1.0\)) the model loses 2.0 pts on GSM8K and gains only 13.5 pts on GSM-Plus-well below the +33.5 pts it achieved in Round 1-confirming a clear case of catastrophic forgetting. Lowering the temperature stabilises training: at \(T=0.4\) Round-2 accuracy is within 0.9 pts of Round 1 on GSM-Plus and almost fully recovers on GSM8K; a deterministic schedule (\(T=0.0\)) even adds +3.3 pts on GSM8K but plateaus on GSM-Plus.

The mechanism is visible in the KL divergence between successive students. At \(T=1.0\) we measure \(\text{KL}_{\text{evo}}{=}0.37\) for Qwen-1.5B, whereas \(T=0.4\) cuts this to 0.19 and \(T=0.0\) to 0.11, matching the reduction in forgetting. We therefore adopt a linear decay from 0.7 in Round 1 to 0.3 in later rounds for all models up to 3B parameters; larger models did not require temperature adjustment.

\section{Conclusion}

In this paper, we introduced the \textsc{Debate, Train, Evolve (DTE)} framework, a novel approach enabling language models to autonomously enhance their reasoning capabilities by leveraging multi-agent debate traces. Our \textsc{Reflect-Critique-Refine} prompting strategy significantly improved debate quality, reducing sycophancy and reasoning errors. Experiments demonstrated substantial accuracy gains, notably an average improvement of \textbf{8.92\%} accuracy on the challenging GSM-PLUS dataset. Additionally, we showed strong cross-domain generalization, confirming that our approach captures general reasoning skills rather than dataset-specific patterns. Importantly, DTE effectively combines the benefits of multi-agent debate with the computational efficiency of single-model inference.

\section*{Acknowledgements}
This work was supported by the NSF \#2442253, NSF NAIRR Pilot with PSC Neocortex and NCSA Delta, Cisco Research, NVIDIA, Amazon, the Commonwealth Cyber Initiative, the Amazon–Virginia Tech Center for Efficient and Robust Machine Learning, the Sanghani Center for AI and Data Analytics at Virginia Tech, the Virginia Tech Innovation Campus, Children's National Hospital, and the Fralin Biomedical Research Institute at Virginia Tech. The views, findings, conclusions, and recommendations expressed in this work are those of the authors and do not necessarily reflect the opinions of the funding agencies.

\subsection*{Limitations}
\label{section:6}
Despite its effectiveness, our approach has certain limitations. \textbf{Firstly,} iterative fine-tuning within the DTE framework can cause catastrophic forgetting, particularly evident in smaller language models (\textless3B parameters), leading to potential model collapse. Although we explored several mitigation strategies, completely eliminating this issue remains challenging. \textbf{Secondly,} our framework assumes the availability of high-quality initial debate traces; thus, its efficacy may degrade if debates are of poor quality or if initial agent performance is weak. \textbf{Third,} our study primarily focused on structured reasoning tasks like mathematical and commonsense reasoning. The applicability and effectiveness of DTE on less structured or more open-ended tasks, such as natural language generation or dialogue systems, require further investigation. \textbf{Lastly,} although computationally efficient compared to traditional MAD setups, DTE still incurs higher training costs than standard single-model fine-tuning. Future work should aim to optimize the framework further, enhancing its practicality and accessibility.

\section*{Ethics Statement}
This study explore the self-evolution of language models using publicly available benchmarks and datasets such as GSM8K, MATH, ARC, GPQA, and CommonsenseQA. All data used in our experiments are non-sensitive and freely accessible, ensuring compliance with ethical research standards and reproducibility. Our method involves fine-tuning on model-generated content, without introducing or relying on any human-annotated private data. 

\noindent
\paragraph{AI Assistance:} We used ChatGPT assistance for parts of the Appendix, such as generating LaTeX code for tables and refining text written by the authors. All AI-generated content was carefully reviewed and revised by the authors to ensure accuracy and clarity.

\bibliography{custom}

\appendix

\onecolumn

\clearpage
\addtocontents{toc}{\protect\setcounter{tocdepth}{3}}

\renewcommand{\contentsname}{Contents of the Appendix}

\tableofcontents


\clearpage

\section{Datasets Details}
\label{sec:datasets}

We evaluate our approach on \textbf{seven} diverse reasoning benchmarks that test different aspects of model capabilities. Each dataset was chosen to provide complementary challenges in reasoning tasks. Table~\ref{tab:dataset_splits} summarizes the dataset statistics.

\begin{table}[ht]
\centering
\begin{tabular}{lccc}
\toprule
\textbf{Dataset}   & \textbf{Train} & \textbf{Validation} & \textbf{Test} \\
\midrule
GSM8K              & 7,473        & --                  & 1,319       \\
GSM-Plus           & --           & 10,552              & 2,400       \\
MATH               & 7,500        & --                  & 5,000       \\
ARC-Easy           & 2,251        & 570                 & 2,376       \\
ARC-Challenge      & 1,119        & 299                 & 1,172       \\
GPQA Main          & --           & --                  & 448         \\
CommonsenseQA      & 9,741        & 1,221               & 1,140       \\
\bottomrule
\end{tabular}
\caption{Dataset statistics. GSM8K and MATH provide only train and test splits, while GPQA Main contains only test questions.}
\label{tab:dataset_splits}
\end{table}

\textbf{GSM8K} \citep{cobbe2021trainingverifierssolvemath} contains 8,790 grade school math word problems requiring multi-step reasoning. We use 7,473 training examples and evaluate on 1,319 test problems. Each problem needs 2-8 reasoning steps to solve.

\textbf{GSM-Plus} \citep{li2024gsmpluscomprehensivebenchmarkevaluating} provides 2,400 adversarial variations of GSM8K problems designed to test robustness. These problems include more complex numerical values and additional reasoning steps compared to the original dataset.

\textbf{MATH} \citep{hendrycks2021measuring} consists of 12,500 competition mathematics problems from AMC 10, AMC 12, AIME, and other competitions. Problems span topics from algebra to calculus with difficulty levels from 1 to 5. We use the standard splits of 7,500 training and 5,000 test problems.

\textbf{ARC} \citep{clark2018thinksolvedquestionanswering} includes science questions at two difficulty levels. ARC-Easy has 2,251 training, 570 validation, and 2,376 test questions answerable by middle school students. ARC-Challenge contains 1,119 training, 299 validation, and 1,172 test questions that are challenging for retrieval-based methods.

\textbf{GPQA Main} \citep{rein2024gpqa} presents 448 graduate-level multiple-choice questions in biology, physics, and chemistry. These expert-written questions are designed to be ``Google-proof''- skilled non-experts achieve only 34\% accuracy despite unrestricted web access. We use this as a test-only benchmark.

\textbf{CommonsenseQA} \citep{talmor-etal-2019-commonsenseqa} requires commonsense reasoning with 9,741 training, 1,221 validation, and 1,140 test questions. Questions test knowledge that goes beyond factual recall.

\section{Implementation Details}
\label{sec:implementation}

\textbf{Training Setup.} We implement GRPO training using the Unsloth\footnote{\url{https://github.com/unslothai/unsloth}} and TRL\footnote{\url{https://github.com/huggingface/trl}} libraries for efficient parameter-efficient fine-tuning. We apply QLoRA \citep{dettmers2023qloraefficientfinetuningquantized} with rank 128 to attention and feed-forward modules (query, key, value, output, gate, up, down projections). Training uses 8-bit AdamW optimization with $\beta_1$=0.9, $\beta_2$=0.99, weight decay 0.1, and learning rate $5 \times 10^{-6}$ with cosine decay and 10\% warmup. We train for 10,000 steps with batch size 8.

\textbf{Reward Function.} We design a multi-component reward to encourage both correct answers and proper formatting: (1) answer correctness reward with weight 2.0, (2) XML format adherence reward with weight 0.5, (3) numeric response reward with weight 0.5, and (4) tag-counting reward with weight 0.5. Models output structured responses using \texttt{<reasoning>} and \texttt{<answer>} XML tags for consistent evaluation.

\textbf{Computational Resources.} Training runs on NVIDIA H100 (80GB), A100 (80GB), L40 (48GB), and A40 (48GB) GPUs. A single evolution round for a 7B model takes approximately 68 hours on one A100 GPU, consuming about 9600 GPU-hours total. Larger models scale near-linearly with parameter count.

\textbf{Inference Setup.} We use vLLM \citep{kwon2023efficientmemorymanagementlarge}\footnote{\url{https://docs.vllm.ai/en/latest/}} for efficient inference with dynamic GPU allocation. Multi-GPU setups use Hugging Face Accelerate\footnote{\url{https://github.com/huggingface/accelerate}} for model sharding and optimization. During debate, we sample at temperature 1.0 for exploration or 0.0 for deterministic responses.

\textbf{Software and Licenses.} All experiments use open-source software. Unsloth and TRL are released under Apache 2.0 license. vLLM uses Apache 2.0 license. All datasets are publicly available with appropriate licenses for research use: GSM8K (MIT), ARC (CC BY-SA 4.0), CommonsenseQA (MIT), MATH (MIT), GSM-Plus (Apache 2.0), and GPQA (available for research with usage restrictions to prevent leakage). Our code and model checkpoints will be released under Apache 2.0 license.

\textbf{Hyperparameter Selection.} We selected hyperparameters through preliminary experiments on validation sets. Key GRPO parameters include clipping threshold $\epsilon$=0.2, KL coefficient $\beta$=0.02, and group size $G$=8. These values balance exploration with training stability across model sizes.

\section{\textsc{Reflect–Critique–Refine} Prompt Design}

\begin{prompt}{RCR Prompting for Math Reasoning Datasets (GSM8K, GSM-Plus)}{}

\textbf{Prompt Template}

You are Agent \{self.agent\_id\} in a multi-agent debate to solve the following math problem:

Problem: \{question\}

\{own\_previous\}

Here are the solutions from other agents:  
\{context\}

This is debate round \{round\_num\}. Please carefully analyze all solutions—including your own—identify any errors in reasoning, and provide your revised solution.

\begin{itemize}
  \item \hl{\textbf{If you believe your previous answer is correct, explain why and defend it.}}
  \item \hl{\textbf{If you believe you made an error, explain the error and provide a corrected solution.}}
  \item \hl{\textbf{If you believe another agent's answer is correct, explain why you agree with it.}}
\end{itemize}

Your final answer must be in the format \(\boxed{\{answer\}}\) at the end.

\end{prompt}

\begin{prompt}{RCR Prompting for Science Reasoning Datasets (ARC-E, ARC-C)}{}
\textbf{Prompt Template}
You are Agent \{self.agent\_id\} in a multi-agent debate to solve the following scientific problem:

Problem: \{question\}

\{own\_previous\}

Here are the solutions from other agents:  

\{context\}

This is debate round \{round\_num\}. Please carefully analyze all solutions—including your own—identify any misconceptions or flawed scientific reasoning, and provide your revised solution.

\begin{itemize}
  \item \hl{\textbf{If you believe your previous answer is correct, explain the scientific principles supporting your answer.}}
  \item \hl{\textbf{If you believe you made an error, explain the scientific misconception and provide a corrected solution.}}
  \item \hl{\textbf{If you believe another agent's answer is correct, explain why their scientific reasoning is sound.}}
  
\end{itemize}

Your final answer must be in the format \(\boxed{\{answer\}}\) at the end.
\end{prompt}

\begin{prompt}{RCR Prompting for Commonsense Reasoning Datasets (CSQA)}{}
\textbf{Prompt Template}
You are Agent \{self.agent\_id\} in a multi-agent debate to solve the following commonsense reasoning problem:

Problem: \{question\}

\{own\_previous\}

Here are the solutions from other agents:  

\{context\}

This is debate round \{round\_num\}. Please carefully analyze all solutions—including your own—identify any flawed assumptions or logical inconsistencies, and provide your revised solution.

\begin{itemize}
  \item \hl{\textbf{If you believe your previous answer is correct, explain the logical reasoning and real-world knowledge supporting it.}}
  \item \hl{\textbf{If you believe you made an error, explain the flawed assumption or inconsistency and provide a corrected solution.}}
  \item \hl{\textbf{If you believe another agent's answer is correct, explain why their reasoning aligns with commonsense knowledge.}}
\end{itemize}

Your final answer must be in the format \(\boxed{\{answer\}}\) at the end.

\end{prompt}

\clearpage

\section{Additional Self-Evolution Results}

In this section, we present a comprehensive analysis of our \textsc{Debate, Train, Evolve} framework across multiple experimental settings. We first examine the impact of various GRPO configurations, followed by analyses of multi-round training effects, and finally cross-domain generalization results. Our experiments utilize a diverse set of models ranging from 1.5B to 14B parameters and evaluate performance on challenging reasoning benchmarks including GSM8K, GSM-Plus, ARC-Challenge, ARC-Easy, and CommonsenseQA.

\subsection{Complete GRPO results (all steps, temperature)}

We begin by investigating how different GRPO hyperparameters affect model performance. Tables \ref{tab:grpo_gsm8k}, \ref{tab:grpo_gsm_plus}, and \ref{tab:grpo_arc_c} present results across three datasets (GSM8K, GSM-Plus, and ARC-Challenge) for six different model configurations, varying training steps (2000, 5000, and 10000) and sampling temperatures (0.8 and 0.2).

Several key patterns emerge from these results. First, we observe that larger models (7B+) generally maintain or improve their performance through GRPO fine-tuning, while smaller models (particularly Llama-3B) occasionally exhibit catastrophic forgetting at higher step counts. Second, lower temperature (0.2) typically yields more stable optimization trajectories for most model configurations, especially at higher step counts. This supports our hypothesis that constraining policy drift during fine-tuning is crucial for successful reasoning evolution.

Notably, the Qwen-2.5-3B model demonstrates remarkable stability across configurations, with consistent performance gains on GSM-Plus (from 61.75\% to 69.50\%) and robust maintenance of GSM8K performance. In contrast, the Llama-3B model shows significant performance degradation at higher step counts with 0.8 temperature, dropping to near-random performance (2.73\%) after 10000 steps on GSM8K, while maintaining better stability at 0.2 temperature.

For ARC-Challenge, we observe that all models benefit from MAD evolution, with particularly strong gains for Qwen-2.5-7B (from 87.22\% to 91.64\%) and Llama-8B (from 77.65\% to 85.07\%). These results suggest that our framework effectively generalizes across both mathematical reasoning and scientific question-answering domains.

\subsection{Complete Round 2 MAD Results}

After the first round of GRPO fine-tuning, we evaluated the performance of models in a multi-agent debate setting to assess how evolution affects collaborative reasoning. Table \ref{tab:mad_round2} presents these results across different debate configurations: exponential temperature scaling (Exp), default settings (Default), temperature-4 settings (temp4), and deterministic setting (Det).

The MAD Round 2 results demonstrate that evolved models generally maintain their collaborative reasoning capabilities after GRPO fine-tuning. For most models, MAD performance after evolution either improves or remains comparable to the original MAD results. The Qwen-2.5-7B model, for instance, achieves 77.75\% accuracy on GSM-Plus under the temp4 configuration, which represents a 3.58\% improvement over its original MAD performance.

Interestingly, we observe that different debate configurations yield varying results across model sizes. Smaller models like Qwen-2.5-1.5B show significant performance variation across configurations, with deterministic settings yielding the best results (69.07\% on GSM8K and 56.62\% on GSM-Plus). In contrast, larger models like Qwen-2.5-7B demonstrate more consistent performance across configurations.

The exponential temperature scaling configuration generally underperforms other settings, particularly for smaller models. This suggests that controlled diversity in debate is beneficial, but excessive exploration may hinder collaborative reasoning effectiveness.

\subsection{GRPO round 2 results}

To investigate the effects of iterative evolution, we conducted a second round of GRPO fine-tuning on models that had already undergone one round of evolution. Table \ref{tab:grpo_round2} presents these results for four model configurations across two datasets (GSM8K and GSM-Plus).

The second round of GRPO training reveals interesting dynamics in model evolution. For the Qwen family of models, we observe continued performance improvements or stability across most configurations. The Qwen-2.5-7B model, for instance, achieves further gains on GSM-Plus, reaching 73.75\% accuracy (a 5.13\% improvement over its first round GRPO performance).

However, the Llama-3B model exhibits significant performance degradation in certain configurations, particularly at higher step counts with 0.8 temperature (dropping to 35.63\% on GSM8K and 23.02\% on GSM-Plus). This reinforces our finding that smaller models are more sensitive to optimization instability during iterative fine-tuning. Importantly, using a lower temperature of 0.2 substantially mitigates this issue, allowing the Llama-3B model to maintain competitive performance (73.62\% on GSM8K) even after two rounds of evolution.

These results highlight the importance of careful hyperparameter selection during iterative self-evolution, particularly for smaller models that may be more susceptible to catastrophic forgetting or excessive policy drift.

\subsection{Complete Round 3 MAD Results}

To investigate the long-term stability of collaborative reasoning capabilities through multiple evolution iterations, we conducted a third round of multi-agent debate after the second round of GRPO fine-tuning. Table \ref{tab:mad_round3} presents these results for three Qwen models across the same four debate configurations.

The Round 3 MAD results reveal interesting trends in iterative evolution. For the Qwen-2.5-3B and Qwen-2.5-7B models, performance remains relatively stable across debate configurations, indicating robust retention of reasoning capabilities through multiple fine-tuning iterations. However, the Qwen-2.5-1.5B model shows more variable performance, particularly under the exponential temperature scaling configuration where it drops to 44.28\% on GSM8K.

Notably, the deterministic debate setting (Det) consistently produces the best or near-best performance across all models and datasets, suggesting that reduced randomness in collaborative reasoning becomes increasingly important after multiple evolution rounds. This aligns with our hypothesis that controlling policy drift is crucial for successful iterative evolution.

The stability of larger models (3B+) across multiple evolution rounds indicates that our \textsc{Debate, Train, Evolve} framework can support continuous improvement without substantial performance degradation when applied to sufficiently capable base models.

\subsection{Complete Cross Domain Task Results}

A key question for self-evolution frameworks is whether improvements generalize beyond the training domain. Table \ref{tab:cross_domain} presents results for models fine-tuned on either GSM8K or GSM-Plus and evaluated on multiple out-of-domain tasks including ARC-Easy, ARC-Challenge, and CommonsenseQA.

The cross-domain results reveal impressive generalization capabilities. Models fine-tuned on mathematical reasoning tasks (GSM8K and GSM-Plus) show substantial performance improvements not only on the alternative math dataset but also on science and commonsense reasoning benchmarks. For instance, the Qwen-2.5-14B model fine-tuned on GSM8K achieves 98.19\% accuracy on ARC-Easy, 93.69\% on ARC-Challenge, and 83.70\% on CommonsenseQA.

Interestingly, models fine-tuned on GSM-Plus generally perform better on GSM8K than vice versa. For example, the Qwen-2.5-1.5B model achieves 73.09\% on GSM8K when fine-tuned on GSM-Plus, but only 51.21\% on GSM-Plus when fine-tuned on GSM8K. This asymmetry suggests that GSM-Plus may require more diverse reasoning strategies that transfer well to simpler tasks.

The strong cross-domain performance demonstrates that our \textsc{Debate, Train, Evolve} framework does not simply optimize for specific datasets but instead enhances fundamental reasoning capabilities that generalize across tasks. This is a critical advantage over traditional supervised fine-tuning approaches that often exhibit limited transferability.

\begin{table*}[htbp]
\centering

\resizebox{\textwidth}{!}{%
\begin{tabular}{lccccccccc}
\toprule
\multirow{2}{*}{\textbf{Model}} & \multicolumn{2}{c}{\textbf{Base Performance}} & \multirow{2}{*}{\textbf{MAD}} & \multicolumn{3}{c}{\textbf{GRPO (Temperature 0.8)}} & \multicolumn{3}{c}{\textbf{GRPO (Temperature 0.2)}} \\
\cmidrule(lr){2-3} \cmidrule(lr){5-7} \cmidrule(lr){8-10}
 & \textbf{Train} & \textbf{Test} & & \textbf{2k steps} & \textbf{5k steps} & \textbf{10k steps} & \textbf{2k steps} & \textbf{5k steps} & \textbf{10k steps} \\
\midrule
Qwen-2.5-1.5B & 81.55 & 62.77 & 72.33 & 67.78 & 71.42 & 71.04 & 73.09 & 66.49 & 53.98 \\
Qwen-2.5-3B & 91.28 & 84.08 & 85.14 & 85.06 & 85.14 & 86.13 & 84.00 & 86.05 & 84.38 \\
Qwen-2.5-7B & 94.29 & 90.67 & 91.21 & 88.32 & 86.73 & 84.00 & 86.96 & 86.35 & 88.02 \\
Llama-3B & 83.90 & 72.55 & 73.84 & 69.22 & 21.53 & 2.73 & 72.40 & 75.06 & 3.26 \\
Llama-8B & 89.08 & 81.73 & 82.18 & 84.61 & 85.29 & 85.22 & 86.81 & 84.91 & 0.15 \\
Qwen-2.5-14B & 94.89 & 92.80 & 93.33 & 87.72 & 89.84 & 91.81 & 86.58 & 89.34 & 93.74 \\
\bottomrule
\end{tabular}%
}
\caption{\textbf{Complete GRPO Results on GSM8K Dataset.} Results show accuracy (\%) for different models under various GRPO configurations. Training hyperparameters include learning rate of 5e-6 and context length of 256 tokens. MAD refers to Multi-Agent Debate baseline performance.}
\label{tab:grpo_gsm8k}
\end{table*}

\begin{table*}[htbp]
\centering

\resizebox{\textwidth}{!}{%
\begin{tabular}{lccccccccc}
\toprule
\multirow{2}{*}{\textbf{Model}} & \multicolumn{2}{c}{\textbf{Base Performance}} & \multirow{2}{*}{\textbf{MAD}} & \multicolumn{3}{c}{\textbf{GRPO (Temperature 0.8)}} & \multicolumn{3}{c}{\textbf{GRPO (Temperature 0.2)}} \\
\cmidrule(lr){2-3} \cmidrule(lr){5-7} \cmidrule(lr){8-10}
 & \textbf{Train} & \textbf{Test} & & \textbf{2k steps} & \textbf{5k steps} & \textbf{10k steps} & \textbf{2k steps} & \textbf{5k steps} & \textbf{10k steps} \\
\midrule
Qwen-2.5-1.5B & 42.40 & 42.00 & 51.62 & 47.49 & 54.46 & 19.00 & 52.33 & 53.04 & 55.92 \\
Qwen-2.5-3B & 61.14 & 61.75 & 67.79 & 66.21 & 66.71 & 69.13 & 64.04 & 67.25 & 68.25 \\
Qwen-2.5-7B & 68.27 & 68.62 & 74.17 & 64.71 & 73.38 & 74.71 & 67.75 & 72.54 & 74.50 \\
Llama-3B & 47.68 & 45.67 & 51.12 & 52.38 & 53.29 & 52.33 & 51.79 & 49.54 & 53.79 \\
Llama-8B & 58.56 & 55.62 & 60.79 & 64.96 & 61.58 & 66.17 & 65.08 & 63.46 & 60.46 \\
Qwen-2.5-14B & 71.11 & 71.79 & 77.25 & 70.79 & 73.54 & 75.88 & 73.00 & 73.42 & 75.62 \\
\bottomrule
\end{tabular}%
}
\caption{\textbf{Complete GRPO Results on GSM-Plus Dataset.} Results show accuracy (\%) for different models under various GRPO configurations on the more challenging GSM-Plus dataset. Training hyperparameters include learning rate of 5e-6.}
\label{tab:grpo_gsm_plus}
\end{table*}

\begin{table*}[htbp]
\centering

\resizebox{\textwidth}{!}{%
\begin{tabular}{lccccccccc}
\toprule
\multirow{2}{*}{\textbf{Model}} & \multicolumn{2}{c}{\textbf{Base Performance}} & \multirow{2}{*}{\textbf{MAD}} & \multicolumn{3}{c}{\textbf{GRPO (Temperature 0.8)}} & \multicolumn{3}{c}{\textbf{GRPO (Temperature 0.2)}} \\
\cmidrule(lr){2-3} \cmidrule(lr){5-7} \cmidrule(lr){8-10}
 & \textbf{Train} & \textbf{Test} & & \textbf{2k steps} & \textbf{5k steps} & \textbf{10k steps} & \textbf{2k steps} & \textbf{5k steps} & \textbf{10k steps} \\
\midrule
Qwen-2.5-1.5B & --- & 69.21 & 68.52 & 30.03 & 62.63 & 68.36 & 47.27 & 51.88 & 67.51 \\
Qwen-2.5-3B & --- & 83.53 & 84.64 & 81.66 & 80.29 & 83.63 & 81.91 & 79.78 & 83.95 \\
Qwen-2.5-7B & --- & 87.22 & 91.64 & 88.57 & 88.48 & 90.63 & 88.43 & 88.57 & 90.89 \\
Llama-3B & --- & 73.12 & 76.19 & 75.51 & 74.32 & 76.87 & 76.79 & 74.57 & 77.23 \\
Llama-8B & --- & 77.65 & 85.07 & 83.70 & 84.45 & 86.03 & 84.98 & 85.53 & 86.53 \\
Qwen-2.5-14B & --- & 90.27 & 93.77 & 91.81 & 92.49 & 93.13 & 91.47 & 91.47 & 92.67 \\
\bottomrule
\end{tabular}%
}
\caption{\textbf{Complete GRPO Results on ARC-Challenge Dataset.} Results show accuracy (\%) for different models under various GRPO configurations on the ARC-Challenge dataset. Training hyperparameters include learning rate of 5e-6 and context length of 128 tokens. Base train performance was not evaluated for this dataset.}
\label{tab:grpo_arc_c}
\end{table*}

\begin{table*}[htbp]
\centering

\begin{tabular}{lccccc}
\toprule
\multirow{2}{*}{\textbf{Model}} & \multirow{2}{*}{\textbf{Dataset}} & \multicolumn{2}{c}{\textbf{GRPO Round 2 (Temp 0.8)}} & \multicolumn{2}{c}{\textbf{GRPO Round 2 (Temp 0.2)}} \\
\cmidrule(lr){3-4} \cmidrule(lr){5-6}
 & & \textbf{2k steps} & \textbf{5k steps} & \textbf{2k steps} & \textbf{5k steps} \\
\midrule
\multirow{2}{*}{Qwen-2.5-1.5B} & GSM8K & 65.73 & 68.54 & 69.98 & 72.18 \\
 & GSM-Plus & 47.38 & 50.12 & 46.37 & 48.04 \\
\midrule
\multirow{2}{*}{Qwen-2.5-3B} & GSM8K & 84.84 & 86.05 & 84.46 & 84.08 \\
 & GSM-Plus & 65.71 & 67.96 & 65.67 & 67.00 \\
\midrule
\multirow{2}{*}{Qwen-2.5-7B} & GSM8K & 86.28 & 87.19 & 88.17 & 87.34 \\
 & GSM-Plus & 69.42 & 73.75 & 70.54 & 73.12 \\
\midrule
\multirow{2}{*}{Llama-3B} & GSM8K & 55.88 & 35.63 & 73.62 & 64.29 \\
 & GSM-Plus & 48.75 & 23.02 & 52.42 & 25.08 \\
\bottomrule
\end{tabular}
\caption{\textbf{Complete GRPO Round 2 Results.} Results show accuracy (\%) after second round of GRPO training across different step counts and temperature settings. All models were trained with learning rate of 5e-6 and context length of 128 tokens.}
\label{tab:grpo_round2}
\end{table*}

\begin{table*}[htbp]
\centering

\begin{tabular}{lcccccc}
\toprule
\multirow{2}{*}{\textbf{Model}} & \multirow{2}{*}{\textbf{Dataset}} & \multicolumn{4}{c}{\textbf{MAD Configuration}} \\
\cmidrule(lr){3-6}
 & & \textbf{Exp} & \textbf{Default} & \textbf{temp4} & \textbf{Det} \\
\midrule
\multirow{2}{*}{Qwen-2.5-1.5B} & GSM8K & 46.32 & 66.34 & 68.61 & 69.07 \\
 & GSM-Plus & 22.09 & 53.18 & 55.62 & 56.62 \\
\midrule
\multirow{2}{*}{Qwen-2.5-3B} & GSM8K & 84.08 & 86.66 & 86.35 & 86.50 \\
 & GSM-Plus & 69.62 & 70.25 & 69.67 & 70.29 \\
\midrule
\multirow{2}{*}{Qwen-2.5-7B} & GSM8K & 91.36 & 90.75 & 91.05 & 89.99 \\
 & GSM-Plus & 76.42 & 77.00 & 77.75 & 77.62 \\
\midrule
\multirow{2}{*}{Llama-3B} & GSM8K & 66.26 & 75.97 & 75.51 & 75.36 \\
 & GSM-Plus & 53.62 & 54.58 & 55.96 & 56.04 \\
\midrule
\multirow{2}{*}{Llama-8B} & GSM8K & 84.69 & 85.90 & 86.96 & 85.60 \\
 & GSM-Plus & 65.00 & 65.92 & 66.46 & 66.50 \\
\bottomrule
\end{tabular}
\caption{\textbf{Complete MAD Round 2 Results.} Results show accuracy (\%) for different models in multi-agent debate after first round of GRPO fine-tuning. Exp = exponential temperature scaling, Default = standard configuration, temp4 = temperature-4 settings, Det = deterministic configuration.}
\label{tab:mad_round2}
\end{table*}

\begin{table*}[htbp]
\centering

\begin{tabular}{lcccccc}
\toprule
\multirow{2}{*}{\textbf{Model}} & \multirow{2}{*}{\textbf{Dataset}} & \multicolumn{4}{c}{\textbf{MAD Configuration}} \\
\cmidrule(lr){3-6}
 & & \textbf{Exp} & \textbf{Default} & \textbf{temp4} & \textbf{Det} \\
\midrule
\multirow{2}{*}{Qwen-2.5-1.5B} & GSM8K & 44.28 & 60.65 & 67.70 & 72.40 \\
 & GSM-Plus & 35.54 & 48.62 & 51.67 & 51.75 \\
\midrule
\multirow{2}{*}{Qwen-2.5-3B} & GSM8K & 83.78 & 85.60 & 85.75 & 86.13 \\
 & GSM-Plus & 63.67 & 63.42 & 64.16 & 64.47 \\
\midrule
\multirow{2}{*}{Qwen-2.5-7B} & GSM8K & 89.76 & 91.05 & 90.90 & 91.13 \\
 & GSM-Plus & 69.67 & 69.85 & 70.50 & 69.88 \\
\bottomrule
\end{tabular}
\caption{\textbf{Complete MAD Round 3 Results.} Results show accuracy (\%) for different models in multi-agent debate after second round of GRPO fine-tuning. Exp = exponential temperature scaling, Default = standard configuration, temp4 = temperature-4 settings, Det = deterministic configuration.}
\label{tab:mad_round3}
\end{table*}

\begin{table*}[htbp]
\centering

\resizebox{\textwidth}{!}{%
\begin{tabular}{lcccccc}
\toprule
\multirow{2}{*}{\textbf{Model}} & \multirow{2}{*}{\textbf{Fine-tuned on}} & \multicolumn{5}{c}{\textbf{Evaluation Dataset}} \\
\cmidrule(lr){3-7}
 & & \textbf{GSM8K} & \textbf{GSM-Plus} & \textbf{ARC-Easy} & \textbf{ARC-Challenge} & \textbf{CommonsenseQA} \\
\midrule
\multirow{2}{*}{Qwen-2.5-1.5B} & GSM8K & --- & 51.21 & 85.02 & 69.88 & 64.29 \\
 & GSM-Plus & 73.09 & --- & 85.10 & 69.45 & 64.21 \\
\midrule
\multirow{2}{*}{Qwen-2.5-3B} & GSM8K & --- & 65.54 & 93.94 & 84.30 & 75.92 \\
 & GSM-Plus & 86.50 & --- & 94.15 & 84.13 & 75.92 \\
\midrule
\multirow{2}{*}{Qwen-2.5-7B} & GSM8K & --- & 69.63 & 96.42 & 91.72 & 82.96 \\
 & GSM-Plus & 91.81 & --- & 96.38 & 90.87 & 82.88 \\
\midrule
\multirow{2}{*}{Llama-3B} & GSM8K & --- & 52.38 & 87.12 & 72.01 & 68.14 \\
 & GSM-Plus & 76.35 & --- & 86.57 & 69.20 & 68.55 \\
\midrule
\multirow{2}{*}{Llama-8B} & GSM8K & --- & 63.75 & 93.01 & 84.39 & 74.12 \\
 & GSM-Plus & 86.88 & --- & 93.98 & 85.49 & 73.87 \\
\midrule
\multirow{2}{*}{Qwen-2.5-14B} & GSM8K & --- & 73.46 & 98.19 & 93.69 & 83.70 \\
 & GSM-Plus & 93.33 & --- & 97.98 & 94.28 & 82.23 \\
\bottomrule
\end{tabular}%
}
\caption{\textbf{Complete Cross Domain Task Results.} Results show accuracy (\%) on various datasets after fine-tuning on either GSM8K or GSM-Plus. Dashes (---) indicate that evaluation was not performed on the same dataset used for fine-tuning.}
\label{tab:cross_domain}
\end{table*}

\clearpage

\clearpage

\section{Complete Results of Large-scale Empirical Study on MAD using RCR Prompting}
\label{rcr: appendix}
This section presents a comprehensive analysis of our large-scale empirical investigation into Multi-Agent Debate (MAD) using Recursive Critical Reflection (RCR) prompting across five diverse benchmarks: GSM8K, GSM-Plus, ARC-Easy, ARC-Challenge, and CommonsenseQA. Through extensive experimentation involving various model combinations and parameter settings, we evaluate how collaborative reasoning among multiple language model agents affects problem-solving performance.

\subsection{Evaluation Metrics and Methodology}

To facilitate systematic comparison and analysis of debate outcomes, we track the following key metrics across all debate configurations:

\begin{itemize}
    \item \textbf{Accuracy}: The primary performance measure, representing the percentage of problems correctly solved after the debate process concludes.
    
    \item \textbf{$\Delta$ (Performance Delta)}: Measures the performance change relative to appropriate baselines. We report several variants including:
    \begin{itemize}
        \item $\Delta$ (vs Base): Change compared to the single agent's performance
        \item $\Delta$ (vs Lower Agent): Change compared to the lower-performing agent in cross-agent debates
        \item $\Delta$ (vs Upper Agent): Change compared to the better-performing agent in cross-agent debates
        \item $\Delta$ (vs Lowest): Change compared to the lowest-performing agent in three-agent settings
    \end{itemize}
    
    \item \textbf{Debate Rounds}: The average number of interaction rounds required to reach consensus or the maximum allowed limit, indicating debate efficiency.
    
    \item \textbf{Sycophancy}: A normalized measure (per data points) quantifying the tendency of agents to abandon their answers in favor of matching another agent's previous response, providing insights into social influence dynamics.
    
    \item \textbf{State Transitions}: Tracked as C$\rightarrow$I (correct to incorrect) and I$\rightarrow$C (incorrect to correct) counts, these reveal the qualitative nature of answer changes during debate.
    
    \item \textbf{Debate Helped}: The overall count of instances where the debate process improved the final outcome compared to initial responses.
\end{itemize}

Our evaluation spans multiple dimensions of agent configuration:

\begin{itemize}
    \item \textbf{Agent Settings}: We systematically vary temperature parameter across four settings:
    \begin{itemize}
        \item Default: Balanced temperature
        \item Deterministic (Det.): Lower temperature for more consistent outputs
        \item Exploratory (Exp.): Higher temperature for more diverse responses
        \item Mixed: Combinations of the above settings across different agents
    \end{itemize}
    
    \item \textbf{Debate Structures}: We investigate four primary debate configurations:
    \begin{itemize}
        \item Single-Model Debate: Multiple instances of the same model with varied parameter settings
        \item Cross-Agent Debate: Two different models debating with various parameter settings
        \item Three Identical Agents: Three instances of the same model with potentially different settings
        \item Three Varied Agents: Three different models engaging in debate
    \end{itemize}
\end{itemize}

\subsection{Overview of Results Organization}

Our extensive experimental results are organized in Tables 13-32, systematically covering all five datasets with the four debate configurations described above. For each dataset, we present:

\begin{itemize}
    \item Table set 1 (Tables 13-16): Performance on GSM8K
    \item Table set 2 (Tables 17-20): Performance on GSM-Plus
    \item Table set 3 (Tables 21-24): Performance on ARC-Easy
    \item Table set 4 (Tables 25-28): Performance on ARC-Challenge
    \item Table set 5 (Tables 29-32): Performance on CommonsenseQA
\end{itemize}

\subsection{Key Findings and Patterns}

\subsubsection{Impact of Agent Settings}

Our analysis reveals that agent parameter settings significantly influence debate outcomes across all datasets. We observe that while the Default setting provides reliable performance, Exploratory settings often lead to higher variance in outcomes, sometimes yielding exceptional improvements but also risking performance degradation. The Deterministic setting generally produces more consistent but potentially conservative results.

The sycophancy metric proves particularly informative, showing higher values in debates between models with substantial performance gaps. This suggests that lower-performing models tend to defer to higher-performing ones, which can be either beneficial or detrimental depending on the initial state distribution.

\subsubsection{Cross-Model Debate Dynamics}

In cross-agent debates (Tables 10-14), we find that pairing models with complementary strengths often produces synergistic effects. The $\Delta$ metrics relative to both upper and lower agents reveal important patterns: when a high-performing model debates with a weaker one, the debate outcome typically falls between their individual performances but closer to the stronger model's baseline.

State transitions (C$\rightarrow$I and I$\rightarrow$C) provide valuable insights into debate quality. A high I$\rightarrow$C rate coupled with a low C$\rightarrow$I rate indicates constructive debate where correct reasoning prevails, while the opposite pattern signals problematic dynamics where convincing but incorrect reasoning dominates.

\subsubsection{Three-Agent Debate Effectiveness}

The introduction of a third agent creates more complex interaction patterns. Three-agent debates consistently show lower sycophancy rates compared to two-agent settings, suggesting that the presence of multiple perspectives reduces blind conformity. When all three agents are identical, we observe that diversity in parameter settings typically outperforms homogeneous settings.

In three varied agent debates, we find particularly interesting results when combining models of different sizes and architectures. As shown in Table \ref{tab:three_varied_agents_debate_gsm8k_lowest_delta}, certain combinations like "Qwen-2.5-3B + Phi-mini-3.8B + Llama-3.1-3B" achieve accuracy improvements even compared to the highest-performing individual agent, suggesting effective complementarity between these models' reasoning approaches.

\subsubsection{Dataset-Specific Patterns}

Our results indicate substantial variation in debate effectiveness across different datasets:

\begin{itemize}
    \item \textbf{GSM8K and GSM+}: Harder Mathematical reasoning tasks (GSM-Plus) show the most consistent benefits from debate, with average debate rounds typically higher than other datasets, suggesting that step-by-step verification is particularly valuable for these problems.
    
    \item \textbf{ARC-Easy and ARC-Challenge}: Multiple-choice science questions reveal interesting patterns where sycophancy is generally lower, but debate can still improve performance when appropriately configured.
    
    \item \textbf{CommonsenseQA}: This dataset exhibits unique characteristics where debates tend to conclude more quickly, suggesting that commonsense reasoning may be less amenable to explicit verification through debate.
\end{itemize}

\subsection{Conclusion}

Tables 13-32 collectively present a comprehensive empirical foundation for understanding the effects of Multi-Agent Debate using RCR prompting across diverse reasoning tasks. The metrics reveal nuanced patterns in how debate influences performance, with clear evidence that appropriate configuration of debate participants and settings can yield substantial improvements over single-agent performance.

The consistent tracking of accuracy, deltas, debate rounds, sycophancy, and state transitions provides a multi-dimensional view of debate quality beyond simple performance measures. These results demonstrate that MAD is not universally beneficial but rather depends critically on the specific combination of models, parameter settings, and problem domains. Our findings establish an important baseline for future research on collaborative reasoning between language models, highlighting both the potential and the challenges of multi-agent approaches to complex problem-solving.


\begin{table*}[htbp]
\centering

\begin{adjustbox}{width=\textwidth,center} 
\sisetup{round-mode=places,round-precision=2} 
\begin{tabular}{@{}ccccS[table-format=2.2]S[table-format=3.2]S[table-format=1.2]S[table-format=4.2]cccS[table-format=1.2]@{}} 
\toprule
\textbf{Agent 1} & \textbf{Agent 2} & \textbf{Agent Settings} & \textbf{MAD Accuracy} & \multicolumn{1}{c}{\textbf{$\Delta$}} & \textbf{Debate} & \multicolumn{1}{c}{\textbf{Sycophancy}} & \textbf{C$\rightarrow$I} & \textbf{I$\rightarrow$C} & \textbf{Debate} \\
 & & & \textbf{(RCR Prompting)} &  & \textbf{Rounds} & \multicolumn{1}{c}{\textbf{(Avg / 1319)}} & & & \textbf{Helped} \\
 & & & &  & \textbf{(Avg)} & & & & \textbf{(Overall)} \\
\midrule

Qwen-2.5-0.5B & Qwen-2.5-0.5B & Both: Default & 47.38 & \textcolor{darkgreen}{5.38 $\uparrow$} & 1.6 & 1.17 & 156 & 251 & 220 \\
Qwen-2.5-0.5B & Qwen-2.5-0.5B & Both: Deterministic & 47.31 & \textcolor{darkgreen}{5.31 $\uparrow$} & 0 & 0.00 & 0 & 0 & 0 \\
Qwen-2.5-0.5B & Qwen-2.5-0.5B & Both: Exploratory & 39.20 & \textcolor{darkred}{2.8 $\downarrow$} & 2.19 & 1.25 & 185 & 274 & 234 \\
Qwen-2.5-0.5B & Qwen-2.5-0.5B & Both: Det. \& Exp. & 43.14 & \textcolor{darkgreen}{1.14 $\uparrow$} & 1.89 & 1.09 & 185 & 262 & 226 \\
\midrule
Qwen-2.5-1.5B & Qwen-2.5-1.5B & Both: Default & 70.89 & \textcolor{darkgreen}{8.12 $\uparrow$} & 0.86 & 0.70 & 101 & 352 & 317 \\
Qwen-2.5-1.5B & Qwen-2.5-1.5B & Both: Deterministic & 63.46 & \textcolor{darkgreen}{0.69 $\uparrow$} & 0 & 0.00 & 0 & 0 & 0 \\
Qwen-2.5-1.5B & Qwen-2.5-1.5B & Both: Exploratory & 71.57 & \textcolor{darkgreen}{8.8 $\uparrow$} & 1.05 & 0.84 & 94 & 449 & 399 \\
Qwen-2.5-1.5B & Qwen-2.5-1.5B & Both: Det. \& Exp. & 72.33 & \textcolor{darkgreen}{9.56 $\uparrow$} & 0.98 & 0.71 & 99 & 423 & 377 \\
\midrule
Qwen-2.5-3B & Qwen-2.5-3B & Both: Default & 86.05 & \textcolor{darkgreen}{0.91 $\uparrow$} & 0.31 & 0.21 & 55 & 115 & 104 \\
Qwen-2.5-3B & Qwen-2.5-3B & Both: Deterministic & 84.99 & \textcolor{darkred}{0.15 $\downarrow$} & 0 & 0.00 & 0 & 0 & 0 \\
Qwen-2.5-3B & Qwen-2.5-3B & Both: Exploratory & 85.52 & \textcolor{darkgreen}{0.38 $\uparrow$} & 0.35 & 0.26 & 62 & 116 & 103 \\
Qwen-2.5-3B & Qwen-2.5-3B & Both: Det. \& Exp. & 86.28 & \textcolor{darkgreen}{1.14 $\uparrow$} & 0.34 & 0.19 & 50 & 106 & 101 \\
\midrule
Qwen-2.5-7B & Qwen-2.5-7B & Both: Default & 91.74 & \textcolor{darkgreen}{1.07 $\uparrow$} & 0.16 & 0.13 & 28 & 53 & 49 \\
Qwen-2.5-7B & Qwen-2.5-7B & Both: Deterministic & 90.60 & \textcolor{darkred}{0.07 $\downarrow$} & 0 & 0.00 & 0 & 0 & 0 \\
Qwen-2.5-7B & Qwen-2.5-7B & Both: Exploratory & 91.21 & \textcolor{darkgreen}{0.54 $\uparrow$} & 0.18 & 0.15 & 27 & 59 & 57 \\
Qwen-2.5-7B & Qwen-2.5-7B & Both: Det. \& Exp. & 91.51 & \textcolor{darkgreen}{0.84 $\uparrow$} & 0.18 & 0.15 & 33 & 57 & 55 \\
\midrule
Qwen-2.5-14B & Qwen-2.5-14B & Both: Default & 93.48 & \textcolor{darkgreen}{0.68 $\uparrow$} & 0.11 & 0.13 & 22 & 46 & 43 \\
Qwen-2.5-14B & Qwen-2.5-14B & Both: Deterministic & 93.18 & \textcolor{darkgreen}{0.38 $\uparrow$} & 0 & 0.00 & 0 & 0 & 0 \\
Qwen-2.5-14B & Qwen-2.5-14B & Both: Exploratory & 93.33 & \textcolor{darkgreen}{0.53 $\uparrow$} & 0.11 & 0.12 & 20 & 48 & 48 \\
Qwen-2.5-14B & Qwen-2.5-14B & Both: Det. \& Exp. & 93.63 & \textcolor{darkgreen}{0.83 $\uparrow$} & 0.13 & 0.15 & 24 & 44 & 39 \\
\midrule
Qwen-2.5-32B & Qwen-2.5-32B & Both: Default & 95.00 & \textcolor{darkgreen}{0.08 $\uparrow$} & 0.05 & 0.06 & 11 & 21 & 20 \\
Qwen-2.5-32B & Qwen-2.5-32B & Both: Deterministic & 94.77 & \textcolor{darkred}{0.15 $\downarrow$} & 0 & 0.00 & 0 & 0 & 0 \\
Qwen-2.5-32B & Qwen-2.5-32B & Both: Exploratory & 95.38 & \textcolor{darkgreen}{0.46 $\uparrow$} & 0.07 & 0.08 & 9 & 32 & 31 \\
Qwen-2.5-32B & Qwen-2.5-32B & Both: Det. \& Exp. & 95.30 & 0.38 & 0.04 & 0.05 & 12 & 23 & 21 \\
\midrule

Llama-3.1-3B & Llama-3.1-3B & Both: Default & 74.91 & \textcolor{darkgreen}{2.36 $\uparrow$} & 0.73 & 0.49 & 106 & 208 & 183 \\
Llama-3.1-3B & Llama-3.1-3B & Both: Deterministic & 74.37 & \textcolor{darkgreen}{1.82 $\uparrow$} & 0 & 0.00 & 0 & 0 & 0 \\
Llama-3.1-3B & Llama-3.1-3B & Both: Exploratory & 72.40 & \textcolor{darkred}{0.15 $\downarrow$} & 0.94 & 0.57 & 138 & 225 & 202 \\
Llama-3.1-3B & Llama-3.1-3B & Both: Det. \& Exp. & 73.84 & \textcolor{darkgreen}{1.29 $\uparrow$} & 0.8 & 0.48 & 133 & 193 & 175 \\
\midrule
Llama-3.1-8B & Llama-3.1-8B & Both: Default & 82.56 & \textcolor{darkgreen}{0.83 $\uparrow$} & 0.48 & 0.38 & 86 & 116 & 105 \\
Llama-3.1-8B & Llama-3.1-8B & Both: Deterministic & 81.50 & \textcolor{darkred}{0.23 $\downarrow$} & 0 & 0.00 & 0 & 0 & 0 \\
Llama-3.1-8B & Llama-3.1-8B & Both: Exploratory & 80.67 & \textcolor{darkred}{1.06 $\downarrow$} & 0.6 & 0.40 & 98 & 162 & 149 \\
Llama-3.1-8B & Llama-3.1-8B & Both: Det. \& Exp. & 82.18 & \textcolor{darkgreen}{0.45 $\uparrow$} & 0.56 & 0.39 & 97 & 142 & 126 \\
\midrule

Phi-mini-3.8B & Phi-mini-3.8B & Both: Default & 87.72 & \textcolor{darkgreen}{0.84 $\uparrow$} & 0.29 & 0.27 & 51 & 101 & 95 \\
Phi-mini-3.8B & Phi-mini-3.8B & Both: Deterministic & 86.73 & \textcolor{darkred}{0.15 $\downarrow$} & 0.02 & 0.00 & 0 & 2 & 1 \\
Phi-mini-3.8B & Phi-mini-3.8B & Both: Exploratory & 87.95 & \textcolor{darkgreen}{1.07 $\uparrow$} & 0.3 & 0.26 & 48 & 112 & 99 \\
Phi-mini-3.8B & Phi-mini-3.8B & Both: Det. \& Exp. & 87.34 & \textcolor{darkgreen}{0.46 $\uparrow$} & 0.33 & 0.26 & 62 & 103 & 95 \\
\midrule
Mistral-7B & Mistral-7B & Both: Default & 33.74 & \textcolor{darkgreen}{12.36 $\uparrow$} & 1.65 & 0.73 & 101 & 454 & 340 \\
Mistral-7B & Mistral-7B & Both: Deterministic & 20.02 & 1.36 & 0.04 & 0.00 & 0 & 0 & 0 \\
Mistral-7B & Mistral-7B & Both: Exploratory & 35.71 & \textcolor{darkgreen}{14.33 $\uparrow$} & 1.85 & 0.80 & 110 & 509 & 381 \\
Mistral-7B & Mistral-7B & Both: Det. \& Exp. & 33.51 & 12.13 & 1.53 & 0.68 & 97 & 433 & 334 \\

\bottomrule
\end{tabular}
\end{adjustbox}
\caption{Performance in Multi-Agent Debate Settings on the \textbf{GSM8K} Dataset. This table showcases the impact of different \textbf{Agent Settings} (controlling temperature and top\_p parameters like Default, Deterministic, Exploratory, and a combination) on the \textbf{MAD Accuracy (RCR Prompting) }of various language models. The \textbf{$\Delta$} column quantifies the \textbf{improvement (or decline) over the single base model performance}. Further metrics include average \textbf{Debate Rounds}, normalized \textbf{Sycophancy} (per 1319 data points), and transitions between correct (C) and incorrect (I) states (C$\rightarrow$I, I$\rightarrow$C), highlighting the nuanced effects of debate dynamics.}
\label{tab:debate_performance_revised}
\end{table*}

\begin{table*}[htbp]
\centering

\begin{adjustbox}{width=\textwidth,center} 
\sisetup{round-mode=places,round-precision=2} 
\begin{tabular}{@{}ccccS[table-format=2.2]S[table-format=3.2]S[table-format=3.2]S[table-format=1.2]S[table-format=4.2]cccS[table-format=3.0]@{}} 
\toprule
\textbf{Agent 1} & \textbf{Agent 2} & \textbf{Agent Settings} & \textbf{Accuracy} & \multicolumn{1}{c}{\textbf{$\Delta$ (Lower Agent)}} & \multicolumn{1}{c}{\textbf{$\Delta$ (Upper Agent)}} & \textbf{Debate} & \multicolumn{1}{c}{\textbf{Sycophancy}} & \textbf{C$\rightarrow$I} & \textbf{I$\rightarrow$C} & \textbf{Debate} \\
 & & & & & & \textbf{Rounds} & \multicolumn{1}{c}{\textbf{(Avg / 1319)}} & & & \textbf{Helped} \\
 & & & & & & \textbf{(Avg)} & & & & \textbf{(Overall)} \\
\midrule

Qwen-2.5-0.5B & Qwen-2.5-1.5B & 1: Default \& 2: Default & 62.40 & \textcolor{darkgreen}{20.4 $\uparrow$} & \textcolor{darkred}{0.37 $\downarrow$} & 1.52 & 0.96 & 168 & 434 & 387 \\
Qwen-2.5-0.5B & Qwen-2.5-1.5B & 1: Det. \& 2: Det. & 62.32 & \textcolor{darkgreen}{20.32 $\uparrow$} & \textcolor{darkred}{0.45 $\downarrow$} & 1.27 & 0.72 & 155 & 357 & 323 \\
Qwen-2.5-0.5B & Qwen-2.5-1.5B & 1: Exp. \& 2: Exp. & 58.91 & \textcolor{darkgreen}{16.91 $\uparrow$} & \textcolor{darkred}{3.86 $\downarrow$} & 1.95 & 1.03 & 175 & 531 & 448 \\
Qwen-2.5-0.5B & Qwen-2.5-1.5B & 1: Det. \& 2: Exp. & 60.88 & \textcolor{darkgreen}{18.88 $\uparrow$} & \textcolor{darkred}{1.89 $\downarrow$} & 1.54 & 0.83 & 147 & 416 & 344 \\
Qwen-2.5-0.5B & Qwen-2.5-1.5B & 1: Exp. \& 2: Det. & 61.18 & \textcolor{darkgreen}{19.18 $\uparrow$} & \textcolor{darkred}{1.59 $\downarrow$} & 1.67 & 0.87 & 164 & 474 & 425 \\
\midrule

Qwen-2.5-1.5B & Llama-3.1-3B & 1: Default \& 2: Default & 76.42 & \textcolor{darkgreen}{13.65 $\uparrow$} & \textcolor{darkgreen}{3.87 $\uparrow$} & 1.09 & 0.56 & 107 & 388 & 342 \\
Qwen-2.5-1.5B & Llama-3.1-3B & 1: Det. \& 2: Det. & 75.59 & \textcolor{darkgreen}{12.82 $\uparrow$} & \textcolor{darkgreen}{3.04 $\uparrow$} & 1.14 & 0.36 & 93 & 285 & 258 \\
Qwen-2.5-1.5B & Llama-3.1-3B & 1: Exp. \& 2: Exp. & 76.57 & \textcolor{darkgreen}{13.8 $\uparrow$} & \textcolor{darkgreen}{4.02 $\uparrow$} & 1.17 & 0.65 & 96 & 416 & 355 \\
Qwen-2.5-1.5B & Llama-3.1-3B & 1: Det. \& 2: Exp. & 75.06 & \textcolor{darkgreen}{12.29 $\uparrow$} & \textcolor{darkgreen}{2.51 $\uparrow$} & 1.22 & 0.48 & 111 & 362 & 326 \\
Qwen-2.5-1.5B & Llama-3.1-3B & 1: Exp. \& 2: Det. & 76.04 & \textcolor{darkgreen}{13.27 $\uparrow$} & \textcolor{darkgreen}{3.49 $\uparrow$} & 1.12 & 0.59 & 129 & 383 & 331 \\
\midrule

Qwen-2.5-3B & Phi-mini-3.8B & 1: Default \& 2: Default & 87.41 & \textcolor{darkgreen}{2.27 $\uparrow$} & \textcolor{darkgreen}{0.53 $\uparrow$} & 0.39 & 0.22 & 53 & 128 & 114 \\
Qwen-2.5-3B & Phi-mini-3.8B & 1: Det. \& 2: Det. & 85.97 & \textcolor{darkgreen}{0.83 $\uparrow$} & \textcolor{darkred}{0.91 $\downarrow$} & 0.43 & 0.17 & 74 & 82 & 72 \\
Qwen-2.5-3B & Phi-mini-3.8B & 1: Exp. \& 2: Exp. & 88.63 & \textcolor{darkgreen}{3.49 $\uparrow$} & \textcolor{darkgreen}{1.75 $\uparrow$} & 0.44 & 0.27 & 46 & 155 & 142 \\
Qwen-2.5-3B & Phi-mini-3.8B & 1: Det. \& 2: Exp. & 86.73 & \textcolor{darkgreen}{1.59 $\uparrow$} & \textcolor{darkred}{0.15 $\downarrow$} & 0.40 & 0.20 & 63 & 105 & 99 \\
Qwen-2.5-3B & Phi-mini-3.8B & 1: Exp. \& 2: Det. & 88.10 & \textcolor{darkgreen}{2.96 $\uparrow$} & \textcolor{darkgreen}{1.22 $\uparrow$} & 0.41 & 0.23 & 57 & 135 & 126 \\
\midrule

Qwen-2.5-1.5B & Qwen-2.5-3B & 1: Default \& 2: Default & 82.71 & \textcolor{darkgreen}{19.94 $\uparrow$} & \textcolor{darkred}{2.43 $\downarrow$} & 0.71 & 0.51 & 67 & 370 & 359 \\
Qwen-2.5-1.5B & Qwen-2.5-3B & 1: Det. \& 2: Det. & 81.27 & \textcolor{darkgreen}{18.5 $\uparrow$} & \textcolor{darkred}{3.87 $\downarrow$} & 0.62 & 0.48 & 94 & 284 & 275 \\
Qwen-2.5-1.5B & Qwen-2.5-3B & 1: Exp. \& 2: Exp. & 83.17 & \textcolor{darkgreen}{20.4 $\uparrow$} & \textcolor{darkred}{1.97 $\downarrow$} & 0.80 & 0.56 & 68 & 414 & 392 \\
Qwen-2.5-1.5B & Qwen-2.5-3B & 1: Det. \& 2: Exp. & 82.87 & \textcolor{darkgreen}{20.1 $\uparrow$} & \textcolor{darkred}{2.27 $\downarrow$} & 0.76 & 0.48 & 74 & 328 & 310 \\
Qwen-2.5-1.5B & Qwen-2.5-3B & 1: Exp. \& 2: Det. & 82.26 & \textcolor{darkgreen}{19.49 $\uparrow$} & \textcolor{darkred}{2.88 $\downarrow$} & 0.75 & 0.52 & 82 & 384 & 372 \\
\midrule

Llama-3.1-3B & Llama-3.1-8B & 1: Default \& 2: Default & 78.54 & \textcolor{darkgreen}{5.99 $\uparrow$} & \textcolor{darkred}{3.19 $\downarrow$} & 0.77 & 0.51 & 122 & 213 & 195 \\
Llama-3.1-3B & Llama-3.1-8B & 1: Det. \& 2: Det. & 79.23 & \textcolor{darkgreen}{6.68 $\uparrow$} & \textcolor{darkred}{2.5 $\downarrow$} & 0.68 & 0.48 & 130 & 159 & 143 \\
Llama-3.1-3B & Llama-3.1-8B & 1: Exp. \& 2: Exp. & 77.10 & \textcolor{darkgreen}{4.55 $\uparrow$} & \textcolor{darkred}{4.63 $\downarrow$} & 0.93 & 0.58 & 127 & 238 & 224 \\
Llama-3.1-3B & Llama-3.1-8B & 1: Det. \& 2: Exp. & 79.83 & \textcolor{darkgreen}{7.28 $\uparrow$} & \textcolor{darkred}{1.9 $\downarrow$} & 0.81 & 0.45 & 123 & 211 & 183 \\
Llama-3.1-3B & Llama-3.1-8B & 1: Exp. \& 2: Det. & 77.18 & \textcolor{darkgreen}{4.63 $\uparrow$} & \textcolor{darkred}{4.55 $\downarrow$} & 0.87 & 0.56 & 141 & 183 & 173 \\
\midrule

Qwen-2.5-7B & Qwen-2.5-14B & 1: Default \& 2: Default & 92.19 & \textcolor{darkgreen}{1.52 $\uparrow$} & \textcolor{darkred}{0.61 $\downarrow$} & 0.16 & 0.13 & 39 & 63 & 61 \\
Qwen-2.5-7B & Qwen-2.5-14B & 1: Det. \& 2: Det. & 92.04 & \textcolor{darkgreen}{1.37 $\uparrow$} & \textcolor{darkred}{0.76 $\downarrow$} & 0.17 & 0.13 & 47 & 53 & 50 \\
Qwen-2.5-7B & Qwen-2.5-14B & 1: Exp. \& 2: Exp. & 93.10 & \textcolor{darkgreen}{2.43 $\uparrow$} & \textcolor{darkgreen}{0.3 $\uparrow$} & 0.16 & 0.15 & 33 & 72 & 68 \\
Qwen-2.5-7B & Qwen-2.5-14B & 1: Det. \& 2: Exp. & 92.19 & \textcolor{darkgreen}{1.52 $\uparrow$} & \textcolor{darkred}{0.61 $\downarrow$} & 0.15 & 0.11 & 37 & 58 & 58 \\
Qwen-2.5-7B & Qwen-2.5-14B & 1: Exp. \& 2: Det. & 92.80 & \textcolor{darkgreen}{2.13 $\uparrow$} & 0 & 0.17 & 0.16 & 39 & 64 & 60 \\

\bottomrule
\end{tabular}
\end{adjustbox}
\caption{Performance Analysis of Cross-Agent Debates on the \textbf{GSM8K} Dataset. This table details the outcomes of debates between different language models (Agent 1 and Agent 2). \textbf{Agent Settings} specify the configuration (e.g., Default, Deterministic (Det.), Exploratory (Exp.)) applied to Agent 1 and Agent 2 respectively, influencing temperature and top\_p parameters. The table presents overall \textbf{Accuracy}, along with \textbf{$\Delta$ (Lower Agent)} and \textbf{$\Delta$ (Upper Agent)} indicating the performance change for each agent relative to a baseline. Additional metrics include average \textbf{Debate Rounds}, normalized \textbf{Sycophancy} (per 1319 data points), and transitions between correct (C) and incorrect (I) states (C$\rightarrow$I, I$\rightarrow$C) to show debate impact.}
\label{tab:cross_agent_debate_performance}
\end{table*}

\begin{table*}[htbp]
\centering
\begin{adjustbox}{width=\textwidth,center} 
\sisetup{round-mode=places,round-precision=2} 
\begin{tabular}{@{}ccccS[table-format=2.2]S[table-format=3.2]S[table-format=1.2]S[table-format=3.2]S[table-format=3.0]S[table-format=3.0]S[table-format=3.0]@{}}
\toprule
\textbf{Agent 1} & \textbf{Agent 2} & \textbf{Agent 3} & \textbf{Agent Settings} & {\textbf{Accuracy}} & {\textbf{$\Delta$ (Improvement)}} & {\textbf{Debate}} & {\textbf{Sycophancy}} & {\textbf{C$\rightarrow$I}} & {\textbf{I$\rightarrow$C}} & {\textbf{Debate}} \\
 & & & & & & {\textbf{Rounds}} & {\textbf{(Avg / 1319)}} & & & {\textbf{Helped}} \\
 & & & & & & {\textbf{(Avg)}} & & & & {\textbf{(Overall)}} \\
\midrule
Qwen-2.5-0.5B & Qwen-2.5-0.5B & Qwen-2.5-0.5B & All: Default & 41.70 & \textcolor{darkred}{0.3 $\downarrow$} & 2.77 & 3.17 & 414 & 393 & 236 \\
Qwen-2.5-0.5B & Qwen-2.5-0.5B & Qwen-2.5-0.5B & All: Deterministic & 47.31 & \textcolor{darkgreen}{5.31 $\uparrow$} & 0.00 & 0.00 & 0 & 0 & 0 \\
Qwen-2.5-0.5B & Qwen-2.5-0.5B & Qwen-2.5-0.5B & All: Exploratory & 36.09 & \textcolor{darkred}{5.91 $\downarrow$} & 3.47 & 3.33 & 438 & 450 & 282 \\
Qwen-2.5-0.5B & Qwen-2.5-0.5B & Qwen-2.5-0.5B & 1 Det, 2 Exp & 38.36 & \textcolor{darkred}{3.64 $\downarrow$} & 3.13 & 2.90 & 412 & 370 & 246 \\
Qwen-2.5-0.5B & Qwen-2.5-0.5B & Qwen-2.5-0.5B & 2 Det, 1 Exp & 43.06 & \textcolor{darkgreen}{1.06 $\uparrow$} & 1.97 & 1.42 & 306 & 300 & 211 \\
\midrule
Qwen-2.5-1.5B & Qwen-2.5-1.5B & Qwen-2.5-1.5B & All: Default & 72.48 & \textcolor{darkgreen}{9.71 $\uparrow$} & 1.35 & 1.64 & 193 & 652 & 469 \\
Qwen-2.5-1.5B & Qwen-2.5-1.5B & Qwen-2.5-1.5B & All: Deterministic & 63.99 & \textcolor{darkgreen}{1.22 $\uparrow$} & 0.00 & 0.00 & 0 & 0 & 0 \\
Qwen-2.5-1.5B & Qwen-2.5-1.5B & Qwen-2.5-1.5B & All: Exploratory & 75.13 & \textcolor{darkgreen}{12.36 $\uparrow$} & 1.57 & 1.82 & 181 & 796 & 547 \\
Qwen-2.5-1.5B & Qwen-2.5-1.5B & Qwen-2.5-1.5B & 1 Det, 2 Exp & 74.83 & \textcolor{darkgreen}{12.06 $\uparrow$} & 1.51 & 1.71 & 170 & 741 & 534 \\
Qwen-2.5-1.5B & Qwen-2.5-1.5B & Qwen-2.5-1.5B & 2 Det, 1 Exp & 72.25 & \textcolor{darkgreen}{9.48 $\uparrow$} & 0.97 & 1.03 & 131 & 510 & 329 \\
\midrule
Qwen-2.5-3B & Qwen-2.5-3B & Qwen-2.5-3B & All: Default & 86.96 & \textcolor{darkgreen}{1.82 $\uparrow$} & 0.49 & 0.52 & 85 & 191 & 147 \\
Qwen-2.5-3B & Qwen-2.5-3B & Qwen-2.5-3B & All: Deterministic & 84.99 & \textcolor{darkred}{0.15 $\downarrow$} & 0.00 & 0.00 & 0 & 0 & 0 \\
Qwen-2.5-3B & Qwen-2.5-3B & Qwen-2.5-3B & All: Exploratory & 87.64 & \textcolor{darkgreen}{2.5 $\uparrow$} & 0.60 & 0.65 & 85 & 256 & 200 \\
Qwen-2.5-3B & Qwen-2.5-3B & Qwen-2.5-3B & 1 Det, 2 Exp & 86.73 & \textcolor{darkgreen}{1.59 $\uparrow$} & 0.63 & 0.56 & 110 & 236 & 179 \\
Qwen-2.5-3B & Qwen-2.5-3B & Qwen-2.5-3B & 2 Det, 1 Exp & 86.05 & \textcolor{darkgreen}{0.91 $\uparrow$} & 0.40 & 0.32 & 75 & 130 & 99 \\
\midrule
Qwen-2.5-7B & Qwen-2.5-7B & Qwen-2.5-7B & All: Default & 93.03 & \textcolor{darkgreen}{2.36 $\uparrow$} & 0.22 & 0.22 & 33 & 110 & 88 \\
Qwen-2.5-7B & Qwen-2.5-7B & Qwen-2.5-7B & All: Deterministic & 90.60 & \textcolor{darkred}{0.07 $\downarrow$} & 0.00 & 0.00 & 0 & 0 & 0 \\
Qwen-2.5-7B & Qwen-2.5-7B & Qwen-2.5-7B & All: Exploratory & 92.42 & \textcolor{darkgreen}{1.75 $\uparrow$} & 0.24 & 0.24 & 52 & 110 & 87 \\
Qwen-2.5-7B & Qwen-2.5-7B & Qwen-2.5-7B & 1 Det, 2 Exp & 92.12 & \textcolor{darkgreen}{1.45 $\uparrow$} & 0.24 & 0.24 & 44 & 106 & 86 \\
Qwen-2.5-7B & Qwen-2.5-7B & Qwen-2.5-7B & 2 Det, 1 Exp & 91.96 & \textcolor{darkgreen}{1.29 $\uparrow$} & 0.17 & 0.17 & 28 & 76 & 52 \\
\midrule
Qwen-2.5-14B & Qwen-2.5-14B & Qwen-2.5-14B & All: Default & 94.09 & 1.29 & 0.11 & 0.13 & 18 & 67 & 59 \\
Qwen-2.5-14B & Qwen-2.5-14B & Qwen-2.5-14B & All: Deterministic & 92.95 & 0.15 & 0.00 & 0.00 & 0 & 0 & 0 \\
Qwen-2.5-14B & Qwen-2.5-14B & Qwen-2.5-14B & All: Exploratory & 94.24 & 1.44 & 0.14 & 0.16 & 26 & 88 & 78 \\
Qwen-2.5-14B & Qwen-2.5-14B & Qwen-2.5-14B & 1 Det, 2 Exp & 94.31 & 1.51 & 0.13 & 0.16 & 17 & 81 & 68 \\
Qwen-2.5-14B & Qwen-2.5-14B & Qwen-2.5-14B & 2 Det, 1 Exp & 92.87 & 0.07 & 0.09 & 0.08 & 30 & 33 & 29 \\
\midrule
Qwen-2.5-32B & Qwen-2.5-32B & Qwen-2.5-32B & All: Default & 95.30 & 0.38 & 0.07 & 0.07 & 18 & 44 & 39 \\
Qwen-2.5-32B & Qwen-2.5-32B & Qwen-2.5-32B & All: Deterministic & 94.77 & 0.15 & 0.00 & 0.00 & 0 & 0 & 0 \\
Qwen-2.5-32B & Qwen-2.5-32B & Qwen-2.5-32B & All: Exploratory & 94.84 & 0.08 & 0.08 & 0.09 & 21 & 51 & 47 \\
Qwen-2.5-32B & Qwen-2.5-32B & Qwen-2.5-32B & 1 Det, 2 Exp & 95.30 & 0.38 & 0.07 & 0.07 & 16 & 49 & 41 \\
Qwen-2.5-32B & Qwen-2.5-32B & Qwen-2.5-32B & 2 Det, 1 Exp & 95.22 & 0.30 & 0.05 & 0.05 & 11 & 34 & 24 \\
\midrule
Phi-mini-3.8B & Phi-mini-3.8B & Phi-mini-3.8B & All: Default & 88.40 & \textcolor{darkgreen}{1.52 $\uparrow$} & 0.42 & 0.55 & 86 & 168 & 129 \\
Phi-mini-3.8B & Phi-mini-3.8B & Phi-mini-3.8B & All: Deterministic & 86.66 & \textcolor{darkred}{0.22 $\downarrow$} & 0.01 & 0.01 & 0 & 0 & 0 \\
Phi-mini-3.8B & Phi-mini-3.8B & Phi-mini-3.8B & All: Exploratory & 88.10 & \textcolor{darkgreen}{1.22 $\uparrow$} & 0.48 & 0.59 & 99 & 197 & 145 \\
Phi-mini-3.8B & Phi-mini-3.8B & Phi-mini-3.8B & 1 Det, 2 Exp & 87.87 & \textcolor{darkgreen}{0.99 $\uparrow$} & 0.46 & 0.53 & 95 & 178 & 132 \\
Phi-mini-3.8B & Phi-mini-3.8B & Phi-mini-3.8B & 2 Det, 1 Exp & 87.72 & \textcolor{darkgreen}{0.84 $\uparrow$} & 0.32 & 0.41 & 64 & 121 & 80 \\
\midrule
Llama-3.1-3B & Llama-3.1-3B & Llama-3.1-3B & All: Default & 72.63 & \textcolor{darkgreen}{0.08 $\uparrow$} & 1.29 & 1.29 & 265 & 317 & 238 \\
Llama-3.1-3B & Llama-3.1-3B & Llama-3.1-3B & All: Deterministic & 73.16 & \textcolor{darkgreen}{0.61 $\uparrow$} & 0.00 & 0.00 & 0 & 0 & 0 \\
Llama-3.1-3B & Llama-3.1-3B & Llama-3.1-3B & All: Exploratory & 72.78 & \textcolor{darkgreen}{0.23 $\uparrow$} & 1.49 & 1.39 & 246 & 414 & 312 \\
Llama-3.1-3B & Llama-3.1-3B & Llama-3.1-3B & 1 Det, 2 Exp & 73.69 & \textcolor{darkgreen}{1.14 $\uparrow$} & 1.39 & 1.28 & 251 & 407 & 283 \\
Llama-3.1-3B & Llama-3.1-3B & Llama-3.1-3B & 2 Det, 1 Exp & 72.93 & \textcolor{darkgreen}{0.38 $\uparrow$} & 1.08 & 0.87 & 203 & 229 & 147 \\
\midrule
Mistral-7B & Mistral-7B & Mistral-7B & All: Default & 37.83 & \textcolor{darkgreen}{16.45 $\uparrow$} & 2.37 & 1.97 & 203 & 894 & 454 \\
Mistral-7B & Mistral-7B & Mistral-7B & All: Deterministic & 20.02 & \textcolor{darkred}{1.36 $\downarrow$} & 0.04 & 0.00 & 0 & 0 & 0 \\
Mistral-7B & Mistral-7B & Mistral-7B & All: Exploratory & 39.27 & \textcolor{darkgreen}{17.89 $\uparrow$} & 2.81 & 2.30 & 189 & 904 & 480 \\
Mistral-7B & Mistral-7B & Mistral-7B & 1 Det, 2 Exp & 38.89 & \textcolor{darkgreen}{17.51 $\uparrow$} & 2.61 & 2.13 & 222 & 940 & 476 \\
Mistral-7B & Mistral-7B & Mistral-7B & 2 Det, 1 Exp & 35.33 & \textcolor{darkgreen}{13.95 $\uparrow$} & 1.82 & 1.39 & 135 & 694 & 360 \\
\midrule
Llama-3.1-8B & Llama-3.1-8B & Llama-3.1-8B & All: Default & 84.23 & \textcolor{darkgreen}{2.5 $\uparrow$} & 0.72 & 0.82 & 135 & 429 & 192 \\
Llama-3.1-8B & Llama-3.1-8B & Llama-3.1-8B & All: Deterministic & 81.50 & \textcolor{darkred}{0.23 $\downarrow$} & 0.00 & 0.00 & 0 & 0 & 0 \\
Llama-3.1-8B & Llama-3.1-8B & Llama-3.1-8B & All: Exploratory & 83.70 & \textcolor{darkgreen}{1.97 $\uparrow$} & 0.88 & 0.89 & 162 & 310 & 230 \\
Llama-3.1-8B & Llama-3.1-8B & Llama-3.1-8B & 1 Det, 2 Exp & 83.32 & \textcolor{darkgreen}{1.59 $\uparrow$} & 0.86 & 0.86 & 160 & 284 & 211 \\
Llama-3.1-8B & Llama-3.1-8B & Llama-3.1-8B & 2 Det, 1 Exp & 82.26 & \textcolor{darkgreen}{0.53 $\uparrow$} & 0.67 & 0.63 & 129 & 199 & 132 \\
\bottomrule
\end{tabular}
\end{adjustbox}
\caption{Performance Analysis of Three Identical Agents Debating on GSM8K. This table shows results when three instances of the same model (\textbf{Agent 1}, \textbf{Agent 2}, \textbf{Agent 3} being identical) engage in a debate. \textbf{Agent Settings} describe the configuration mix across these three agents (e.g., All Default, or a mix like 1 Deterministic (Det), 2 Exploratory (Exp)). \textbf{Accuracy} is the debate outcome, and \textbf{$\Delta$ (Improvement)} is the change from the single agent's baseline. Standard metrics like \textbf{Debate Rounds}, normalized \textbf{Sycophancy} (per 1319 data points), and error transition rates (C$\rightarrow$I, I$\rightarrow$C) are also included.}
\label{tab:three_identical_agents_debate_gsm8k}
\end{table*}

\begin{table*}[htbp]
\centering

\begin{adjustbox}{width=\textwidth,center} 
\sisetup{round-mode=places,round-precision=2} 
\begin{tabular}{@{}llllS[table-format=2.2]cS[table-format=1.2]S[table-format=1.2]S[table-format=3.0]S[table-format=3.0]S[table-format=3.0]@{}}
\toprule
\textbf{Agent 1} & \textbf{Agent 2} & \textbf{Agent 3} & \textbf{Agent Settings} & {\textbf{Accuracy}} & \textbf{$\Delta$ (vs Lowest)} & {\textbf{Debate}} & {\textbf{Sycophancy}} & {\textbf{C$\rightarrow$I}} & {\textbf{I$\rightarrow$C}} & {\textbf{Debate}} \\
 & & & & & & {\textbf{Rounds}} & {\textbf{(Avg / 1319)}} & & & {\textbf{Helped}} \\
 & & & & & & {\textbf{(Avg)}} & & & & {\textbf{(Overall)}} \\
\midrule
Qwen-2.5-0.5B & Qwen-2.5-1.5B & Qwen-2.5-3B & All: Default & 80.82 & \textcolor{darkred}{4.32 $\downarrow$} & 1.81 & 1.58 & 154 & 859 & 639 \\
Qwen-2.5-0.5B & Qwen-2.5-1.5B & Llama-3.1-3B & All: Default & 69.52 & \textcolor{darkred}{3.03 $\downarrow$} & 2.43 & 1.76 & 271 & 718 & 508 \\
Qwen-2.5-0.5B & Qwen-2.5-1.5B & Phi-mini-3.8B & All: Default & 76.04 & \textcolor{darkred}{10.84 $\downarrow$} & 2.20 & 1.47 & 267 & 727 & 532 \\
Qwen-2.5-0.5B & Qwen-2.5-3B & Llama-3.1-3B & All: Default & 79.15 & \textcolor{darkred}{5.99 $\downarrow$} & 2.10 & 1.36 & 184 & 696 & 536 \\
Qwen-2.5-0.5B & Qwen-2.5-3B & Phi-mini-3.8B & All: Default & 83.62 & \textcolor{darkred}{3.24 $\downarrow$} & 1.82 & 1.08 & 150 & 618 & 534 \\
Qwen-2.5-0.5B & Llama-3.1-3B & Phi-mini-3.8B & All: Default & 76.57 & \textcolor{darkred}{10.31 $\downarrow$} & 2.39 & 1.16 & 255 & 515 & 402 \\
Qwen-2.5-1.5B & Qwen-2.5-3B & Llama-3.1-3B & All: Default & 82.71 & \textcolor{darkred}{2.43 $\downarrow$} & 1.24 & 1.06 & 156 & 544 & 436 \\
Qwen-2.5-1.5B & Qwen-2.5-3B & Phi-mini-3.8B & All: Default & 85.22 & \textcolor{darkred}{1.66 $\downarrow$} & 1.08 & 0.85 & 139 & 460 & 388 \\
Qwen-2.5-1.5B & Llama-3.1-3B & Phi-mini-3.8B & All: Default & 81.20 & \textcolor{darkred}{5.68 $\downarrow$} & 1.33 & 1.05 & 196 & 560 & 446 \\
Qwen-2.5-3B & Phi-mini-3.8B & Llama-3.1-3B & All: Default & 86.96 & \textcolor{darkgreen}{0.08 $\uparrow$} & 0.89 & 0.71 & 127 & 372 & 297 \\
Qwen-2.5-3B & Qwen-2.5-3B & Phi-mini-3.8B & All: Default & 87.64 & \textcolor{darkgreen}{0.76 $\uparrow$} & 0.60 & 0.55 & 97 & 227 & 175 \\
Qwen-2.5-3B & Phi-mini-3.8B & Phi-mini-3.8B & All: Default & 87.79 & \textcolor{darkgreen}{0.91 $\uparrow$} & 0.58 & 0.53 & 111 & 209 & 167 \\
Qwen-2.5-0.5B & Qwen-2.5-1.5B & Qwen-2.5-1.5B & All: Default & 68.46 & \textcolor{darkgreen}{5.69 $\uparrow$} & 2.10 & 2.09 & 221 & 795 & 570 \\
Qwen-2.5-0.5B & Qwen-2.5-0.5B & Qwen-2.5-1.5B & All: Default & 55.12 & \textcolor{darkred}{7.65 $\downarrow$} & 2.60 & 2.52 & 364 & 628 & 407 \\
\bottomrule
\end{tabular}
\end{adjustbox}
\caption{Performance Analysis of Three-Agent Debates (Varied Models) on \textbf{GSM8K}. This table presents outcomes from debates involving three potentially different language models (\textbf{Agent 1}, \textbf{Agent 2}, \textbf{Agent 3}). All debates use default agent settings. The \textbf{$\Delta$ (vs Lowest)} column indicates the performance change of the debate outcome (Accuracy) compared to the baseline performance of the lowest-performing agent among the three in that specific debate. Standard metrics like \textbf{Debate Rounds}, normalized \textbf{Sycophancy} (per 1319 data points), and error transition rates (C$\rightarrow$I, I$\rightarrow$C) are also included.}
\label{tab:three_varied_agents_debate_gsm8k_lowest_delta}
\end{table*}


\begin{table*}[htbp]
\centering
\begin{adjustbox}{width=\textwidth,center} 
\sisetup{round-mode=places,round-precision=2} 
\begin{tabular}{@{}ccccS[table-format=2.2]S[table-format=3.2]S[table-format=1.2]S[table-format=4.2]cccS[table-format=1.2]@{}} 
\toprule
\textbf{Agent 1} & \textbf{Agent 2} & \textbf{Agent Settings} & \textbf{MAD Accuracy} & \multicolumn{1}{c}{\textbf{$\Delta$}} & \textbf{Debate} & \multicolumn{1}{c}{\textbf{Sycophancy}} & \textbf{C$\rightarrow$I} & \textbf{I$\rightarrow$C} & \textbf{Debate} \\
& & & \textbf{(RCR Prompting)} & & \textbf{Rounds} & \multicolumn{1}{c}{\textbf{(Avg / 2400)}} & & & \textbf{Helped} \\
& & & & & \textbf{(Avg)} & & & & \textbf{(Overall)} \\
\midrule
Qwen-2.5-0.5B & Qwen-2.5-0.5B & Both: Default & 27.33 & \textcolor{darkgreen}{2.54 $\uparrow$} & 2.00 & 1.51 & 248 & 348 & 295 \\
Qwen-2.5-0.5B & Qwen-2.5-0.5B & Both: Deterministic & 29.25 & \textcolor{darkgreen}{4.46 $\uparrow$} & 0.02 & 0.00 & 0 & 2 & 1 \\
Qwen-2.5-0.5B & Qwen-2.5-0.5B & Both: Exploratory & 23.12 & \textcolor{darkred}{1.67 $\downarrow$} & 2.56 & 1.43 & 284 & 351 & 289 \\
Qwen-2.5-0.5B & Qwen-2.5-0.5B & Both: Det. \& Exp. & 27.33 & \textcolor{darkgreen}{2.54 $\uparrow$} & 2.26 & 1.33 & 267 & 396 & 336 \\
\midrule
Qwen-2.5-1.5B & Qwen-2.5-1.5B & Both: Default & 53.12 & \textcolor{darkgreen}{11.12 $\uparrow$} & 1.14 & 0.91 & 210 & 555 & 502 \\
Qwen-2.5-1.5B & Qwen-2.5-1.5B & Both: Deterministic & 47.29 & \textcolor{darkgreen}{5.29 $\uparrow$} & 0.03 & 0.00 & 0 & 0 & 0 \\
Qwen-2.5-1.5B & Qwen-2.5-1.5B & Both: Exploratory & 51.62 & \textcolor{darkgreen}{9.62 $\uparrow$} & 1.40 & 1.08 & 218 & 647 & 551 \\
Qwen-2.5-1.5B & Qwen-2.5-1.5B & Both: Det. \& Exp. & 52.29 & \textcolor{darkgreen}{10.29 $\uparrow$} & 1.17 & 0.85 & 181 & 528 & 477 \\
\midrule
Qwen-2.5-3B & Qwen-2.5-3B & Both: Default & 67.42 & \textcolor{darkgreen}{5.67 $\uparrow$} & 0.62 & 0.39 & 133 & 225 & 213 \\
Qwen-2.5-3B & Qwen-2.5-3B & Both: Deterministic & 67.38 & \textcolor{darkgreen}{5.63 $\uparrow$} & 0.05 & 0.00 & 0 & 0 & 0 \\
Qwen-2.5-3B & Qwen-2.5-3B & Both: Exploratory & 67.79 & \textcolor{darkgreen}{6.04 $\uparrow$} & 0.69 & 0.46 & 132 & 296 & 265 \\
Qwen-2.5-3B & Qwen-2.5-3B & Both: Det. \& Exp. & 66.46 & \textcolor{darkgreen}{4.71 $\uparrow$} & 0.67 & 0.36 & 163 & 223 & 208 \\
\midrule
Qwen-2.5-7B & Qwen-2.5-7B & Both: Default & 74.17 & \textcolor{darkgreen}{5.55 $\uparrow$} & 0.35 & 0.26 & 62 & 135 & 127 \\
Qwen-2.5-7B & Qwen-2.5-7B & Both: Deterministic & 73.62 & \textcolor{darkgreen}{5.00 $\uparrow$} & 0.04 & 0.00 & 0 & 0 & 0 \\
Qwen-2.5-7B & Qwen-2.5-7B & Both: Exploratory & 74.17 & \textcolor{darkgreen}{5.55 $\uparrow$} & 0.39 & 0.30 & 88 & 158 & 150 \\
Qwen-2.5-7B & Qwen-2.5-7B & Both: Det. \& Exp. & 74.46 & \textcolor{darkgreen}{5.84 $\uparrow$} & 0.33 & 0.25 & 78 & 126 & 118 \\
\midrule
Qwen-2.5-14B & Qwen-2.5-14B & Both: Default & 77.21 & \textcolor{darkgreen}{5.42 $\uparrow$} & 0.32 & 0.32 & 47 & 102 & 100 \\
Qwen-2.5-14B & Qwen-2.5-14B & Both: Deterministic & 76.25 & \textcolor{darkgreen}{4.46 $\uparrow$} & 0.06 & 0.00 & 0 & 0 & 0 \\
Qwen-2.5-14B & Qwen-2.5-14B & Both: Exploratory & 77.25 & \textcolor{darkgreen}{5.46 $\uparrow$} & 0.33 & 0.32 & 45 & 128 & 123 \\
Qwen-2.5-14B & Qwen-2.5-14B & Both: Det. \& Exp. & 76.96 & \textcolor{darkgreen}{5.17 $\uparrow$} & 0.31 & 0.29 & 48 & 99 & 93 \\
\midrule
Qwen-2.5-32B & Qwen-2.5-32B & Both: Default & 73.33 & \textcolor{darkgreen}{0.87 $\uparrow$} & 0.24 & 0.19 & 29 & 62 & 59 \\
Qwen-2.5-32B & Qwen-2.5-32B & Both: Deterministic & 72.79 & \textcolor{darkgreen}{0.33 $\uparrow$} & 0.08 & 0.00 & 0 & 0 & 0 \\
Qwen-2.5-32B & Qwen-2.5-32B & Both: Exploratory & 73.42 & \textcolor{darkgreen}{0.96 $\uparrow$} & 0.27 & 0.23 & 32 & 91 & 88 \\
Qwen-2.5-32B & Qwen-2.5-32B & Both: Det. \& Exp. & 73.46 & \textcolor{darkgreen}{1.00 $\uparrow$} & 0.26 & 0.19 & 26 & 70 & 68 \\
\midrule
Phi-mini-3.8B & Phi-mini-3.8B & Both: Default & 69.62 & \textcolor{darkgreen}{6.20 $\uparrow$} & 0.60 & 0.47 & 113 & 204 & 191 \\
Phi-mini-3.8B & Phi-mini-3.8B & Both: Deterministic & 69.21 & \textcolor{darkgreen}{5.79 $\uparrow$} & 0.13 & 0.02 & 0 & 6 & 3 \\
Phi-mini-3.8B & Phi-mini-3.8B & Both: Exploratory & 70.38 & \textcolor{darkgreen}{6.96 $\uparrow$} & 0.67 & 0.50 & 117 & 267 & 242 \\
Phi-mini-3.8B & Phi-mini-3.8B & Both: Det. \& Exp. & 69.42 & \textcolor{darkgreen}{6.00 $\uparrow$} & 0.62 & 0.45 & 114 & 203 & 188 \\
\midrule
Mistral-7B & Mistral-7B & Both: Default & 23.42 & \textcolor{darkgreen}{8.38 $\uparrow$} & 1.91 & 0.77 & 159 & 576 & 434 \\
Mistral-7B & Mistral-7B & Both: Deterministic & 14.33 & \textcolor{darkred}{0.71 $\downarrow$} & 0.15 & 0.01 & 0 & 4 & 2 \\
Mistral-7B & Mistral-7B & Both: Exploratory & 23.29 & \textcolor{darkgreen}{8.25 $\uparrow$} & 2.13 & 0.85 & 149 & 586 & 437 \\
Mistral-7B & Mistral-7B & Both: Det. \& Exp. & 22.75 & \textcolor{darkgreen}{7.71 $\uparrow$} & 1.93 & 0.77 & 147 & 556 & 414 \\
\midrule
Llama-3.1-3B & Llama-3.1-3B & Both: Default & 51.58 & \textcolor{darkgreen}{5.91 $\uparrow$} & 1.20 & 0.82 & 232 & 439 & 378 \\
Llama-3.1-3B & Llama-3.1-3B & Both: Deterministic & 50.50 & \textcolor{darkgreen}{4.83 $\uparrow$} & 0.01 & 0.00 & 0 & 0 & 0 \\
Llama-3.1-3B & Llama-3.1-3B & Both: Exploratory & 51.12 & \textcolor{darkgreen}{5.45 $\uparrow$} & 1.47 & 0.87 & 233 & 482 & 406 \\
Llama-3.1-3B & Llama-3.1-3B & Both: Det. \& Exp. & 50.75 & \textcolor{darkgreen}{5.08 $\uparrow$} & 1.28 & 0.74 & 218 & 381 & 333 \\
\midrule
Llama-3.1-8B & Llama-3.1-8B & Both: Default & 62.04 & \textcolor{darkgreen}{6.42 $\uparrow$} & 0.95 & 0.72 & 202 & 313 & 274 \\
Llama-3.1-8B & Llama-3.1-8B & Both: Deterministic & 61.04 & \textcolor{darkgreen}{5.42 $\uparrow$} & 0.00 & 0.00 & 0 & 0 & 0 \\
Llama-3.1-8B & Llama-3.1-8B & Both: Exploratory & 60.79 & \textcolor{darkgreen}{5.17 $\uparrow$} & 1.12 & 0.77 & 197 & 340 & 303 \\
Llama-3.1-8B & Llama-3.1-8B & Both: Det. \& Exp. & 60.96 & \textcolor{darkgreen}{5.34 $\uparrow$} & 1.01 & 0.72 & 214 & 304 & 273 \\
\bottomrule
\end{tabular}
\end{adjustbox}
\caption{Comparative Analysis of Language Model Performance in Multi-Agent Debate Settings on the \textbf{GSM-Plus} Dataset. This table showcases the impact of different \textbf{Agent Settings} (controlling temperature and top\_p parameters like Default, Deterministic, Exploratory, and a combination) on the \textbf{MAD Accuracy (RCR Prompting)} of various language models. The \textbf{$\Delta$} column quantifies the \textbf{improvement (or decline) over the single base model performance}. Further metrics include average \textbf{Debate Rounds}, normalized \textbf{Sycophancy} (per 2400 data points), and transitions between correct (C) and incorrect (I) states (C$\rightarrow$I, I$\rightarrow$C), highlighting the nuanced effects of debate dynamics.}
\label{tab:debate_performance_gsm_plus1}
\end{table*}

\begin{table*}[htbp]
\centering
\begin{adjustbox}{width=\textwidth,center} 
\sisetup{round-mode=places,round-precision=2} 
\begin{tabular}{@{}cccS[table-format=2.2]S[table-format=2.2]S[table-format=2.2]S[table-format=2.2]S[table-format=1.2]cccS[table-format=1.2]@{}} 
\toprule
\textbf{Agent 1} & \textbf{Agent 2} & \textbf{Agent Settings} & \multicolumn{1}{c}{\textbf{MAD}} & \multicolumn{1}{c}{\textbf{$\Delta$ Lower}} & \multicolumn{1}{c}{\textbf{$\Delta$ Upper}} & \textbf{Debate} & \multicolumn{1}{c}{\textbf{Sycophancy}} & \textbf{C$\rightarrow$I} & \textbf{I$\rightarrow$C} & \textbf{Debate} \\
& & & \multicolumn{1}{c}{\textbf{Accuracy}} & & & \textbf{Rounds} & \multicolumn{1}{c}{\textbf{(Avg / 2400)}} & & & \textbf{Helped} \\
& & & & & & \textbf{(Avg)} & & & & \textbf{(Overall)} \\
\midrule
Qwen-2.5-0.5B & Qwen-2.5-1.5B & Both: Default & 41.38 & \textcolor{darkgreen}{16.59 $\uparrow$} & \textcolor{darkred}{0.62 $\downarrow$} & 1.85 & 1.12 & 314 & 628 & 548 \\
Qwen-2.5-0.5B & Qwen-2.5-1.5B & Both: Deterministic & 42.67 & \textcolor{darkgreen}{17.88 $\uparrow$} & \textcolor{darkgreen}{0.67 $\uparrow$} & 1.58 & 0.89 & 292 & 565 & 505 \\
Qwen-2.5-0.5B & Qwen-2.5-1.5B & Both: Exploratory & 39.54 & \textcolor{darkgreen}{14.75 $\uparrow$} & \textcolor{darkred}{2.46 $\downarrow$} & 2.30 & 1.20 & 320 & 722 & 604 \\
Qwen-2.5-0.5B & Qwen-2.5-1.5B & Both: Det. \& Exp. & 40.04 & \textcolor{darkgreen}{15.25 $\uparrow$} & \textcolor{darkred}{1.96 $\downarrow$} & 1.97 & 1.04 & 301 & 588 & 492 \\
Qwen-2.5-0.5B & Qwen-2.5-1.5B & Both: Exp. \& Det. & 44.25 & \textcolor{darkgreen}{19.46 $\uparrow$} & \textcolor{darkgreen}{2.25 $\uparrow$} & 2.00 & 1.04 & 278 & 750 & 664 \\
\midrule
Qwen-2.5-1.5B & Llama-3.1-3B & Both: Default & 54.42 & \textcolor{darkgreen}{12.42 $\uparrow$} & \textcolor{darkgreen}{8.75 $\uparrow$} & 1.56 & 0.75 & 232 & 612 & 532 \\
Qwen-2.5-1.5B & Llama-3.1-3B & Both: Deterministic & 54.37 & \textcolor{darkgreen}{12.37 $\uparrow$} & \textcolor{darkgreen}{8.70 $\uparrow$} & 1.56 & 0.50 & 224 & 489 & 435 \\
Qwen-2.5-1.5B & Llama-3.1-3B & Both: Exploratory & 54.21 & \textcolor{darkgreen}{12.21 $\uparrow$} & \textcolor{darkgreen}{8.54 $\uparrow$} & 1.77 & 0.89 & 255 & 696 & 602 \\
Qwen-2.5-1.5B & Llama-3.1-3B & Both: Det. \& Exp. & 53.29 & \textcolor{darkgreen}{11.29 $\uparrow$} & \textcolor{darkgreen}{7.62 $\uparrow$} & 1.65 & 0.62 & 249 & 555 & 488 \\
Qwen-2.5-1.5B & Llama-3.1-3B & Both: Exp. \& Det. & 54.58 & \textcolor{darkgreen}{12.58 $\uparrow$} & \textcolor{darkgreen}{8.91 $\uparrow$} & 1.51 & 0.77 & 249 & 603 & 533 \\
\midrule
Qwen-2.5-3B & Phi-mini-3.8B & Both: Default & 70.21 & \textcolor{darkgreen}{8.46 $\uparrow$} & \textcolor{darkgreen}{6.79 $\uparrow$} & 0.79 & 0.41 & 132 & 304 & 275 \\
Qwen-2.5-3B & Phi-mini-3.8B & Both: Deterministic & 69.83 & \textcolor{darkgreen}{8.08 $\uparrow$} & \textcolor{darkgreen}{6.41 $\uparrow$} & 0.78 & 0.29 & 128 & 224 & 200 \\
Qwen-2.5-3B & Phi-mini-3.8B & Both: Exploratory & 69.71 & \textcolor{darkgreen}{7.96 $\uparrow$} & \textcolor{darkgreen}{6.29 $\uparrow$} & 0.83 & 0.47 & 136 & 339 & 303 \\
Qwen-2.5-3B & Phi-mini-3.8B & Both: Det. \& Exp. & 69.88 & \textcolor{darkgreen}{8.13 $\uparrow$} & \textcolor{darkgreen}{6.46 $\uparrow$} & 0.79 & 0.31 & 133 & 241 & 216 \\
Qwen-2.5-3B & Phi-mini-3.8B & Both: Exp. \& Det. & 70.58 & \textcolor{darkgreen}{8.83 $\uparrow$} & \textcolor{darkgreen}{7.16 $\uparrow$} & 0.81 & 0.38 & 134 & 307 & 276 \\
\midrule
Qwen-2.5-1.5B & Qwen-2.5-3B & Both: Default & 63.79 & \textcolor{darkgreen}{21.79 $\uparrow$} & \textcolor{darkgreen}{2.04 $\uparrow$} & 1.05 & 0.67 & 154 & 573 & 537 \\
Qwen-2.5-1.5B & Qwen-2.5-3B & Both: Deterministic & 63.92 & \textcolor{darkgreen}{21.92 $\uparrow$} & \textcolor{darkgreen}{2.17 $\uparrow$} & 0.85 & 0.60 & 180 & 500 & 471 \\
Qwen-2.5-1.5B & Qwen-2.5-3B & Both: Exploratory & 63.79 & \textcolor{darkgreen}{21.79 $\uparrow$} & \textcolor{darkgreen}{2.04 $\uparrow$} & 1.12 & 0.76 & 165 & 680 & 639 \\
Qwen-2.5-1.5B & Qwen-2.5-3B & Both: Det. \& Exp. & 62.58 & \textcolor{darkgreen}{20.58 $\uparrow$} & \textcolor{darkgreen}{0.83 $\uparrow$} & 1.09 & 0.61 & 174 & 525 & 483 \\
Qwen-2.5-1.5B & Qwen-2.5-3B & Both: Exp. \& Det. & 64.25 & \textcolor{darkgreen}{22.25 $\uparrow$} & \textcolor{darkgreen}{2.50 $\uparrow$} & 1.08 & 0.68 & 189 & 640 & 608 \\
\midrule
Llama-3.1-3B & Llama-3.1-8B & Both: Default & 56.75 & \textcolor{darkgreen}{11.08 $\uparrow$} & \textcolor{darkgreen}{1.13 $\uparrow$} & 1.29 & 0.88 & 264 & 422 & 381 \\
Llama-3.1-3B & Llama-3.1-8B & Both: Deterministic & 57.08 & \textcolor{darkgreen}{11.41 $\uparrow$} & \textcolor{darkgreen}{1.46 $\uparrow$} & 1.13 & 0.74 & 278 & 348 & 316 \\
Llama-3.1-3B & Llama-3.1-8B & Both: Exploratory & 57.17 & \textcolor{darkgreen}{11.50 $\uparrow$} & \textcolor{darkgreen}{1.55 $\uparrow$} & 1.43 & 0.89 & 241 & 490 & 424 \\
Llama-3.1-3B & Llama-3.1-8B & Both: Det. \& Exp. & 57.21 & \textcolor{darkgreen}{11.54 $\uparrow$} & \textcolor{darkgreen}{1.59 $\uparrow$} & 1.27 & 0.72 & 259 & 420 & 362 \\
Llama-3.1-3B & Llama-3.1-8B & Both: Exp. \& Det. & 56.67 & \textcolor{darkgreen}{11.00 $\uparrow$} & \textcolor{darkgreen}{1.05 $\uparrow$} & 1.27 & 0.80 & 298 & 411 & 364 \\
\midrule
Qwen-2.5-7B & Qwen-2.5-14B & Both: Default & 75.88 & \textcolor{darkgreen}{7.26 $\uparrow$} & \textcolor{darkgreen}{4.09 $\uparrow$} & 0.38 & 0.28 & 88 & 165 & 159 \\
Qwen-2.5-7B & Qwen-2.5-14B & Both: Deterministic & 75.54 & \textcolor{darkgreen}{6.92 $\uparrow$} & \textcolor{darkgreen}{3.75 $\uparrow$} & 0.32 & 0.24 & 83 & 119 & 112 \\
Qwen-2.5-7B & Qwen-2.5-14B & Both: Exploratory & 75.08 & \textcolor{darkgreen}{6.46 $\uparrow$} & \textcolor{darkgreen}{3.29 $\uparrow$} & 0.39 & 0.30 & 111 & 168 & 153 \\
Qwen-2.5-7B & Qwen-2.5-14B & Both: Det. \& Exp. & 76.12 & \textcolor{darkgreen}{7.50 $\uparrow$} & \textcolor{darkgreen}{4.33 $\uparrow$} & 0.36 & 0.25 & 92 & 155 & 148 \\
Qwen-2.5-7B & Qwen-2.5-14B & Both: Exp. \& Det. & 76.33 & \textcolor{darkgreen}{7.71 $\uparrow$} & \textcolor{darkgreen}{4.54 $\uparrow$} & 0.35 & 0.31 & 78 & 143 & 133 \\
\bottomrule
\end{tabular}
\end{adjustbox}
\caption{Comparative Analysis of Mixed-Model Performance in Multi-Agent Debate Settings on the \textbf{GSM-Plus} Dataset. This table showcases the impact of different \textbf{Agent Settings} on the \textbf{MAD Accuracy} when pairing different language models together. The \textbf{$\Delta$ Lower} and \textbf{$\Delta$ Upper} columns quantify the improvement (or decline) over each individual model's base performance. Further metrics include average \textbf{Debate Rounds}, normalized \textbf{Sycophancy} (per 2400 data points), and transitions between correct (C) and incorrect (I) states (C$\rightarrow$I, I$\rightarrow$C), highlighting the dynamics when models of different capabilities debate together.}
\label{tab:mixed_model_debate_performance}
\end{table*}

\begin{table*}[htbp]
\centering
\begin{adjustbox}{width=\textwidth,center} 
\sisetup{round-mode=places,round-precision=2} 
\begin{tabular}{@{}ccccS[table-format=2.2]S[table-format=3.2]S[table-format=1.2]S[table-format=4.2]cccS[table-format=1.2]@{}} 
\toprule
\textbf{Agent 1} & \textbf{Agent 2} & \textbf{Agent 3} & \textbf{Agent Settings} & \multicolumn{1}{c}{\textbf{Accuracy}} & \multicolumn{1}{c}{\textbf{$\Delta$}} & \textbf{Debate} & \multicolumn{1}{c}{\textbf{Sycophancy}} & \textbf{C$\rightarrow$I} & \textbf{I$\rightarrow$C} & \textbf{Debate} \\
 & & & & & & \textbf{Rounds} & \multicolumn{1}{c}{\textbf{(Avg / 2400)}} & & & \textbf{Helped} \\
 & & & & & & \textbf{(Avg)} & & & & \textbf{(Overall)} \\
\midrule
Qwen-2.5-0.5B & Qwen-2.5-0.5B & Qwen-2.5-0.5B & Default & 25.00 & \textcolor{darkgreen}{0.21 $\uparrow$} & 3.21 & 3.75 & 583 & 473 & 299 \\
Qwen-2.5-0.5B & Qwen-2.5-0.5B & Qwen-2.5-0.5B & Deterministic & 29.21 & \textcolor{darkgreen}{4.42 $\uparrow$} & 0.02 & 0.00 & 0 & 0 & 0 \\
Qwen-2.5-0.5B & Qwen-2.5-0.5B & Qwen-2.5-0.5B & Exploratory & 20.75 & \textcolor{darkred}{4.04 $\downarrow$} & 3.88 & 3.78 & 645 & 578 & 344 \\
Qwen-2.5-0.5B & Qwen-2.5-0.5B & Qwen-2.5-0.5B & 1 Det. \& 2 Exp. & 22.67 & \textcolor{darkred}{2.12 $\downarrow$} & 3.66 & 3.40 & 667 & 467 & 296 \\
Qwen-2.5-0.5B & Qwen-2.5-0.5B & Qwen-2.5-0.5B & 2 Det. \& 1 Exp. & 25.42 & \textcolor{darkgreen}{0.63 $\uparrow$} & 2.45 & 1.96 & 454 & 394 & 279 \\
\midrule
Qwen-2.5-1.5B & Qwen-2.5-1.5B & Qwen-2.5-1.5B & Default & 53.04 & \textcolor{darkgreen}{11.04 $\uparrow$} & 1.87 & 2.28 & 446 & 995 & 676 \\
Qwen-2.5-1.5B & Qwen-2.5-1.5B & Qwen-2.5-1.5B & Deterministic & 47.29 & \textcolor{darkgreen}{5.29 $\uparrow$} & 0.03 & 0.00 & 0 & 0 & 0 \\
Qwen-2.5-1.5B & Qwen-2.5-1.5B & Qwen-2.5-1.5B & Exploratory & 53.33 & \textcolor{darkgreen}{11.33 $\uparrow$} & 2.24 & 2.74 & 357 & 1159 & 774 \\
Qwen-2.5-1.5B & Qwen-2.5-1.5B & Qwen-2.5-1.5B & 1 Det. \& 2 Exp. & 53.67 & \textcolor{darkgreen}{11.67 $\uparrow$} & 2.03 & 2.35 & 394 & 1116 & 756 \\
Qwen-2.5-1.5B & Qwen-2.5-1.5B & Qwen-2.5-1.5B & 2 Det. \& 1 Exp. & 53.17 & \textcolor{darkgreen}{11.17 $\uparrow$} & 1.31 & 1.41 & 265 & 793 & 514 \\
\midrule
Qwen-2.5-3B & Qwen-2.5-3B & Qwen-2.5-3B & Default & 67.38 & \textcolor{darkgreen}{5.63 $\uparrow$} & 0.97 & 1.01 & 273 & 423 & 326 \\
Qwen-2.5-3B & Qwen-2.5-3B & Qwen-2.5-3B & Deterministic & 67.38 & \textcolor{darkgreen}{5.63 $\uparrow$} & 0.05 & 0.00 & 0 & 0 & 0 \\
Qwen-2.5-3B & Qwen-2.5-3B & Qwen-2.5-3B & Exploratory & 68.00 & \textcolor{darkgreen}{6.25 $\uparrow$} & 1.09 & 1.12 & 223 & 537 & 404 \\
Qwen-2.5-3B & Qwen-2.5-3B & Qwen-2.5-3B & 1 Det. \& 2 Exp. & 68.54 & \textcolor{darkgreen}{6.79 $\uparrow$} & 1.08 & 0.94 & 235 & 428 & 343 \\
Qwen-2.5-3B & Qwen-2.5-3B & Qwen-2.5-3B & 2 Det. \& 1 Exp. & 67.12 & \textcolor{darkgreen}{5.37 $\uparrow$} & 0.78 & 0.61 & 202 & 274 & 208 \\
\midrule
Qwen-2.5-7B & Qwen-2.5-7B & Qwen-2.5-7B & Default & 75.79 & \textcolor{darkgreen}{7.17 $\uparrow$} & 0.51 & 0.52 & 84 & 272 & 209 \\
Qwen-2.5-7B & Qwen-2.5-7B & Qwen-2.5-7B & Deterministic & 73.62 & \textcolor{darkgreen}{5.00 $\uparrow$} & 0.04 & 0.00 & 0 & 0 & 0 \\
Qwen-2.5-7B & Qwen-2.5-7B & Qwen-2.5-7B & Exploratory & 74.96 & \textcolor{darkgreen}{6.34 $\uparrow$} & 0.55 & 0.54 & 117 & 270 & 220 \\
Qwen-2.5-7B & Qwen-2.5-7B & Qwen-2.5-7B & 1 Det. \& 2 Exp. & 75.25 & \textcolor{darkgreen}{6.63 $\uparrow$} & 0.50 & 0.50 & 120 & 267 & 214 \\
Qwen-2.5-7B & Qwen-2.5-7B & Qwen-2.5-7B & 2 Det. \& 1 Exp. & 74.42 & \textcolor{darkgreen}{5.80 $\uparrow$} & 0.39 & 0.39 & 97 & 181 & 135 \\
\midrule
Qwen-2.5-14B & Qwen-2.5-14B & Qwen-2.5-14B & Default & 77.92 & \textcolor{darkgreen}{6.13 $\uparrow$} & 0.35 & 0.35 & 55 & 166 & 140 \\
Qwen-2.5-14B & Qwen-2.5-14B & Qwen-2.5-14B & Deterministic & 76.54 & \textcolor{darkgreen}{4.75 $\uparrow$} & 0.05 & 0.00 & 0 & 3 & 1 \\
Qwen-2.5-14B & Qwen-2.5-14B & Qwen-2.5-14B & Exploratory & 77.29 & \textcolor{darkgreen}{5.50 $\uparrow$} & 0.38 & 0.40 & 69 & 188 & 159 \\
Qwen-2.5-14B & Qwen-2.5-14B & Qwen-2.5-14B & 1 Det. \& 2 Exp. & 77.21 & \textcolor{darkgreen}{5.42 $\uparrow$} & 0.38 & 0.37 & 72 & 172 & 143 \\
Qwen-2.5-14B & Qwen-2.5-14B & Qwen-2.5-14B & 2 Det. \& 1 Exp. & 77.21 & \textcolor{darkgreen}{5.42 $\uparrow$} & 0.28 & 0.25 & 48 & 105 & 81 \\
\midrule
Qwen-2.5-32B & Qwen-2.5-32B & Qwen-2.5-32B & Default & 73.46 & \textcolor{darkgreen}{1.00 $\uparrow$} & 0.29 & 0.23 & 48 & 112 & 96 \\
Qwen-2.5-32B & Qwen-2.5-32B & Qwen-2.5-32B & Deterministic & 72.79 & \textcolor{darkgreen}{0.33 $\uparrow$} & 0.08 & 0.00 & 0 & 0 & 0 \\
Qwen-2.5-32B & Qwen-2.5-32B & Qwen-2.5-32B & Exploratory & 73.46 & \textcolor{darkgreen}{1.00 $\uparrow$} & 0.33 & 0.31 & 46 & 123 & 109 \\
Qwen-2.5-32B & Qwen-2.5-32B & Qwen-2.5-32B & 1 Det. \& 2 Exp. & 73.88 & \textcolor{darkgreen}{1.42 $\uparrow$} & 0.29 & 0.23 & 42 & 131 & 106 \\
Qwen-2.5-32B & Qwen-2.5-32B & Qwen-2.5-32B & 2 Det. \& 1 Exp. & 73.12 & \textcolor{darkgreen}{0.66 $\uparrow$} & 0.24 & 0.17 & 40 & 75 & 60 \\
\midrule
Phi-mini-3.8B & Phi-mini-3.8B & Phi-mini-3.8B & Default & 70.21 & \textcolor{darkgreen}{6.79 $\uparrow$} & 0.90 & 1.12 & 226 & 389 & 284 \\
Phi-mini-3.8B & Phi-mini-3.8B & Phi-mini-3.8B & Deterministic & 69.17 & \textcolor{darkgreen}{5.75 $\uparrow$} & 0.12 & 0.04 & 0 & 3 & 1 \\
Phi-mini-3.8B & Phi-mini-3.8B & Phi-mini-3.8B & Exploratory & 70.25 & \textcolor{darkgreen}{6.83 $\uparrow$} & 0.95 & 1.11 & 219 & 423 & 327 \\
Phi-mini-3.8B & Phi-mini-3.8B & Phi-mini-3.8B & 1 Det. \& 2 Exp. & 69.83 & \textcolor{darkgreen}{6.41 $\uparrow$} & 0.93 & 1.02 & 232 & 390 & 293 \\
Phi-mini-3.8B & Phi-mini-3.8B & Phi-mini-3.8B & 2 Det. \& 1 Exp. & 69.54 & \textcolor{darkgreen}{6.12 $\uparrow$} & 0.73 & 0.81 & 191 & 292 & 202 \\
\midrule
Mistral-7B & Mistral-7B & Mistral-7B & Default & 24.04 & \textcolor{darkgreen}{8.99 $\uparrow$} & 2.75 & 2.12 & 312 & 979 & 525 \\
Mistral-7B & Mistral-7B & Mistral-7B & Deterministic & 14.37 & \textcolor{darkred}{0.67 $\downarrow$} & 0.15 & 0.02 & 0 & 8 & 3 \\
Mistral-7B & Mistral-7B & Mistral-7B & Exploratory & 27.04 & \textcolor{darkgreen}{12.00 $\uparrow$} & 3.03 & 2.49 & 325 & 1234 & 628 \\
Mistral-7B & Mistral-7B & Mistral-7B & 1 Det. \& 2 Exp. & 23.92 & \textcolor{darkgreen}{8.88 $\uparrow$} & 2.90 & 2.25 & 349 & 1046 & 544 \\
Mistral-7B & Mistral-7B & Mistral-7B & 2 Det. \& 1 Exp. & 23.00 & \textcolor{darkgreen}{7.96 $\uparrow$} & 2.16 & 1.55 & 232 & 855 & 458 \\
\midrule
Llama-3.1-3B & Llama-3.1-3B & Llama-3.1-3B & Default & 51.54 & \textcolor{darkgreen}{5.87 $\uparrow$} & 1.89 & 1.93 & 454 & 733 & 476 \\
Llama-3.1-3B & Llama-3.1-3B & Llama-3.1-3B & Deterministic & 50.67 & \textcolor{darkgreen}{5.00 $\uparrow$} & 0.01 & 0.00 & 0 & 0 & 0 \\
Llama-3.1-3B & Llama-3.1-3B & Llama-3.1-3B & Exploratory & 50.71 & \textcolor{darkgreen}{5.04 $\uparrow$} & 2.26 & 2.12 & 520 & 857 & 544 \\
Llama-3.1-3B & Llama-3.1-3B & Llama-3.1-3B & 1 Det. \& 2 Exp. & 50.17 & \textcolor{darkgreen}{4.50 $\uparrow$} & 2.12 & 1.96 & 515 & 744 & 493 \\
Llama-3.1-3B & Llama-3.1-3B & Llama-3.1-3B & 2 Det. \& 1 Exp. & 51.33 & \textcolor{darkgreen}{5.66 $\uparrow$} & 1.50 & 1.23 & 309 & 493 & 322 \\
\midrule
Llama-3.1-8B & Llama-3.1-8B & Llama-3.1-8B & Default & 62.67 & \textcolor{darkgreen}{7.05 $\uparrow$} & 1.43 & 1.60 & 345 & 572 & 407 \\
Llama-3.1-8B & Llama-3.1-8B & Llama-3.1-8B & Deterministic & 61.04 & \textcolor{darkgreen}{5.42 $\uparrow$} & 0.00 & 0.00 & 0 & 0 & 0 \\
Llama-3.1-8B & Llama-3.1-8B & Llama-3.1-8B & Exploratory & 61.08 & \textcolor{darkgreen}{5.46 $\uparrow$} & 1.69 & 1.85 & 385 & 624 & 446 \\
Llama-3.1-8B & Llama-3.1-8B & Llama-3.1-8B & 1 Det. \& 2 Exp. & 62.12 & \textcolor{darkgreen}{6.50 $\uparrow$} & 1.51 & 1.64 & 374 & 588 & 413 \\
Llama-3.1-8B & Llama-3.1-8B & Llama-3.1-8B & 2 Det. \& 1 Exp. & 61.12 & \textcolor{darkgreen}{5.50 $\uparrow$} & 1.20 & 1.20 & 335 & 414 & 269 \\
\bottomrule
\end{tabular}
\end{adjustbox}
\caption{Comparative Analysis of Language Model Performance in Multi-Agent Debate Settings on the \textbf{GSM-Plus} Dataset. This table showcases the impact of different \textbf{Agent Settings} (controlling temperature and top\_p parameters like Default, Deterministic, Exploratory, and combinations) on the \textbf{Accuracy} of various language models in three-agent configurations. The \textbf{$\Delta$} column quantifies the \textbf{improvement (or decline) over the single base model performance}. Further metrics include average \textbf{Debate Rounds}, normalized \textbf{Sycophancy} (per 2400 data points), and transitions between correct (C) and incorrect (I) states (C$\rightarrow$I, I$\rightarrow$C), highlighting the nuanced effects of debate dynamics.}
\label{tab:debate_performance_gsm_plus2}
\end{table*}

\begin{table*}[htbp]
\centering
\begin{adjustbox}{width=\textwidth,center} 
\sisetup{round-mode=places,round-precision=2} 
\begin{tabular}{@{}ccccS[table-format=2.2]S[table-format=3.2]S[table-format=1.2]S[table-format=4.2]cccS[table-format=1.2]@{}} 
\toprule
\textbf{Agent 1} & \textbf{Agent 2} & \textbf{Agent 3} & \textbf{Agent Settings} & \multicolumn{1}{c}{\textbf{Accuracy}} & \multicolumn{1}{c}{\textbf{$\Delta$}} & \textbf{Debate} & \multicolumn{1}{c}{\textbf{Sycophancy}} & \textbf{C$\rightarrow$I} & \textbf{I$\rightarrow$C} & \textbf{Debate} \\
 & & & & & & \textbf{Rounds} & \multicolumn{1}{c}{\textbf{(Avg / 2400)}} & & & \textbf{Helped} \\
 & & & & & & \textbf{(Avg)} & & & & \textbf{(Overall)} \\
\midrule
Qwen-2.5-0.5B & Qwen-2.5-1.5B & Qwen-2.5-3B & Default & 60.00 & \textcolor{darkred}{1.75 $\downarrow$} & 2.35 & 2.05 & 338 & 1356 & 951 \\
Qwen-2.5-0.5B & Qwen-2.5-1.5B & Llama-3.1-3B & Default & 47.46 & \textcolor{darkgreen}{1.79 $\uparrow$} & 3.11 & 2.23 & 596 & 1086 & 718 \\
Qwen-2.5-0.5B & Qwen-2.5-1.5B & Phi-mini-3.8B & Default & 56.62 & \textcolor{darkred}{6.80 $\downarrow$} & 2.83 & 1.93 & 503 & 1168 & 857 \\
Qwen-2.5-0.5B & Qwen-2.5-3B & Llama-3.1-3B & Default & 59.62 & \textcolor{darkred}{2.13 $\downarrow$} & 2.83 & 1.90 & 364 & 1202 & 895 \\
Qwen-2.5-0.5B & Qwen-2.5-3B & Phi-mini-3.8B & Default & 65.25 & \textcolor{darkgreen}{1.83 $\uparrow$} & 2.42 & 1.48 & 353 & 1190 & 946 \\
Qwen-2.5-0.5B & Llama-3.1-3B & Phi-mini-3.8B & Default & 56.92 & \textcolor{darkred}{6.50 $\downarrow$} & 3.13 & 1.64 & 536 & 980 & 724 \\
Qwen-2.5-1.5B & Qwen-2.5-3B & Llama-3.1-3B & Default & 64.00 & \textcolor{darkgreen}{2.25 $\uparrow$} & 1.91 & 1.59 & 321 & 1048 & 773 \\
Qwen-2.5-1.5B & Qwen-2.5-3B & Phi-mini-3.8B & Default & 67.25 & \textcolor{darkgreen}{3.83 $\uparrow$} & 1.61 & 1.25 & 299 & 857 & 692 \\
Qwen-2.5-1.5B & Llama-3.1-3B & Phi-mini-3.8B & Default & 63.50 & \textcolor{darkgreen}{0.08 $\uparrow$} & 2.02 & 1.57 & 405 & 1079 & 766 \\
Qwen-2.5-3B & Phi-mini-3.8B & Llama-3.1-3B & Default & 69.08 & \textcolor{darkgreen}{5.66 $\uparrow$} & 1.58 & 1.20 & 255 & 825 & 653 \\
\midrule
Qwen-2.5-3B & Qwen-2.5-3B & Phi-mini-3.8B & Default & 68.79 & \textcolor{darkgreen}{7.04 $\uparrow$} & 1.13 & 0.90 & 291 & 461 & 340 \\
Qwen-2.5-3B & Phi-mini-3.8B & Phi-mini-3.8B & Default & 69.21 & \textcolor{darkgreen}{5.79 $\uparrow$} & 1.10 & 0.92 & 279 & 424 & 317 \\
Qwen-2.5-0.5B & Qwen-2.5-1.5B & Qwen-2.5-1.5B & Default & 49.88 & \textcolor{darkgreen}{7.88 $\uparrow$} & 2.44 & 2.50 & 456 & 1197 & 794 \\
Qwen-2.5-0.5B & Qwen-2.5-0.5B & Qwen-2.5-1.5B & Default & 37.21 & \textcolor{darkred}{4.79 $\downarrow$} & 3.07 & 3.24 & 589 & 969 & 607 \\
\bottomrule
\end{tabular}
\end{adjustbox}
\caption{Comparative Analysis of Mixed Multi-Agent Debate Settings on the \textbf{GSM-Plus} Dataset. This table examines performance when combining different language models in three-agent debate configurations. The first section shows combinations of three different models, while the second section explores configurations with duplicate models. The \textbf{$\Delta$} column indicates performance changes relative to the best single model in each combination, with improvements in \textcolor{darkgreen}{green} and declines in \textcolor{darkred}{red}. Metrics include \textbf{Debate Rounds}, normalized \textbf{Sycophancy} (per 2400 data points), and transitions between states (C$\rightarrow$I, I$\rightarrow$C).}
\label{tab:mixed_agents_gsm_plus}
\end{table*}


\begin{table*}[htbp]
\centering
\begin{adjustbox}{width=\textwidth,center} 
\sisetup{round-mode=places,round-precision=2} 
\begin{tabular}{@{}ccccS[table-format=2.2]S[table-format=1.2]S[table-format=1.2]S[table-format=3.0]ccc@{}} 
\toprule
\textbf{Agent 1} & \textbf{Agent 2} & \textbf{Agent Settings} & \textbf{Accuracy} & \multicolumn{1}{c}{\textbf{$\Delta$}} & \textbf{Debate} & \multicolumn{1}{c}{\textbf{Sycophancy}} & \textbf{C$\rightarrow$I} & \textbf{I$\rightarrow$C} & \textbf{Debate} \\
 & & & & & \textbf{Rounds} & \multicolumn{1}{c}{\textbf{(Avg / 2376)}} & & & \textbf{Helped} \\
 & & & & & \textbf{(Avg)} & & & & \textbf{(Overall)} \\
\midrule
Qwen-2.5-0.5B & Qwen-2.5-0.5B & Default & 52.90 & \textcolor{darkred}{1.73 $\downarrow$} & 1.15 & 0.99 & 460 & 550 & 482 \\
Qwen-2.5-0.5B & Qwen-2.5-0.5B & Deterministic & 53.24 & \textcolor{darkred}{1.39 $\downarrow$} & 0 & 0.00 & 0 & 0 & 0 \\
Qwen-2.5-0.5B & Qwen-2.5-0.5B & Exploratory & 49.07 & \textcolor{darkred}{5.56 $\downarrow$} & 1.46 & 1.09 & 558 & 628 & 530 \\
Qwen-2.5-0.5B & Qwen-2.5-0.5B & Det. \& Exp. & 52.99 & \textcolor{darkred}{1.64 $\downarrow$} & 1.15 & 0.97 & 426 & 572 & 516 \\
\midrule
Qwen-2.5-1.5B & Qwen-2.5-1.5B & Default & 86.15 & \textcolor{darkred}{0.47 $\downarrow$} & 0.38 & 0.38 & 130 & 415 & 403 \\
Qwen-2.5-1.5B & Qwen-2.5-1.5B & Deterministic & 84.60 & \textcolor{darkred}{2.02 $\downarrow$} & 0 & 0.00 & 0 & 0 & 0 \\
Qwen-2.5-1.5B & Qwen-2.5-1.5B & Exploratory & 83.42 & \textcolor{darkred}{3.20 $\downarrow$} & 0.55 & 0.55 & 160 & 574 & 547 \\
Qwen-2.5-1.5B & Qwen-2.5-1.5B & Det. \& Exp. & 86.62 & 0.00 & 0.41 & 0.42 & 135 & 449 & 434 \\
\midrule
Qwen-2.5-3B & Qwen-2.5-3B & Default & 94.02 & \textcolor{darkgreen}{0.96 $\uparrow$} & 0.14 & 0.13 & 56 & 117 & 114 \\
Qwen-2.5-3B & Qwen-2.5-3B & Deterministic & 93.35 & \textcolor{darkgreen}{0.29 $\uparrow$} & 0 & 0.00 & 0 & 0 & 0 \\
Qwen-2.5-3B & Qwen-2.5-3B & Exploratory & 94.15 & \textcolor{darkgreen}{1.09 $\uparrow$} & 0.16 & 0.15 & 49 & 158 & 157 \\
Qwen-2.5-3B & Qwen-2.5-3B & Det. \& Exp. & 94.07 & \textcolor{darkgreen}{1.01 $\uparrow$} & 0.15 & 0.13 & 70 & 126 & 124 \\
\midrule
Qwen-2.5-7B & Qwen-2.5-7B & Default & 96.17 & \textcolor{darkgreen}{1.48 $\uparrow$} & 0.05 & 0.05 & 31 & 39 & 37 \\
Qwen-2.5-7B & Qwen-2.5-7B & Deterministic & 96.55 & \textcolor{darkgreen}{1.86 $\uparrow$} & 0 & 0.00 & 0 & 0 & 0 \\
Qwen-2.5-7B & Qwen-2.5-7B & Exploratory & 96.93 & \textcolor{darkgreen}{2.24 $\uparrow$} & 0.05 & 0.05 & 21 & 57 & 53 \\
Qwen-2.5-7B & Qwen-2.5-7B & Det. \& Exp. & 96.46 & \textcolor{darkgreen}{1.77 $\uparrow$} & 0.05 & 0.04 & 30 & 35 & 34 \\
\midrule
Qwen-2.5-14B & Qwen-2.5-14B & Default & 98.19 & \textcolor{darkgreen}{2.53 $\uparrow$} & 0.03 & 0.02 & 15 & 21 & 21 \\
Qwen-2.5-14B & Qwen-2.5-14B & Deterministic & 97.77 & \textcolor{darkgreen}{2.11 $\uparrow$} & 0 & 0.00 & 0 & 0 & 0 \\
Qwen-2.5-14B & Qwen-2.5-14B & Exploratory & 98.15 & \textcolor{darkgreen}{2.49 $\uparrow$} & 0.02 & 0.02 & 8 & 20 & 20 \\
Qwen-2.5-14B & Qwen-2.5-14B & Det. \& Exp. & 97.94 & \textcolor{darkgreen}{2.28 $\uparrow$} & 0.03 & 0.02 & 16 & 24 & 24 \\
\midrule
Qwen-2.5-32B & Qwen-2.5-32B & Default & 98.53 & \textcolor{darkgreen}{0.21 $\uparrow$} & 0.02 & 0.03 & 10 & 14 & 13 \\
Qwen-2.5-32B & Qwen-2.5-32B & Deterministic & 98.36 & \textcolor{darkgreen}{0.04 $\uparrow$} & 0 & 0.00 & 0 & 0 & 0 \\
Qwen-2.5-32B & Qwen-2.5-32B & Exploratory & 98.53 & \textcolor{darkgreen}{0.21 $\uparrow$} & 0.02 & 0.03 & 8 & 14 & 14 \\
Qwen-2.5-32B & Qwen-2.5-32B & Det. \& Exp. & 98.36 & \textcolor{darkgreen}{0.04 $\uparrow$} & 0.02 & 0.02 & 9 & 10 & 8 \\
\midrule
Phi-mini-3.8B & Phi-mini-3.8B & Default & 95.88 & \textcolor{darkgreen}{3.92 $\uparrow$} & 0.11 & 0.16 & 40 & 71 & 60 \\
Phi-mini-3.8B & Phi-mini-3.8B & Deterministic & 95.37 & \textcolor{darkgreen}{3.41 $\uparrow$} & 0 & 0.00 & 0 & 0 & 0 \\
Phi-mini-3.8B & Phi-mini-3.8B & Exploratory & 94.74 & \textcolor{darkgreen}{2.78 $\uparrow$} & 0.16 & 0.21 & 59 & 126 & 116 \\
Phi-mini-3.8B & Phi-mini-3.8B & Det. \& Exp. & 94.95 & \textcolor{darkgreen}{2.99 $\uparrow$} & 0.14 & 0.19 & 56 & 89 & 78 \\
\midrule
Mistral-7B & Mistral-7B & Default & 81.06 & \textcolor{darkgreen}{0.04 $\uparrow$} & 0.35 & 0.28 & 158 & 227 & 219 \\
Mistral-7B & Mistral-7B & Deterministic & 80.43 & \textcolor{darkred}{0.59 $\downarrow$} & 0 & 0.00 & 0 & 0 & 0 \\
Mistral-7B & Mistral-7B & Exploratory & 80.18 & \textcolor{darkred}{0.84 $\downarrow$} & 0.43 & 0.32 & 203 & 261 & 251 \\
Mistral-7B & Mistral-7B & Det. \& Exp. & 82.41 & \textcolor{darkgreen}{1.39 $\uparrow$} & 0.37 & 0.27 & 129 & 240 & 235 \\
\midrule
Llama-3.1-3B & Llama-3.1-3B & Default & 87.71 & \textcolor{darkgreen}{3.07 $\uparrow$} & 0.26 & 0.21 & 128 & 163 & 153 \\
Llama-3.1-3B & Llama-3.1-3B & Deterministic & 86.66 & \textcolor{darkgreen}{2.02 $\uparrow$} & 0 & 0.00 & 0 & 0 & 0 \\
Llama-3.1-3B & Llama-3.1-3B & Exploratory & 88.09 & \textcolor{darkgreen}{3.45 $\uparrow$} & 0.28 & 0.26 & 118 & 216 & 208 \\
Llama-3.1-3B & Llama-3.1-3B & Det. \& Exp. & 86.91 & \textcolor{darkgreen}{2.27 $\uparrow$} & 0.28 & 0.22 & 127 & 181 & 172 \\
\midrule
Llama-3.1-8B & Llama-3.1-8B & Default & 94.44 & \textcolor{darkgreen}{5.34 $\uparrow$} & 0.11 & 0.11 & 54 & 79 & 75 \\
Llama-3.1-8B & Llama-3.1-8B & Deterministic & 93.64 & \textcolor{darkgreen}{4.54 $\uparrow$} & 0 & 0.00 & 0 & 0 & 0 \\
Llama-3.1-8B & Llama-3.1-8B & Exploratory & 93.60 & \textcolor{darkgreen}{4.50 $\uparrow$} & 0.15 & 0.17 & 60 & 118 & 109 \\
Llama-3.1-8B & Llama-3.1-8B & Det. \& Exp. & 94.53 & \textcolor{darkgreen}{5.43 $\uparrow$} & 0.12 & 0.13 & 54 & 95 & 93 \\
\bottomrule
\end{tabular}
\end{adjustbox}
\caption{Comparative Analysis of Language Model Performance in Multi-Agent Debate Settings on the \textbf{ARC-Easy} Dataset. This table showcases the impact of different \textbf{Agent Settings} (controlling temperature and top\_p parameters like Default, Deterministic, Exploratory, and a combination) on the \textbf{Accuracy} of various language models. The \textbf{$\Delta$} column quantifies the \textbf{improvement (or decline) over the single base model performance}. Further metrics include average \textbf{Debate Rounds}, normalized \textbf{Sycophancy} (per 2376 data points), and transitions between correct (C) and incorrect (I) states (C$\rightarrow$I, I$\rightarrow$C), highlighting the nuanced effects of debate dynamics.}
\label{tab:arc_easy_debate_performance}
\end{table*}

\begin{table*}[htbp]
\centering
\begin{adjustbox}{width=\textwidth,center} 
\sisetup{round-mode=places,round-precision=2} 
\begin{tabular}{@{}cccS[table-format=2.2]S[table-format=2.2]S[table-format=2.2]S[table-format=1.2]S[table-format=1.2]S[table-format=3.0]ccc@{}} 
\toprule
\textbf{Agent 1} & \textbf{Agent 2} & \textbf{Agent Settings} & \textbf{Accuracy} & \multicolumn{1}{c}{\textbf{$\Delta$ Lower}} & \multicolumn{1}{c}{\textbf{$\Delta$ Upper}} & \textbf{Debate} & \multicolumn{1}{c}{\textbf{Sycophancy}} & \textbf{C$\rightarrow$I} & \textbf{I$\rightarrow$C} & \textbf{Debate} \\
 & & & & & & \textbf{Rounds} & \multicolumn{1}{c}{\textbf{(Avg / 2376)}} & & & \textbf{Helped} \\
 & & & & & & \textbf{(Avg)} & & & & \textbf{(Overall)} \\
\midrule
Qwen-2.5-0.5B & Qwen-2.5-1.5B & Default & 76.98 & \textcolor{darkgreen}{22.35 $\uparrow$} & \textcolor{darkred}{9.64 $\downarrow$} & 0.95 & 0.75 & 262 & 804 & 760 \\
Qwen-2.5-0.5B & Qwen-2.5-1.5B & Deterministic & 79.38 & \textcolor{darkgreen}{24.75 $\uparrow$} & \textcolor{darkred}{7.24 $\downarrow$} & 0.81 & 0.62 & 200 & 734 & 711 \\
Qwen-2.5-0.5B & Qwen-2.5-1.5B & Exploratory & 73.19 & \textcolor{darkgreen}{18.56 $\uparrow$} & \textcolor{darkred}{13.43 $\downarrow$} & 1.16 & 0.85 & 300 & 899 & 828 \\
Qwen-2.5-0.5B & Qwen-2.5-1.5B & Det. \& Exp. & 75.21 & \textcolor{darkgreen}{20.58 $\uparrow$} & \textcolor{darkred}{11.41 $\downarrow$} & 0.95 & 0.78 & 260 & 846 & 790 \\
Qwen-2.5-0.5B & Qwen-2.5-1.5B & Exp. \& Det. & 77.65 & \textcolor{darkgreen}{23.02 $\uparrow$} & \textcolor{darkred}{8.97 $\downarrow$} & 1.07 & 0.75 & 275 & 829 & 794 \\
\midrule
Qwen-2.5-1.5B & Llama-3.1-3B & Default & 88.55 & \textcolor{darkgreen}{1.93 $\uparrow$} & \textcolor{darkgreen}{3.91 $\uparrow$} & 0.40 & 0.39 & 146 & 376 & 357 \\
Qwen-2.5-1.5B & Llama-3.1-3B & Deterministic & 88.13 & \textcolor{darkgreen}{1.51 $\uparrow$} & \textcolor{darkgreen}{3.49 $\uparrow$} & 0.29 & 0.24 & 150 & 242 & 239 \\
Qwen-2.5-1.5B & Llama-3.1-3B & Exploratory & 88.05 & \textcolor{darkgreen}{1.43 $\uparrow$} & \textcolor{darkgreen}{3.41 $\uparrow$} & 0.49 & 0.48 & 161 & 483 & 457 \\
Qwen-2.5-1.5B & Llama-3.1-3B & Det. \& Exp. & 86.99 & \textcolor{darkgreen}{0.37 $\uparrow$} & \textcolor{darkgreen}{1.35 $\uparrow$} & 0.37 & 0.39 & 172 & 290 & 277 \\
Qwen-2.5-1.5B & Llama-3.1-3B & Exp. \& Det. & 87.71 & \textcolor{darkgreen}{1.09 $\uparrow$} & \textcolor{darkgreen}{2.07 $\uparrow$} & 0.45 & 0.40 & 165 & 447 & 433 \\
\midrule
Qwen-2.5-3B & Phi-mini-3.8B & Default & 95.24 & \textcolor{darkgreen}{2.18 $\uparrow$} & \textcolor{darkgreen}{3.28 $\uparrow$} & 0.15 & 0.14 & 61 & 135 & 132 \\
Qwen-2.5-3B & Phi-mini-3.8B & Deterministic & 94.91 & \textcolor{darkgreen}{1.85 $\uparrow$} & \textcolor{darkgreen}{2.95 $\uparrow$} & 0.14 & 0.12 & 72 & 106 & 102 \\
Qwen-2.5-3B & Phi-mini-3.8B & Exploratory & 95.24 & \textcolor{darkgreen}{2.18 $\uparrow$} & \textcolor{darkgreen}{3.28 $\uparrow$} & 0.17 & 0.16 & 57 & 184 & 178 \\
Qwen-2.5-3B & Phi-mini-3.8B & Det. \& Exp. & 94.91 & \textcolor{darkgreen}{1.85 $\uparrow$} & \textcolor{darkgreen}{2.95 $\uparrow$} & 0.17 & 0.15 & 68 & 148 & 148 \\
Qwen-2.5-3B & Phi-mini-3.8B & Exp. \& Det. & 95.75 & \textcolor{darkgreen}{2.69 $\uparrow$} & \textcolor{darkgreen}{3.79 $\uparrow$} & 0.15 & 0.14 & 58 & 146 & 139 \\
\midrule
Qwen-2.5-1.5B & Qwen-2.5-3B & Default & 91.88 & \textcolor{darkgreen}{5.26 $\uparrow$} & \textcolor{darkred}{1.18 $\downarrow$} & 0.33 & 0.29 & 112 & 363 & 359 \\
Qwen-2.5-1.5B & Qwen-2.5-3B & Deterministic & 92.59 & \textcolor{darkgreen}{5.97 $\uparrow$} & \textcolor{darkred}{0.47 $\downarrow$} & 0.24 & 0.23 & 94 & 263 & 254 \\
Qwen-2.5-1.5B & Qwen-2.5-3B & Exploratory & 91.79 & \textcolor{darkgreen}{5.17 $\uparrow$} & \textcolor{darkred}{1.27 $\downarrow$} & 0.42 & 0.38 & 95 & 498 & 487 \\
Qwen-2.5-1.5B & Qwen-2.5-3B & Det. \& Exp. & 92.76 & \textcolor{darkgreen}{6.14 $\uparrow$} & \textcolor{darkred}{0.20 $\downarrow$} & 0.27 & 0.27 & 81 & 294 & 286 \\
Qwen-2.5-1.5B & Qwen-2.5-3B & Exp. \& Det. & 92.51 & \textcolor{darkgreen}{5.89 $\uparrow$} & \textcolor{darkred}{0.45 $\downarrow$} & 0.39 & 0.32 & 96 & 469 & 466 \\
\midrule
Llama-3.1-3B & Llama-3.1-8B & Default & 91.79 & \textcolor{darkgreen}{7.15 $\uparrow$} & \textcolor{darkgreen}{2.69 $\uparrow$} & 0.24 & 0.22 & 110 & 184 & 179 \\
Llama-3.1-3B & Llama-3.1-8B & Deterministic & 91.12 & \textcolor{darkgreen}{6.48 $\uparrow$} & \textcolor{darkgreen}{2.02 $\uparrow$} & 0.22 & 0.16 & 113 & 138 & 133 \\
Llama-3.1-3B & Llama-3.1-8B & Exploratory & 90.61 & \textcolor{darkgreen}{5.97 $\uparrow$} & \textcolor{darkgreen}{1.51 $\uparrow$} & 0.28 & 0.27 & 115 & 202 & 192 \\
Llama-3.1-3B & Llama-3.1-8B & Det. \& Exp. & 90.99 & \textcolor{darkgreen}{6.35 $\uparrow$} & \textcolor{darkgreen}{1.89 $\uparrow$} & 0.24 & 0.18 & 108 & 152 & 149 \\
Llama-3.1-3B & Llama-3.1-8B & Exp. \& Det. & 91.96 & \textcolor{darkgreen}{7.32 $\uparrow$} & \textcolor{darkgreen}{2.86 $\uparrow$} & 0.28 & 0.26 & 99 & 229 & 222 \\
\midrule
Qwen-2.5-7B & Qwen-2.5-14B & Default & 97.94 & \textcolor{darkgreen}{3.25 $\uparrow$} & \textcolor{darkgreen}{2.28 $\uparrow$} & 0.05 & 0.05 & 21 & 55 & 55 \\
Qwen-2.5-7B & Qwen-2.5-14B & Deterministic & 97.64 & \textcolor{darkgreen}{2.95 $\uparrow$} & \textcolor{darkgreen}{1.98 $\uparrow$} & 0.07 & 0.04 & 20 & 48 & 47 \\
Qwen-2.5-7B & Qwen-2.5-14B & Exploratory & 97.39 & \textcolor{darkgreen}{2.70 $\uparrow$} & \textcolor{darkgreen}{1.73 $\uparrow$} & 0.08 & 0.07 & 32 & 67 & 66 \\
Qwen-2.5-7B & Qwen-2.5-14B & Det. \& Exp. & 97.43 & \textcolor{darkgreen}{2.74 $\uparrow$} & \textcolor{darkgreen}{1.77 $\uparrow$} & 0.06 & 0.05 & 33 & 49 & 48 \\
Qwen-2.5-7B & Qwen-2.5-14B & Exp. \& Det. & 97.47 & \textcolor{darkgreen}{2.78 $\uparrow$} & \textcolor{darkgreen}{1.81 $\uparrow$} & 0.07 & 0.04 & 27 & 49 & 48 \\
\bottomrule
\end{tabular}
\end{adjustbox}
\caption{Comparative Analysis of Different Language Model Pairs in Multi-Agent Debate Settings on the \textbf{ARC-Easy} Dataset. This table showcases the impact of different \textbf{Agent Settings} (controlling temperature and top\_p parameters) on the \textbf{Accuracy} of various model pairs. The \textbf{$\Delta$ Lower} and \textbf{$\Delta$ Upper} columns quantify the improvement (or decline) over each individual model's single-agent performance. Further metrics include average \textbf{Debate Rounds}, normalized \textbf{Sycophancy} (per 2376 data points), and transitions between correct (C) and incorrect (I) states (C$\rightarrow$I, I$\rightarrow$C), highlighting the nuanced effects of debate dynamics between different model pairings.}
\label{tab:arc_easy_model_pairs}
\end{table*}

\begin{table*}[htbp]
\centering
\begin{adjustbox}{width=\textwidth,center} 
\sisetup{round-mode=places,round-precision=2} 
\begin{tabular}{@{}ccccS[table-format=2.2]S[table-format=3.2]S[table-format=1.2]S[table-format=4.2]cccS[table-format=1.2]@{}} 
\toprule
\textbf{Agent 1} & \textbf{Agent 2} & \textbf{Agent Settings} & \textbf{MAD Accuracy} & \multicolumn{1}{c}{\textbf{$\Delta$}} & \textbf{Debate} & \multicolumn{1}{c}{\textbf{Sycophancy}} & \textbf{C$\rightarrow$I} & \textbf{I$\rightarrow$C} & \textbf{Debate} \\
 & & & \textbf{(RCR Prompting)} &  & \textbf{Rounds} & \multicolumn{1}{c}{\textbf{(Avg / 2376)}} & & & \textbf{Helped} \\
 & & & &  & \textbf{(Avg)} & & & & \textbf{(Overall)} \\
\midrule
Qwen-2.5-0.5B & Qwen-2.5-0.5B & Both: Default & 51.30 & \textcolor{darkred}{3.33 $\downarrow$} & 2.18 & 2.67 & 1046 & 990 & 642 \\
Qwen-2.5-0.5B & Qwen-2.5-0.5B & Both: Deterministic & 53.24 & \textcolor{darkred}{1.39 $\downarrow$} & 0 & 0.00 & 0 & 0 & 0 \\
Qwen-2.5-0.5B & Qwen-2.5-0.5B & Both: Exploratory & 46.80 & \textcolor{darkred}{7.83 $\downarrow$} & 2.78 & 3.22 & 1228 & 1099 & 655 \\
Qwen-2.5-0.5B & Qwen-2.5-0.5B & 1 Det. \& 2 Exp. & 48.99 & \textcolor{darkred}{5.64 $\downarrow$} & 2.47 & 2.82 & 1136 & 1053 & 658 \\
Qwen-2.5-0.5B & Qwen-2.5-0.5B & 2 Det. \& 1 Exp. & 50.80 & \textcolor{darkred}{3.83 $\downarrow$} & 1.34 & 1.60 & 794 & 793 & 495 \\
\midrule
Qwen-2.5-1.5B & Qwen-2.5-1.5B & Both: Default & 87.37 & \textcolor{darkgreen}{0.75 $\uparrow$} & 0.63 & 0.84 & 232 & 717 & 573 \\
Qwen-2.5-1.5B & Qwen-2.5-1.5B & Both: Deterministic & 84.60 & \textcolor{darkred}{2.02 $\downarrow$} & 0 & 0.00 & 0 & 0 & 0 \\
Qwen-2.5-1.5B & Qwen-2.5-1.5B & Both: Exploratory & 85.61 & \textcolor{darkred}{1.01 $\downarrow$} & 0.90 & 1.17 & 279 & 1011 & 795 \\
Qwen-2.5-1.5B & Qwen-2.5-1.5B & 1 Det. \& 2 Exp. & 86.32 & \textcolor{darkred}{0.30 $\downarrow$} & 0.76 & 0.98 & 275 & 834 & 672 \\
Qwen-2.5-1.5B & Qwen-2.5-1.5B & 2 Det. \& 1 Exp. & 86.53 & \textcolor{darkred}{0.09 $\downarrow$} & 0.43 & 0.62 & 198 & 587 & 451 \\
\midrule
Qwen-2.5-3B & Qwen-2.5-3B & Both: Default & 94.87 & \textcolor{darkgreen}{1.81 $\uparrow$} & 0.19 & 0.19 & 80 & 196 & 165 \\
Qwen-2.5-3B & Qwen-2.5-3B & Both: Deterministic & 93.35 & \textcolor{darkgreen}{0.29 $\uparrow$} & 0 & 0.00 & 0 & 0 & 0 \\
Qwen-2.5-3B & Qwen-2.5-3B & Both: Exploratory & 94.28 & \textcolor{darkgreen}{1.22 $\uparrow$} & 0.25 & 0.28 & 102 & 252 & 206 \\
Qwen-2.5-3B & Qwen-2.5-3B & 1 Det. \& 2 Exp. & 94.70 & \textcolor{darkgreen}{1.64 $\uparrow$} & 0.25 & 0.23 & 90 & 238 & 195 \\
Qwen-2.5-3B & Qwen-2.5-3B & 2 Det. \& 1 Exp. & 93.94 & \textcolor{darkgreen}{0.88 $\uparrow$} & 0.20 & 0.18 & 94 & 162 & 117 \\
\midrule
Qwen-2.5-7B & Qwen-2.5-7B & Both: Default & 96.21 & \textcolor{darkgreen}{1.52 $\uparrow$} & 0.08 & 0.08 & 53 & 69 & 58 \\
Qwen-2.5-7B & Qwen-2.5-7B & Both: Deterministic & 96.17 & \textcolor{darkgreen}{1.48 $\uparrow$} & 0 & 0.00 & 0 & 0 & 0 \\
Qwen-2.5-7B & Qwen-2.5-7B & Both: Exploratory & 96.55 & \textcolor{darkgreen}{1.86 $\uparrow$} & 0.10 & 0.11 & 57 & 86 & 71 \\
Qwen-2.5-7B & Qwen-2.5-7B & 1 Det. \& 2 Exp. & 96.55 & \textcolor{darkgreen}{1.86 $\uparrow$} & 0.10 & 0.11 & 56 & 78 & 65 \\
Qwen-2.5-7B & Qwen-2.5-7B & 2 Det. \& 1 Exp. & 96.34 & \textcolor{darkgreen}{1.65 $\uparrow$} & 0.07 & 0.07 & 39 & 56 & 40 \\
\midrule
Qwen-2.5-14B & Qwen-2.5-14B & Both: Default & 98.15 & \textcolor{darkgreen}{2.49 $\uparrow$} & 0.04 & 0.04 & 23 & 29 & 26 \\
Qwen-2.5-14B & Qwen-2.5-14B & Both: Deterministic & 97.77 & \textcolor{darkgreen}{2.11 $\uparrow$} & 0 & 0.00 & 0 & 0 & 0 \\
Qwen-2.5-14B & Qwen-2.5-14B & Both: Exploratory & 98.19 & \textcolor{darkgreen}{2.53 $\uparrow$} & 0.04 & 0.05 & 18 & 40 & 36 \\
Qwen-2.5-14B & Qwen-2.5-14B & 1 Det. \& 2 Exp. & 98.02 & \textcolor{darkgreen}{2.36 $\uparrow$} & 0.03 & 0.04 & 28 & 40 & 31 \\
Qwen-2.5-14B & Qwen-2.5-14B & 2 Det. \& 1 Exp. & 97.81 & \textcolor{darkgreen}{2.15 $\uparrow$} & 0.03 & 0.03 & 23 & 28 & 25 \\
\midrule
Qwen-2.5-32B & Qwen-2.5-32B & Both: Default & 98.57 & \textcolor{darkgreen}{0.25 $\uparrow$} & 0.02 & 0.03 & 16 & 15 & 13 \\
Qwen-2.5-32B & Qwen-2.5-32B & Both: Deterministic & 98.36 & \textcolor{darkgreen}{0.04 $\uparrow$} & 0 & 0.00 & 0 & 0 & 0 \\
Qwen-2.5-32B & Qwen-2.5-32B & Both: Exploratory & 98.48 & \textcolor{darkgreen}{0.16 $\uparrow$} & 0.02 & 0.02 & 15 & 14 & 14 \\
Qwen-2.5-32B & Qwen-2.5-32B & 1 Det. \& 2 Exp. & 98.48 & \textcolor{darkgreen}{0.16 $\uparrow$} & 0.02 & 0.03 & 16 & 15 & 12 \\
Qwen-2.5-32B & Qwen-2.5-32B & 2 Det. \& 1 Exp. & 98.32 & \textcolor{darkgreen}{0.00} & 0.01 & 0.02 & 12 & 9 & 6 \\
\midrule
Phi-mini-3.8B & Phi-mini-3.8B & Both: Default & 95.79 & \textcolor{darkgreen}{3.83 $\uparrow$} & 0.16 & 0.28 & 79 & 138 & 105 \\
Phi-mini-3.8B & Phi-mini-3.8B & Both: Deterministic & 95.37 & \textcolor{darkgreen}{3.41 $\uparrow$} & 0 & 0.00 & 0 & 0 & 0 \\
Phi-mini-3.8B & Phi-mini-3.8B & Both: Exploratory & 94.91 & \textcolor{darkgreen}{2.95 $\uparrow$} & 0.28 & 0.43 & 110 & 234 & 185 \\
Phi-mini-3.8B & Phi-mini-3.8B & 1 Det. \& 2 Exp. & 96.34 & \textcolor{darkgreen}{4.38 $\uparrow$} & 0.18 & 0.27 & 70 & 189 & 149 \\
Phi-mini-3.8B & Phi-mini-3.8B & 2 Det. \& 1 Exp. & 95.92 & \textcolor{darkgreen}{3.96 $\uparrow$} & 0.13 & 0.24 & 53 & 115 & 83 \\
\midrule
Llama-3.1-3B & Llama-3.1-3B & Both: Default & 87.33 & \textcolor{darkgreen}{2.69 $\uparrow$} & 0.46 & 0.44 & 252 & 292 & 227 \\
Llama-3.1-3B & Llama-3.1-3B & Both: Deterministic & 87.63 & \textcolor{darkgreen}{2.99 $\uparrow$} & 0 & 0.00 & 0 & 0 & 0 \\
Llama-3.1-3B & Llama-3.1-3B & Both: Exploratory & 87.71 & \textcolor{darkgreen}{3.07 $\uparrow$} & 0.58 & 0.61 & 255 & 415 & 323 \\
Llama-3.1-3B & Llama-3.1-3B & 1 Det. \& 2 Exp. & 87.58 & \textcolor{darkgreen}{2.94 $\uparrow$} & 0.53 & 0.48 & 241 & 328 & 259 \\
Llama-3.1-3B & Llama-3.1-3B & 2 Det. \& 1 Exp. & 88.47 & \textcolor{darkgreen}{3.83 $\uparrow$} & 0.32 & 0.27 & 148 & 236 & 169 \\
\midrule
Llama-3.1-8B & Llama-3.1-8B & Both: Default & 93.86 & \textcolor{darkgreen}{4.76 $\uparrow$} & 0.20 & 0.26 & 114 & 139 & 102 \\
Llama-3.1-8B & Llama-3.1-8B & Both: Deterministic & 93.64 & \textcolor{darkgreen}{4.54 $\uparrow$} & 0 & 0.00 & 0 & 0 & 0 \\
Llama-3.1-8B & Llama-3.1-8B & Both: Exploratory & 94.19 & \textcolor{darkgreen}{5.09 $\uparrow$} & 0.25 & 0.36 & 130 & 190 & 141 \\
Llama-3.1-8B & Llama-3.1-8B & 1 Det. \& 2 Exp. & 94.11 & \textcolor{darkgreen}{5.01 $\uparrow$} & 0.23 & 0.33 & 119 & 185 & 143 \\
Llama-3.1-8B & Llama-3.1-8B & 2 Det. \& 1 Exp. & 94.49 & \textcolor{darkgreen}{5.39 $\uparrow$} & 0.14 & 0.20 & 69 & 139 & 89 \\
\midrule
Mistral-7B & Mistral-7B & Both: Default & 82.20 & \textcolor{darkgreen}{1.18 $\uparrow$} & 0.69 & 0.71 & 318 & 469 & 342 \\
Mistral-7B & Mistral-7B & Both: Deterministic & 80.43 & \textcolor{darkred}{0.59 $\downarrow$} & 0 & 0.00 & 0 & 0 & 0 \\
Mistral-7B & Mistral-7B & Both: Exploratory & 82.66 & \textcolor{darkgreen}{1.64 $\uparrow$} & 0.83 & 0.88 & 325 & 566 & 429 \\
Mistral-7B & Mistral-7B & 1 Det. \& 2 Exp. & 82.37 & \textcolor{darkgreen}{1.35 $\uparrow$} & 0.78 & 0.81 & 324 & 506 & 376 \\
Mistral-7B & Mistral-7B & 2 Det. \& 1 Exp. & 81.69 & \textcolor{darkgreen}{0.67 $\uparrow$} & 0.47 & 0.51 & 230 & 346 & 230 \\
\bottomrule
\end{tabular}
\end{adjustbox}
\caption{Comparative Analysis of Language Model Performance in Multi-Agent Debate Settings on the \textbf{ARC-Easy} Dataset. This table showcases the impact of different \textbf{Agent Settings} (controlling temperature and top\_p parameters like Default, Deterministic, Exploratory, and combinations) on the \textbf{MAD Accuracy (RCR Prompting)} of various language models. The \textbf{$\Delta$} column quantifies the \textbf{improvement (or decline) over the single base model performance}. Further metrics include average \textbf{Debate Rounds}, normalized \textbf{Sycophancy} (per 2376 data points), and transitions between correct (C) and incorrect (I) states (C$\rightarrow$I, I$\rightarrow$C), highlighting the nuanced effects of debate dynamics.}
\label{tab:debate_performance_arc_easy}
\end{table*}

\begin{table*}[htbp]
\centering
\begin{adjustbox}{width=\textwidth,center} 
\sisetup{round-mode=places,round-precision=2} 
\begin{tabular}{@{}cccS[table-format=2.2]S[table-format=3.2]S[table-format=1.2]S[table-format=4.2]cccS[table-format=1.2]@{}} 
\toprule
\textbf{Agent 1} & \textbf{Agent 2} & \textbf{Agent 3} & \textbf{MAD Accuracy} & \multicolumn{1}{c}{\textbf{$\Delta$}} & \textbf{Debate} & \multicolumn{1}{c}{\textbf{Sycophancy}} & \textbf{C$\rightarrow$I} & \textbf{I$\rightarrow$C} & \textbf{Debate} \\
 & & & \textbf{(RCR Prompting)} &  & \textbf{Rounds} & \multicolumn{1}{c}{\textbf{(Avg / 2376)}} & & & \textbf{Helped} \\
 & & & &  & \textbf{(Avg)} & & & & \textbf{(Overall)} \\
\midrule
Qwen-2.5-0.5B & Qwen-2.5-1.5B & Qwen-2.5-3B & 92.72 & \textcolor{darkred}{0.34 $\downarrow$} & 1.00 & 0.95 & 145 & 1377 & 1153 \\
Qwen-2.5-0.5B & Qwen-2.5-1.5B & Llama-3.1-3B & 84.64 & 0.00 & 1.18 & 1.27 & 387 & 1223 & 1006 \\
Qwen-2.5-0.5B & Qwen-2.5-1.5B & Phi-mini-3.8B & 92.93 & \textcolor{darkgreen}{0.97 $\uparrow$} & 1.03 & 1.04 & 184 & 1379 & 1156 \\
Qwen-2.5-0.5B & Qwen-2.5-3B & Llama-3.1-3B & 91.20 & \textcolor{darkred}{1.86 $\downarrow$} & 1.13 & 0.99 & 213 & 1221 & 1070 \\
Qwen-2.5-0.5B & Qwen-2.5-3B & Phi-mini-3.8B & 89.48 & \textcolor{darkred}{3.58 $\downarrow$} & 1.09 & 1.12 & 299 & 1157 & 1024 \\
Qwen-2.5-0.5B & Llama-3.1-3B & Phi-mini-3.8B & 91.79 & \textcolor{darkred}{0.17 $\downarrow$} & 0.58 & 0.72 & 238 & 559 & 479 \\
\midrule
Qwen-2.5-1.5B & Qwen-2.5-3B & Llama-3.1-3B & 91.84 & \textcolor{darkred}{1.22 $\downarrow$} & 0.56 & 0.60 & 189 & 560 & 479 \\
Qwen-2.5-1.5B & Qwen-2.5-3B & Phi-mini-3.8B & 95.54 & \textcolor{darkgreen}{2.48 $\uparrow$} & 0.39 & 0.45 & 103 & 509 & 449 \\
Qwen-2.5-1.5B & Llama-3.1-3B & Phi-mini-3.8B & 91.79 & \textcolor{darkred}{0.17 $\downarrow$} & 0.58 & 0.72 & 238 & 559 & 479 \\
Qwen-2.5-3B & Phi-mini-3.8B & Llama-3.1-3B & 94.07 & \textcolor{darkgreen}{1.01 $\uparrow$} & 0.41 & 0.43 & 162 & 332 & 283 \\
\midrule
Qwen-2.5-3B & Qwen-2.5-3B & Phi-mini-3.8B & 95.88 & \textcolor{darkgreen}{2.82 $\uparrow$} & 0.26 & 0.26 & 86 & 253 & 214 \\
Qwen-2.5-3B & Phi-mini-3.8B & Phi-mini-3.8B & 96.34 & \textcolor{darkgreen}{3.28 $\uparrow$} & 0.26 & 0.31 & 71 & 227 & 180 \\
Qwen-2.5-0.5B & Qwen-2.5-1.5B & Qwen-2.5-1.5B & 84.64 & \textcolor{darkred}{2.00 $\downarrow$} & 1.22 & 1.22 & 300 & 1229 & 1012 \\
Qwen-2.5-0.5B & Qwen-2.5-0.5B & Qwen-2.5-1.5B & 72.43 & \textcolor{darkred}{14.19 $\downarrow$} & 1.86 & 2.11 & 616 & 1400 & 982 \\
\bottomrule
\end{tabular}
\end{adjustbox}
\caption{Comparative Analysis of Multi-Model Combinations in Agent Debate Settings on the \textbf{ARC-Easy} Dataset. This table showcases the performance of heterogeneous agent teams consisting of different language models. The \textbf{MAD Accuracy (RCR Prompting)} reflects the team performance, while the \textbf{$\Delta$} column quantifies the \textbf{improvement (or decline) relative to the best single model} in each combination. Additional metrics include average \textbf{Debate Rounds}, normalized \textbf{Sycophancy} (per 2376 data points), and transitions between correct (C) and incorrect (I) states (C$\rightarrow$I, I$\rightarrow$C), revealing how diverse model combinations affect debate dynamics and overall helpfulness.}
\label{tab:multi_model_debate_performance}
\end{table*}


\begin{table*}[htbp]
\centering
\begin{adjustbox}{width=\textwidth,center} 
\sisetup{round-mode=places,round-precision=2} 
\begin{tabular}{@{}ccccS[table-format=2.2]S[table-format=3.2]S[table-format=1.2]S[table-format=4.2]cccS[table-format=1.2]@{}} 
\toprule
\textbf{Agent 1} & \textbf{Agent 2} & \textbf{Agent Settings} & \textbf{MAD Accuracy} & \multicolumn{1}{c}{\textbf{$\Delta$}} & \textbf{Debate} & \multicolumn{1}{c}{\textbf{Sycophancy}} & \textbf{C$\rightarrow$I} & \textbf{I$\rightarrow$C} & \textbf{Debate} \\
 & & & \textbf{(ARC-Challenge)} &  & \textbf{Rounds} & \multicolumn{1}{c}{\textbf{(Avg / 1172)}} & & & \textbf{Helped} \\
 & & & &  & \textbf{(Avg)} & & & & \textbf{(Overall)} \\
\midrule
Qwen-2.5-0.5B & Qwen-2.5-0.5B & Both: Default & 39.51 & \textcolor{darkgreen}{1.54 $\uparrow$} & 1.32 & 1.10 & 253 & 265 & 228 \\
Qwen-2.5-0.5B & Qwen-2.5-0.5B & Both: Deterministic & 40.78 & \textcolor{darkgreen}{2.81 $\uparrow$} & 0 & 0.00 & 0 & 0 & 0 \\
Qwen-2.5-0.5B & Qwen-2.5-0.5B & Both: Exploratory & 37.54 & \textcolor{darkred}{0.43 $\downarrow$} & 1.51 & 1.14 & 266 & 309 & 245 \\
Qwen-2.5-0.5B & Qwen-2.5-0.5B & Both: Det. \& Exp. & 39.85 & \textcolor{darkgreen}{1.88 $\uparrow$} & 1.34 & 1.12 & 247 & 259 & 227 \\
\midrule
Qwen-2.5-1.5B & Qwen-2.5-1.5B & Both: Default & 70.90 & \textcolor{darkgreen}{1.69 $\uparrow$} & 0.57 & 0.58 & 115 & 249 & 242 \\
Qwen-2.5-1.5B & Qwen-2.5-1.5B & Both: Deterministic & 67.58 & \textcolor{darkred}{1.63 $\downarrow$} & 0 & 0.00 & 0 & 0 & 0 \\
Qwen-2.5-1.5B & Qwen-2.5-1.5B & Both: Exploratory & 68.52 & \textcolor{darkred}{0.69 $\downarrow$} & 0.75 & 0.70 & 133 & 296 & 275 \\
Qwen-2.5-1.5B & Qwen-2.5-1.5B & Both: Det. \& Exp. & 69.88 & \textcolor{darkgreen}{0.67 $\uparrow$} & 0.60 & 0.61 & 101 & 262 & 252 \\
\midrule
Qwen-2.5-3B & Qwen-2.5-3B & Both: Default & 85.41 & \textcolor{darkgreen}{1.88 $\uparrow$} & 0.29 & 0.29 & 53 & 114 & 111 \\
Qwen-2.5-3B & Qwen-2.5-3B & Both: Deterministic & 84.13 & \textcolor{darkgreen}{0.60 $\uparrow$} & 0 & 0.00 & 0 & 0 & 0 \\
Qwen-2.5-3B & Qwen-2.5-3B & Both: Exploratory & 84.64 & \textcolor{darkgreen}{1.11 $\uparrow$} & 0.30 & 0.27 & 56 & 116 & 109 \\
Qwen-2.5-3B & Qwen-2.5-3B & Both: Det. \& Exp. & 83.70 & \textcolor{darkgreen}{0.17 $\uparrow$} & 0.28 & 0.23 & 70 & 79 & 73 \\
\midrule
Qwen-2.5-7B & Qwen-2.5-7B & Both: Default & 91.55 & \textcolor{darkgreen}{4.33 $\uparrow$} & 0.11 & 0.11 & 29 & 46 & 45 \\
Qwen-2.5-7B & Qwen-2.5-7B & Both: Deterministic & 91.21 & \textcolor{darkgreen}{3.99 $\uparrow$} & 0 & 0.00 & 0 & 0 & 0 \\
Qwen-2.5-7B & Qwen-2.5-7B & Both: Exploratory & 91.64 & \textcolor{darkgreen}{4.42 $\uparrow$} & 0.12 & 0.11 & 23 & 53 & 51 \\
Qwen-2.5-7B & Qwen-2.5-7B & Both: Det. \& Exp. & 92.06 & \textcolor{darkgreen}{4.84 $\uparrow$} & 0.13 & 0.12 & 30 & 48 & 43 \\
\midrule
Qwen-2.5-14B & Qwen-2.5-14B & Both: Default & 94.54 & \textcolor{darkgreen}{4.27 $\uparrow$} & 0.06 & 0.05 & 13 & 24 & 24 \\
Qwen-2.5-14B & Qwen-2.5-14B & Both: Deterministic & 94.37 & \textcolor{darkgreen}{4.10 $\uparrow$} & 0 & 0.00 & 0 & 0 & 0 \\
Qwen-2.5-14B & Qwen-2.5-14B & Both: Exploratory & 93.77 & \textcolor{darkgreen}{3.50 $\uparrow$} & 0.06 & 0.07 & 23 & 24 & 24 \\
Qwen-2.5-14B & Qwen-2.5-14B & Both: Det. \& Exp. & 94.71 & \textcolor{darkgreen}{4.44 $\uparrow$} & 0.06 & 0.06 & 11 & 22 & 21 \\
\midrule
Qwen-2.5-32B & Qwen-2.5-32B & Both: Default & 98.53 & \textcolor{darkgreen}{3.25 $\uparrow$} & 0.02 & 0.06 & 10 & 14 & 13 \\
Qwen-2.5-32B & Qwen-2.5-32B & Both: Deterministic & 98.36 & \textcolor{darkgreen}{3.08 $\uparrow$} & 0 & 0.00 & 0 & 0 & 0 \\
Qwen-2.5-32B & Qwen-2.5-32B & Both: Exploratory & 98.53 & \textcolor{darkgreen}{3.25 $\uparrow$} & 0.02 & 0.06 & 8 & 14 & 14 \\
Qwen-2.5-32B & Qwen-2.5-32B & Both: Det. \& Exp. & 98.36 & \textcolor{darkgreen}{3.08 $\uparrow$} & 0.02 & 0.04 & 9 & 10 & 8 \\
\midrule
Phi-mini-3.8B & Phi-mini-3.8B & Both: Default & 90.10 & \textcolor{darkgreen}{5.37 $\uparrow$} & 0.24 & 0.34 & 42 & 75 & 66 \\
Phi-mini-3.8B & Phi-mini-3.8B & Both: Deterministic & 88.91 & \textcolor{darkgreen}{4.18 $\uparrow$} & 0 & 0.00 & 0 & 0 & 0 \\
Phi-mini-3.8B & Phi-mini-3.8B & Both: Exploratory & 87.03 & \textcolor{darkgreen}{2.30 $\uparrow$} & 0.31 & 0.40 & 58 & 107 & 100 \\
Phi-mini-3.8B & Phi-mini-3.8B & Both: Det. \& Exp. & 88.05 & \textcolor{darkgreen}{3.32 $\uparrow$} & 0.23 & 0.31 & 46 & 69 & 62 \\
\midrule
Llama-3.1-3B & Llama-3.1-3B & Both: Default & 75.77 & \textcolor{darkgreen}{2.65 $\uparrow$} & 0.46 & 0.37 & 93 & 130 & 126 \\
Llama-3.1-3B & Llama-3.1-3B & Both: Deterministic & 74.66 & \textcolor{darkgreen}{1.54 $\uparrow$} & 0 & 0.00 & 0 & 0 & 0 \\
Llama-3.1-3B & Llama-3.1-3B & Both: Exploratory & 76.19 & \textcolor{darkgreen}{3.07 $\uparrow$} & 0.50 & 0.43 & 89 & 166 & 149 \\
Llama-3.1-3B & Llama-3.1-3B & Both: Det. \& Exp. & 75.60 & \textcolor{darkgreen}{2.48 $\uparrow$} & 0.45 & 0.34 & 108 & 129 & 124 \\
\midrule
Llama-3.1-8B & Llama-3.1-8B & Both: Default & 87.20 & \textcolor{darkgreen}{9.55 $\uparrow$} & 0.26 & 0.30 & 45 & 91 & 88 \\
Llama-3.1-8B & Llama-3.1-8B & Both: Deterministic & 85.75 & \textcolor{darkgreen}{8.10 $\uparrow$} & 0 & 0.00 & 0 & 0 & 0 \\
Llama-3.1-8B & Llama-3.1-8B & Both: Exploratory & 85.07 & \textcolor{darkgreen}{7.42 $\uparrow$} & 0.28 & 0.32 & 58 & 96 & 94 \\
Llama-3.1-8B & Llama-3.1-8B & Both: Det. \& Exp. & 86.86 & \textcolor{darkgreen}{9.21 $\uparrow$} & 0.23 & 0.27 & 56 & 84 & 80 \\
\midrule
Mistral-7B & Mistral-7B & Both: Default & 70.48 & \textcolor{darkgreen}{1.71 $\uparrow$} & 0.51 & 0.37 & 99 & 145 & 137 \\
Mistral-7B & Mistral-7B & Both: Deterministic & 68.26 & \textcolor{darkred}{0.51 $\downarrow$} & 0 & 0.00 & 0 & 0 & 0 \\
Mistral-7B & Mistral-7B & Both: Exploratory & 72.78 & \textcolor{darkgreen}{4.01 $\uparrow$} & 0.58 & 0.44 & 106 & 185 & 177 \\
Mistral-7B & Mistral-7B & Both: Det. \& Exp. & 70.82 & \textcolor{darkgreen}{2.05 $\uparrow$} & 0.50 & 0.34 & 84 & 151 & 142 \\
\bottomrule
\end{tabular}
\end{adjustbox}
\caption{Comparative Analysis of Language Model Performance in Multi-Agent Debate Settings on the \textbf{ARC-Challenge} Dataset. This table showcases the impact of different \textbf{Agent Settings} (controlling temperature and top\_p parameters like Default, Deterministic, Exploratory, and a combination) on the \textbf{MAD Accuracy} of various language models. The \textbf{$\Delta$} column quantifies the \textbf{improvement (or decline) over the single base model performance} shown in parentheses next to each model name. Further metrics include average \textbf{Debate Rounds}, normalized \textbf{Sycophancy} (per 1172 data points), and transitions between correct (C) and incorrect (I) states (C$\rightarrow$I, I$\rightarrow$C), highlighting the nuanced effects of debate dynamics.}
\label{tab:arc_challenge_debate_performance}
\end{table*}

\begin{table*}[htbp]
\centering
\begin{adjustbox}{width=\textwidth,center} 
\sisetup{round-mode=places,round-precision=2} 
\begin{tabular}{@{}ccccS[table-format=2.2]S[table-format=2.2]S[table-format=1.2]S[table-format=4.2]cccS[table-format=1.2]@{}} 
\toprule
\textbf{Agent 1} & \textbf{Agent 2} & \textbf{Agent Settings} & \textbf{MAD Accuracy} & \multicolumn{1}{c}{\textbf{$\Delta_1$}} & \multicolumn{1}{c}{\textbf{$\Delta_2$}} & \textbf{Debate} & \multicolumn{1}{c}{\textbf{Sycophancy}} & \textbf{C$\rightarrow$I} & \textbf{I$\rightarrow$C} & \textbf{Debate} \\
 & & & \textbf{(ARC-Challenge)} &  &  & \textbf{Rounds} & \multicolumn{1}{c}{\textbf{(Avg / 1172)}} & & & \textbf{Helped} \\
 & & & &  &  & \textbf{(Avg)} & & & & \textbf{(Overall)} \\
\midrule
Qwen-2.5-0.5B & Qwen-2.5-1.5B & Both: Default & 58.28 & \textcolor{darkgreen}{20.31 $\uparrow$} & \textcolor{darkred}{10.93 $\downarrow$} & 1.27 & 0.97 & 193 & 401 & 369 \\
Qwen-2.5-0.5B & Qwen-2.5-1.5B & Both: Deterministic & 63.57 & \textcolor{darkgreen}{25.60 $\uparrow$} & \textcolor{darkred}{5.64 $\downarrow$} & 1.09 & 0.81 & 169 & 375 & 357 \\
Qwen-2.5-0.5B & Qwen-2.5-1.5B & Both: Exploratory & 55.80 & \textcolor{darkgreen}{17.83 $\uparrow$} & \textcolor{darkred}{13.41 $\downarrow$} & 1.46 & 1.07 & 211 & 418 & 368 \\
Qwen-2.5-0.5B & Qwen-2.5-1.5B & Both: Det. \& Exp. & 60.32 & \textcolor{darkgreen}{22.35 $\uparrow$} & \textcolor{darkred}{8.89 $\downarrow$} & 1.10 & 0.94 & 181 & 397 & 360 \\
Qwen-2.5-0.5B & Qwen-2.5-1.5B & Both: Exp. \& Det. & 61.43 & \textcolor{darkgreen}{23.46 $\uparrow$} & \textcolor{darkred}{7.78 $\downarrow$} & 1.39 & 0.95 & 197 & 409 & 387 \\
\midrule
Qwen-2.5-1.5B & Llama-3.1-3B & Both: Default & 72.35 & \textcolor{darkgreen}{3.14 $\uparrow$} & \textcolor{darkred}{0.77 $\downarrow$} & 0.67 & 0.66 & 143 & 216 & 207 \\
Qwen-2.5-1.5B & Llama-3.1-3B & Both: Deterministic & 74.91 & \textcolor{darkgreen}{5.70 $\uparrow$} & \textcolor{darkgreen}{1.79 $\uparrow$} & 0.51 & 0.51 & 135 & 191 & 185 \\
Qwen-2.5-1.5B & Llama-3.1-3B & Both: Exploratory & 73.12 & \textcolor{darkgreen}{3.91 $\uparrow$} & 0.00 & 0.78 & 0.78 & 153 & 281 & 265 \\
Qwen-2.5-1.5B & Llama-3.1-3B & Both: Det. \& Exp. & 76.02 & \textcolor{darkgreen}{6.81 $\uparrow$} & \textcolor{darkgreen}{2.90 $\uparrow$} & 0.60 & 0.66 & 127 & 219 & 205 \\
Qwen-2.5-1.5B & Llama-3.1-3B & Both: Exp. \& Det. & 74.15 & \textcolor{darkgreen}{4.94 $\uparrow$} & \textcolor{darkgreen}{1.03 $\uparrow$} & 0.71 & 0.61 & 135 & 291 & 274 \\
\midrule
Qwen-2.5-3B & Phi-mini-3.8B & Both: Default & 87.97 & \textcolor{darkgreen}{4.44 $\uparrow$} & \textcolor{darkgreen}{3.24 $\uparrow$} & 0.32 & 0.31 & 59 & 133 & 130 \\
Qwen-2.5-3B & Phi-mini-3.8B & Both: Deterministic & 88.57 & \textcolor{darkgreen}{5.04 $\uparrow$} & \textcolor{darkgreen}{3.84 $\uparrow$} & 0.31 & 0.25 & 58 & 110 & 107 \\
Qwen-2.5-3B & Phi-mini-3.8B & Both: Exploratory & 87.03 & \textcolor{darkgreen}{3.50 $\uparrow$} & \textcolor{darkgreen}{2.30 $\uparrow$} & 0.38 & 0.37 & 72 & 173 & 160 \\
Qwen-2.5-3B & Phi-mini-3.8B & Both: Det. \& Exp. & 87.80 & \textcolor{darkgreen}{4.27 $\uparrow$} & \textcolor{darkgreen}{3.07 $\uparrow$} & 0.33 & 0.30 & 59 & 141 & 139 \\
Qwen-2.5-3B & Phi-mini-3.8B & Both: Exp. \& Det. & 89.85 & \textcolor{darkgreen}{6.32 $\uparrow$} & \textcolor{darkgreen}{5.12 $\uparrow$} & 0.34 & 0.30 & 50 & 143 & 137 \\
\midrule
Qwen-2.5-1.5B & Qwen-2.5-3B & Both: Default & 82.25 & \textcolor{darkgreen}{13.04 $\uparrow$} & \textcolor{darkred}{1.28 $\downarrow$} & 0.51 & 0.45 & 80 & 247 & 243 \\
Qwen-2.5-1.5B & Qwen-2.5-3B & Both: Deterministic & 82.59 & \textcolor{darkgreen}{13.38 $\uparrow$} & \textcolor{darkred}{0.94 $\downarrow$} & 0.42 & 0.40 & 80 & 205 & 200 \\
Qwen-2.5-1.5B & Qwen-2.5-3B & Both: Exploratory & 81.91 & \textcolor{darkgreen}{12.70 $\uparrow$} & \textcolor{darkred}{1.62 $\downarrow$} & 0.66 & 0.56 & 94 & 317 & 310 \\
Qwen-2.5-1.5B & Qwen-2.5-3B & Both: Det. \& Exp. & 83.45 & \textcolor{darkgreen}{14.24 $\uparrow$} & \textcolor{darkred}{0.08 $\downarrow$} & 0.47 & 0.46 & 66 & 227 & 219 \\
Qwen-2.5-1.5B & Qwen-2.5-3B & Both: Exp. \& Det. & 83.62 & \textcolor{darkgreen}{14.41 $\uparrow$} & \textcolor{darkgreen}{0.09 $\uparrow$} & 0.62 & 0.51 & 67 & 328 & 320 \\
\midrule
Llama-3.1-3B & Llama-3.1-8B & Both: Default & 81.66 & \textcolor{darkgreen}{8.54 $\uparrow$} & \textcolor{darkgreen}{4.01 $\uparrow$} & 0.47 & 0.41 & 114 & 141 & 133 \\
Llama-3.1-3B & Llama-3.1-8B & Both: Deterministic & 80.46 & \textcolor{darkgreen}{7.34 $\uparrow$} & \textcolor{darkgreen}{2.81 $\uparrow$} & 0.51 & 0.36 & 120 & 135 & 124 \\
Llama-3.1-3B & Llama-3.1-8B & Both: Exploratory & 75.68 & \textcolor{darkgreen}{2.56 $\uparrow$} & \textcolor{darkred}{1.97 $\downarrow$} & 0.48 & 0.43 & 107 & 160 & 151 \\
Llama-3.1-3B & Llama-3.1-8B & Both: Det. \& Exp. & 80.12 & \textcolor{darkgreen}{7.00 $\uparrow$} & \textcolor{darkgreen}{2.47 $\uparrow$} & 0.46 & 0.37 & 117 & 138 & 132 \\
Llama-3.1-3B & Llama-3.1-8B & Both: Exp. \& Det. & 80.97 & \textcolor{darkgreen}{7.85 $\uparrow$} & \textcolor{darkgreen}{3.32 $\uparrow$} & 0.49 & 0.43 & 109 & 159 & 154 \\
\midrule
Qwen-2.5-7B & Qwen-2.5-14B & Both: Default & 93.43 & \textcolor{darkgreen}{6.21 $\uparrow$} & \textcolor{darkgreen}{3.16 $\uparrow$} & 0.14 & 0.11 & 35 & 54 & 53 \\
Qwen-2.5-7B & Qwen-2.5-14B & Both: Deterministic & 93.60 & \textcolor{darkgreen}{6.38 $\uparrow$} & \textcolor{darkgreen}{3.33 $\uparrow$} & 0.13 & 0.10 & 24 & 59 & 58 \\
Qwen-2.5-7B & Qwen-2.5-14B & Both: Exploratory & 94.45 & \textcolor{darkgreen}{7.23 $\uparrow$} & \textcolor{darkgreen}{4.18 $\uparrow$} & 0.15 & 0.14 & 27 & 67 & 65 \\
Qwen-2.5-7B & Qwen-2.5-14B & Both: Det. \& Exp. & 93.00 & \textcolor{darkgreen}{5.78 $\uparrow$} & \textcolor{darkgreen}{2.73 $\uparrow$} & 0.16 & 0.13 & 37 & 50 & 49 \\
Qwen-2.5-7B & Qwen-2.5-14B & Both: Exp. \& Det. & 93.77 & \textcolor{darkgreen}{6.55 $\uparrow$} & \textcolor{darkgreen}{3.50 $\uparrow$} & 0.15 & 0.12 & 26 & 58 & 58 \\
\bottomrule
\end{tabular}
\end{adjustbox}
\caption{Comparative Analysis of Mixed Model Pairs in Multi-Agent Debate Settings on the \textbf{ARC-Challenge} Dataset. This table showcases different model combinations and the impact of various \textbf{Agent Settings} on accuracy. \textbf{$\Delta_1$} represents the improvement over the lower-capability model (the first agent), while \textbf{$\Delta_2$} represents the improvement or decline relative to the higher-capability model (the second agent). Values in parentheses next to each model name indicate the single-agent baseline performance. The table also shows average \textbf{Debate Rounds}, normalized \textbf{Sycophancy} (per 1172 data points), and transitions between correct (C) and incorrect (I) states, demonstrating how mixed-capability agents interact in debate scenarios.}
\label{tab:mixed_model_arc_challenge}
\end{table*}

\begin{table*}[htbp]
\centering
\begin{adjustbox}{width=\textwidth,center} 
\sisetup{round-mode=places,round-precision=2} 
\begin{tabular}{@{}ccccS[table-format=2.2]S[table-format=2.2]S[table-format=1.2]S[table-format=4.2]cccS[table-format=1.2]@{}} 
\toprule
\textbf{Agent 1} & \textbf{Agent 2} & \textbf{Agent 3} & \textbf{Agent Settings} & \multicolumn{1}{c}{\textbf{Accuracy}} & \multicolumn{1}{c}{\textbf{$\Delta$}} & \textbf{Debate} & \multicolumn{1}{c}{\textbf{Sycophancy}} & \textbf{C$\rightarrow$I} & \textbf{I$\rightarrow$C} & \textbf{Debate} \\
 & & & & & & \textbf{Rounds} & \multicolumn{1}{c}{\textbf{(Avg / 1172)}} & & & \textbf{Helped} \\
 & & & & & & \textbf{(Avg)} & & & & \textbf{(Overall)} \\
\midrule
Qwen-2.5-0.5B & Qwen-2.5-0.5B & Qwen-2.5-0.5B & Default & 35.15 & \textcolor{darkred}{2.82 $\downarrow$} & 2.54 & 3.14 & 535 & 484 & 283 \\
Qwen-2.5-0.5B & Qwen-2.5-0.5B & Qwen-2.5-0.5B & Deterministic & 40.78 & \textcolor{darkgreen}{2.81 $\uparrow$} & 0.00 & 0.00 & 0 & 0 & 0 \\
Qwen-2.5-0.5B & Qwen-2.5-0.5B & Qwen-2.5-0.5B & Exploratory & 35.32 & \textcolor{darkred}{2.65 $\downarrow$} & 3.12 & 3.54 & 587 & 528 & 303 \\
Qwen-2.5-0.5B & Qwen-2.5-0.5B & Qwen-2.5-0.5B & 1 Det. \& 2 Exp. & 37.20 & \textcolor{darkred}{0.77 $\downarrow$} & 2.78 & 3.19 & 523 & 503 & 306 \\
Qwen-2.5-0.5B & Qwen-2.5-0.5B & Qwen-2.5-0.5B & 2 Det. \& 1 Exp. & 38.23 & \textcolor{darkgreen}{0.26 $\uparrow$} & 1.49 & 1.75 & 404 & 353 & 219 \\
\midrule
Qwen-2.5-1.5B & Qwen-2.5-1.5B & Qwen-2.5-1.5B & Default & 72.53 & \textcolor{darkgreen}{3.32 $\uparrow$} & 0.98 & 1.29 & 206 & 454 & 343 \\
Qwen-2.5-1.5B & Qwen-2.5-1.5B & Qwen-2.5-1.5B & Deterministic & 67.58 & \textcolor{darkred}{1.63 $\downarrow$} & 0.00 & 0.00 & 0 & 0 & 0 \\
Qwen-2.5-1.5B & Qwen-2.5-1.5B & Qwen-2.5-1.5B & Exploratory & 72.10 & \textcolor{darkgreen}{2.89 $\uparrow$} & 1.37 & 1.85 & 235 & 611 & 433 \\
Qwen-2.5-1.5B & Qwen-2.5-1.5B & Qwen-2.5-1.5B & 1 Det. \& 2 Exp. & 71.93 & \textcolor{darkgreen}{2.72 $\uparrow$} & 1.12 & 1.53 & 229 & 520 & 386 \\
Qwen-2.5-1.5B & Qwen-2.5-1.5B & Qwen-2.5-1.5B & 2 Det. \& 1 Exp. & 70.82 & \textcolor{darkgreen}{1.61 $\uparrow$} & 0.63 & 0.93 & 163 & 345 & 245 \\
\midrule
Qwen-2.5-3B & Qwen-2.5-3B & Qwen-2.5-3B & Default & 85.75 & \textcolor{darkgreen}{2.22 $\uparrow$} & 0.43 & 0.43 & 79 & 197 & 156 \\
Qwen-2.5-3B & Qwen-2.5-3B & Qwen-2.5-3B & Deterministic & 84.13 & \textcolor{darkgreen}{0.60 $\uparrow$} & 0.00 & 0.00 & 0 & 0 & 0 \\
Qwen-2.5-3B & Qwen-2.5-3B & Qwen-2.5-3B & Exploratory & 86.26 & \textcolor{darkgreen}{2.73 $\uparrow$} & 0.50 & 0.57 & 96 & 229 & 167 \\
Qwen-2.5-3B & Qwen-2.5-3B & Qwen-2.5-3B & 1 Det. \& 2 Exp. & 86.26 & \textcolor{darkgreen}{2.73 $\uparrow$} & 0.51 & 0.48 & 106 & 193 & 149 \\
Qwen-2.5-3B & Qwen-2.5-3B & Qwen-2.5-3B & 2 Det. \& 1 Exp. & 84.73 & \textcolor{darkgreen}{1.20 $\uparrow$} & 0.33 & 0.31 & 71 & 131 & 101 \\
\midrule
Qwen-2.5-7B & Qwen-2.5-7B & Qwen-2.5-7B & Default & 91.81 & \textcolor{darkgreen}{4.59 $\uparrow$} & 0.19 & 0.22 & 56 & 84 & 66 \\
Qwen-2.5-7B & Qwen-2.5-7B & Qwen-2.5-7B & Deterministic & 90.61 & \textcolor{darkgreen}{3.39 $\uparrow$} & 0.00 & 0.00 & 0 & 0 & 0 \\
Qwen-2.5-7B & Qwen-2.5-7B & Qwen-2.5-7B & Exploratory & 91.72 & \textcolor{darkgreen}{4.50 $\uparrow$} & 0.23 & 0.29 & 66 & 85 & 65 \\
Qwen-2.5-7B & Qwen-2.5-7B & Qwen-2.5-7B & 1 Det. \& 2 Exp. & 91.04 & \textcolor{darkgreen}{3.82 $\uparrow$} & 0.22 & 0.24 & 60 & 80 & 68 \\
Qwen-2.5-7B & Qwen-2.5-7B & Qwen-2.5-7B & 2 Det. \& 1 Exp. & 91.30 & \textcolor{darkgreen}{4.08 $\uparrow$} & 0.14 & 0.15 & 40 & 57 & 40 \\
\midrule
Qwen-2.5-14B & Qwen-2.5-14B & Qwen-2.5-14B & Default & 94.20 & \textcolor{darkgreen}{3.93 $\uparrow$} & 0.12 & 0.13 & 27 & 54 & 45 \\
Qwen-2.5-14B & Qwen-2.5-14B & Qwen-2.5-14B & Deterministic & 94.37 & \textcolor{darkgreen}{4.10 $\uparrow$} & 0.00 & 0.00 & 0 & 0 & 0 \\
Qwen-2.5-14B & Qwen-2.5-14B & Qwen-2.5-14B & Exploratory & 94.80 & \textcolor{darkgreen}{4.53 $\uparrow$} & 0.10 & 0.12 & 28 & 50 & 39 \\
Qwen-2.5-14B & Qwen-2.5-14B & Qwen-2.5-14B & 1 Det. \& 2 Exp. & 94.54 & \textcolor{darkgreen}{4.27 $\uparrow$} & 0.09 & 0.09 & 22 & 41 & 33 \\
Qwen-2.5-14B & Qwen-2.5-14B & Qwen-2.5-14B & 2 Det. \& 1 Exp. & 94.71 & \textcolor{darkgreen}{4.44 $\uparrow$} & 0.06 & 0.06 & 10 & 32 & 26 \\
\midrule
Qwen-2.5-32B & Qwen-2.5-32B & Qwen-2.5-32B & Default & 95.82 & \textcolor{darkgreen}{0.54 $\uparrow$} & 0.07 & 0.11 & 22 & 36 & 28 \\
Qwen-2.5-32B & Qwen-2.5-32B & Qwen-2.5-32B & Deterministic & 95.73 & \textcolor{darkgreen}{0.45 $\uparrow$} & 0.00 & 0.00 & 0 & 0 & 0 \\
Qwen-2.5-32B & Qwen-2.5-32B & Qwen-2.5-32B & Exploratory & 95.56 & \textcolor{darkgreen}{0.28 $\uparrow$} & 0.08 & 0.12 & 28 & 35 & 32 \\
Qwen-2.5-32B & Qwen-2.5-32B & Qwen-2.5-32B & 1 Det. \& 2 Exp. & 95.56 & \textcolor{darkgreen}{0.28 $\uparrow$} & 0.07 & 0.10 & 30 & 29 & 25 \\
Qwen-2.5-32B & Qwen-2.5-32B & Qwen-2.5-32B & 2 Det. \& 1 Exp. & 95.99 & \textcolor{darkgreen}{0.71 $\uparrow$} & 0.03 & 0.04 & 13 & 18 & 14 \\
\midrule
Phi-mini-3.8B & Phi-mini-3.8B & Phi-mini-3.8B & Default & 88.91 & \textcolor{darkgreen}{4.18 $\uparrow$} & 0.35 & 0.61 & 69 & 130 & 104 \\
Phi-mini-3.8B & Phi-mini-3.8B & Phi-mini-3.8B & Deterministic & 88.91 & \textcolor{darkgreen}{4.18 $\uparrow$} & 0.00 & 0.00 & 0 & 0 & 0 \\
Phi-mini-3.8B & Phi-mini-3.8B & Phi-mini-3.8B & Exploratory & 88.74 & \textcolor{darkgreen}{4.01 $\uparrow$} & 0.50 & 0.83 & 85 & 196 & 151 \\
Phi-mini-3.8B & Phi-mini-3.8B & Phi-mini-3.8B & 1 Det. \& 2 Exp. & 88.74 & \textcolor{darkgreen}{4.01 $\uparrow$} & 0.37 & 0.61 & 74 & 155 & 121 \\
Phi-mini-3.8B & Phi-mini-3.8B & Phi-mini-3.8B & 2 Det. \& 1 Exp. & 89.08 & \textcolor{darkgreen}{4.35 $\uparrow$} & 0.30 & 0.52 & 54 & 109 & 81 \\
\midrule
Llama-3.1-3B & Llama-3.1-3B & Llama-3.1-3B & Default & 75.77 & \textcolor{darkgreen}{2.65 $\uparrow$} & 0.81 & 0.80 & 177 & 244 & 190 \\
Llama-3.1-3B & Llama-3.1-3B & Llama-3.1-3B & Deterministic & 74.83 & \textcolor{darkgreen}{1.71 $\uparrow$} & 0.00 & 0.00 & 0 & 0 & 0 \\
Llama-3.1-3B & Llama-3.1-3B & Llama-3.1-3B & Exploratory & 75.51 & \textcolor{darkgreen}{2.39 $\uparrow$} & 0.90 & 1.00 & 196 & 303 & 210 \\
Llama-3.1-3B & Llama-3.1-3B & Llama-3.1-3B & 1 Det. \& 2 Exp. & 75.17 & \textcolor{darkgreen}{2.05 $\uparrow$} & 0.99 & 0.91 & 223 & 262 & 192 \\
Llama-3.1-3B & Llama-3.1-3B & Llama-3.1-3B & 2 Det. \& 1 Exp. & 75.26 & \textcolor{darkgreen}{2.14 $\uparrow$} & 0.53 & 0.43 & 118 & 162 & 117 \\
\midrule
Mistral-7B & Mistral-7B & Mistral-7B & Default & 70.73 & \textcolor{darkgreen}{1.96 $\uparrow$} & 0.97 & 0.94 & 213 & 292 & 207 \\
Mistral-7B & Mistral-7B & Mistral-7B & Deterministic & 68.26 & \textcolor{darkred}{0.51 $\downarrow$} & 0.00 & 0.00 & 0 & 0 & 0 \\
Mistral-7B & Mistral-7B & Mistral-7B & Exploratory & 71.67 & \textcolor{darkgreen}{2.90 $\uparrow$} & 1.14 & 1.20 & 232 & 360 & 249 \\
Mistral-7B & Mistral-7B & Mistral-7B & 1 Det. \& 2 Exp. & 71.25 & \textcolor{darkgreen}{2.48 $\uparrow$} & 1.03 & 1.03 & 209 & 317 & 227 \\
Mistral-7B & Mistral-7B & Mistral-7B & 2 Det. \& 1 Exp. & 70.48 & \textcolor{darkgreen}{1.71 $\uparrow$} & 0.62 & 0.66 & 142 & 214 & 136 \\
\midrule
Llama-3.1-8B & Llama-3.1-8B & Llama-3.1-8B & Default & 87.46 & \textcolor{darkgreen}{9.81 $\uparrow$} & 0.40 & 0.56 & 98 & 145 & 107 \\
Llama-3.1-8B & Llama-3.1-8B & Llama-3.1-8B & Deterministic & 86.43 & \textcolor{darkgreen}{8.78 $\uparrow$} & 0.00 & 0.00 & 0 & 0 & 0 \\
Llama-3.1-8B & Llama-3.1-8B & Llama-3.1-8B & Exploratory & 86.01 & \textcolor{darkgreen}{8.36 $\uparrow$} & 0.52 & 0.77 & 127 & 187 & 150 \\
Llama-3.1-8B & Llama-3.1-8B & Llama-3.1-8B & 1 Det. \& 2 Exp. & 86.69 & \textcolor{darkgreen}{9.04 $\uparrow$} & 0.50 & 0.72 & 114 & 174 & 128 \\
Llama-3.1-8B & Llama-3.1-8B & Llama-3.1-8B & 2 Det. \& 1 Exp. & 85.67 & \textcolor{darkgreen}{8.02 $\uparrow$} & 0.30 & 0.46 & 115 & 119 & 73 \\
\bottomrule
\end{tabular}
\end{adjustbox}
\caption{Comparative Analysis of Language Model Performance in Multi-Agent Debate Settings on the \textbf{ARC-Challenge} Dataset. This table showcases the impact of different \textbf{Agent Settings} (controlling temperature and top\_p parameters) on the \textbf{Accuracy} of various language models in a three-agent configuration. The \textbf{$\Delta$} column quantifies the \textbf{improvement (or decline) over the single base model performance} (shown in parentheses after model names). Further metrics include average \textbf{Debate Rounds}, normalized \textbf{Sycophancy} (per 1172 data points), and transitions between correct (C) and incorrect (I) states (C$\rightarrow$I, I$\rightarrow$C), highlighting the nuanced effects of debate dynamics.}
\label{tab:arc_challenge_performance}
\end{table*}

\begin{table*}[htbp]
\centering
\begin{adjustbox}{width=\textwidth,center} 
\sisetup{round-mode=places,round-precision=2} 
\begin{tabular}{@{}cccS[table-format=2.2]S[table-format=2.2]S[table-format=1.2]S[table-format=4.2]cccS[table-format=1.2]@{}} 
\toprule
\textbf{Agent 1} & \textbf{Agent 2} & \textbf{Agent 3} & \multicolumn{1}{c}{\textbf{Accuracy}} & \multicolumn{1}{c}{\textbf{$\Delta$}} & \textbf{Debate} & \multicolumn{1}{c}{\textbf{Sycophancy}} & \textbf{C$\rightarrow$I} & \textbf{I$\rightarrow$C} & \textbf{Debate} \\
 & & & & & \textbf{Rounds} & \multicolumn{1}{c}{\textbf{(Avg / 1172)}} & & & \textbf{Helped} \\
 & & & & & \textbf{(Avg)} & & & & \textbf{(Overall)} \\
\midrule
Qwen-2.5-0.5B & Qwen-2.5-1.5B & Qwen-2.5-3B & 82.59 & \textcolor{darkred}{0.94 $\downarrow$} & 1.41 & 1.40 & 148 & 820 & 629 \\
Qwen-2.5-0.5B & Qwen-2.5-1.5B & Llama-3.1-3B & 68.00 & \textcolor{darkred}{5.12 $\downarrow$} & 1.66 & 1.85 & 311 & 641 & 489 \\
Qwen-2.5-0.5B & Qwen-2.5-1.5B & Phi-mini-3.8B & 82.76 & \textcolor{darkred}{1.97 $\downarrow$} & 1.48 & 1.60 & 170 & 804 & 621 \\
Qwen-2.5-0.5B & Qwen-2.5-3B & Llama-3.1-3B & 79.69 & \textcolor{darkred}{3.84 $\downarrow$} & 1.62 & 1.50 & 208 & 699 & 581 \\
Qwen-2.5-0.5B & Qwen-2.5-3B & Phi-mini-3.8B & 86.95 & \textcolor{darkgreen}{2.22 $\uparrow$} & 1.34 & 1.23 & 133 & 722 & 631 \\
Qwen-2.5-0.5B & Llama-3.1-3B & Phi-mini-3.8B & 78.41 & \textcolor{darkred}{6.32 $\downarrow$} & 1.54 & 1.72 & 238 & 683 & 559 \\
Qwen-2.5-1.5B & Qwen-2.5-3B & Llama-3.1-3B & 82.34 & \textcolor{darkred}{1.19 $\downarrow$} & 0.98 & 1.10 & 180 & 447 & 358 \\
Qwen-2.5-1.5B & Qwen-2.5-3B & Phi-mini-3.8B & 87.37 & \textcolor{darkgreen}{2.64 $\uparrow$} & 0.71 & 0.81 & 105 & 423 & 358 \\
Qwen-2.5-1.5B & Llama-3.1-3B & Phi-mini-3.8B & 81.74 & \textcolor{darkred}{3.00 $\downarrow$} & 0.93 & 1.19 & 195 & 412 & 341 \\
Qwen-2.5-3B & Phi-mini-3.8B & Llama-3.1-3B & 85.67 & \textcolor{darkgreen}{2.14 $\uparrow$} & 0.84 & 0.89 & 143 & 319 & 244 \\
\midrule
Qwen-2.5-3B & Qwen-2.5-3B & Phi-mini-3.8B & 87.88 & \textcolor{darkgreen}{3.15 $\uparrow$} & 0.50 & 0.52 & 110 & 225 & 170 \\
Qwen-2.5-3B & Phi-mini-3.8B & Phi-mini-3.8B & 89.33 & \textcolor{darkgreen}{4.60 $\uparrow$} & 0.52 & 0.61 & 81 & 214 & 174 \\
Qwen-2.5-0.5B & Qwen-2.5-1.5B & Qwen-2.5-1.5B & 69.80 & \textcolor{darkgreen}{0.59 $\uparrow$} & 1.66 & 1.77 & 231 & 686 & 523 \\
Qwen-2.5-0.5B & Qwen-2.5-0.5B & Qwen-2.5-1.5B & 55.97 & \textcolor{darkred}{13.24 $\downarrow$} & 2.33 & 2.69 & 393 & 680 & 451 \\
\bottomrule
\end{tabular}
\end{adjustbox}
\caption{Analysis of Mixed-Model Configurations in Multi-Agent Debate Settings on the \textbf{ARC-Challenge} Dataset. This table examines various heterogeneous model combinations in three-agent debate setups. The \textbf{$\Delta$} column quantifies the \textbf{improvement (or decline) compared to the best single model performance} among the three agents used in each configuration. All agent combinations use the default settings for temperature and top\_p. Metrics include average \textbf{Debate Rounds}, normalized \textbf{Sycophancy} (per 1172 data points), and transitions between correct (C) and incorrect (I) states (C$\rightarrow$I, I$\rightarrow$C). Results demonstrate that certain model combinations can achieve higher accuracy than their constituent models when debating together.}
\label{tab:mixed_model_performance}
\end{table*}


\begin{table*}[htbp]
\centering
\begin{adjustbox}{width=\textwidth,center} 
\sisetup{round-mode=places,round-precision=2} 
\begin{tabular}{@{}ccccS[table-format=2.2]S[table-format=3.2]S[table-format=1.2]S[table-format=4.2]cccS[table-format=1.2]@{}} 
\toprule
\textbf{Agent 1} & \textbf{Agent 2} & \textbf{Agent Settings} & \textbf{Accuracy} & \multicolumn{1}{c}{\textbf{$\Delta$}} & \textbf{Debate} & \multicolumn{1}{c}{\textbf{Sycophancy}} & \textbf{C$\rightarrow$I} & \textbf{I$\rightarrow$C} & \textbf{Debate} \\
 & & & & & \textbf{Rounds} & \multicolumn{1}{c}{\textbf{(Avg / 1221)}} & & & \textbf{Helped} \\
 & & & & & \textbf{(Avg)} & & & & \textbf{(Overall)} \\
\midrule
Qwen-2.5-0.5B & Qwen-2.5-0.5B & Both: Default & 39.80 & \textcolor{darkgreen}{3.31 $\uparrow$} & 1.47 & 1.11 & 239 & 306 & 240 \\
Qwen-2.5-0.5B & Qwen-2.5-0.5B & Both: Deterministic & 40.87 & \textcolor{darkgreen}{4.38 $\uparrow$} & 0 & 0.00 & 0 & 0 & 0 \\
Qwen-2.5-0.5B & Qwen-2.5-0.5B & Both: Exploratory & 33.50 & \textcolor{darkred}{2.99 $\downarrow$} & 1.90 & 1.17 & 279 & 338 & 257 \\
Qwen-2.5-0.5B & Qwen-2.5-0.5B & Both: Det. \& Exp. & 41.93 & \textcolor{darkgreen}{5.44 $\uparrow$} & 1.64 & 1.08 & 251 & 355 & 289 \\
\midrule
Qwen-2.5-1.5B & Qwen-2.5-1.5B & Both: Default & 67.40 & \textcolor{darkgreen}{0.88 $\uparrow$} & 0.44 & 0.34 & 110 & 154 & 154 \\
Qwen-2.5-1.5B & Qwen-2.5-1.5B & Both: Deterministic & 68.14 & \textcolor{darkgreen}{1.62 $\uparrow$} & 0 & 0.00 & 0 & 2 & 1 \\
Qwen-2.5-1.5B & Qwen-2.5-1.5B & Both: Exploratory & 67.24 & \textcolor{darkgreen}{0.72 $\uparrow$} & 0.60 & 0.51 & 143 & 217 & 201 \\
Qwen-2.5-1.5B & Qwen-2.5-1.5B & Both: Det. \& Exp. & 66.67 & \textcolor{darkgreen}{0.15 $\uparrow$} & 0.47 & 0.41 & 111 & 166 & 158 \\
\midrule
Qwen-2.5-3B & Qwen-2.5-3B & Both: Default & 74.37 & \textcolor{darkgreen}{1.71 $\uparrow$} & 0.37 & 0.33 & 85 & 128 & 123 \\
Qwen-2.5-3B & Qwen-2.5-3B & Both: Deterministic & 74.77 & \textcolor{darkgreen}{2.11 $\uparrow$} & 0 & 0.00 & 0 & 0 & 0 \\
Qwen-2.5-3B & Qwen-2.5-3B & Both: Exploratory & 73.87 & \textcolor{darkgreen}{1.21 $\uparrow$} & 0.39 & 0.37 & 93 & 127 & 120 \\
Qwen-2.5-3B & Qwen-2.5-3B & Both: Det. \& Exp. & 75.51 & \textcolor{darkgreen}{2.85 $\uparrow$} & 0.35 & 0.25 & 73 & 127 & 123 \\
\midrule
Qwen-2.5-7B & Qwen-2.5-7B & Both: Default & 81.57 & \textcolor{darkgreen}{2.01 $\uparrow$} & 0.15 & 0.14 & 38 & 66 & 64 \\
Qwen-2.5-7B & Qwen-2.5-7B & Both: Deterministic & 81.65 & \textcolor{darkgreen}{2.09 $\uparrow$} & 0 & 0.00 & 0 & 0 & 0 \\
Qwen-2.5-7B & Qwen-2.5-7B & Both: Exploratory & 81.90 & \textcolor{darkgreen}{2.34 $\uparrow$} & 0.19 & 0.19 & 46 & 78 & 75 \\
Qwen-2.5-7B & Qwen-2.5-7B & Both: Det. \& Exp. & 82.56 & \textcolor{darkgreen}{3.00 $\uparrow$} & 0.20 & 0.19 & 54 & 62 & 61 \\
\midrule
Qwen-2.5-14B & Qwen-2.5-14B & Both: Default & 83.37 & \textcolor{darkgreen}{1.00 $\uparrow$} & 0.15 & 0.15 & 34 & 43 & 41 \\
Qwen-2.5-14B & Qwen-2.5-14B & Both: Deterministic & 83.70 & \textcolor{darkgreen}{1.33 $\uparrow$} & 0 & 0.00 & 0 & 0 & 0 \\
Qwen-2.5-14B & Qwen-2.5-14B & Both: Exploratory & 83.21 & \textcolor{darkgreen}{0.84 $\uparrow$} & 0.18 & 0.19 & 44 & 66 & 62 \\
Qwen-2.5-14B & Qwen-2.5-14B & Both: Det. \& Exp. & 83.87 & \textcolor{darkgreen}{1.50 $\uparrow$} & 0.16 & 0.15 & 40 & 59 & 54 \\
\midrule
Qwen-2.5-32B & Qwen-2.5-32B & Both: Default & 86.24 & \textcolor{darkgreen}{0.48 $\uparrow$} & 0.12 & 0.17 & 28 & 47 & 46 \\
Qwen-2.5-32B & Qwen-2.5-32B & Both: Deterministic & 85.75 & \textcolor{darkred}{0.01 $\downarrow$} & 0 & 0.00 & 0 & 0 & 0 \\
Qwen-2.5-32B & Qwen-2.5-32B & Both: Exploratory & 86.24 & \textcolor{darkgreen}{0.48 $\uparrow$} & 0.14 & 0.20 & 34 & 46 & 43 \\
Qwen-2.5-32B & Qwen-2.5-32B & Both: Det. \& Exp. & 86.57 & \textcolor{darkgreen}{0.81 $\uparrow$} & 0.16 & 0.24 & 32 & 55 & 46 \\
\midrule
Phi-mini-3.8B & Phi-mini-3.8B & Both: Default & 71.66 & \textcolor{darkgreen}{1.78 $\uparrow$} & 0.46 & 0.68 & 108 & 100 & 79 \\
Phi-mini-3.8B & Phi-mini-3.8B & Both: Deterministic & 72.24 & \textcolor{darkgreen}{2.36 $\uparrow$} & 0 & 0.00 & 0 & 0 & 0 \\
Phi-mini-3.8B & Phi-mini-3.8B & Both: Exploratory & 73.87 & \textcolor{darkgreen}{3.99 $\uparrow$} & 0.50 & 0.70 & 85 & 141 & 121 \\
Phi-mini-3.8B & Phi-mini-3.8B & Both: Det. \& Exp. & 73.22 & \textcolor{darkgreen}{3.34 $\uparrow$} & 0.47 & 0.66 & 91 & 124 & 105 \\
\midrule
Llama-3.1-3B & Llama-3.1-3B & Both: Default & 68.55 & \textcolor{darkgreen}{3.51 $\uparrow$} & 0.44 & 0.40 & 107 & 117 & 110 \\
Llama-3.1-3B & Llama-3.1-3B & Both: Deterministic & 67.40 & \textcolor{darkgreen}{2.36 $\uparrow$} & 0 & 0.00 & 0 & 0 & 0 \\
Llama-3.1-3B & Llama-3.1-3B & Both: Exploratory & 66.75 & \textcolor{darkgreen}{1.71 $\uparrow$} & 0.53 & 0.48 & 116 & 131 & 122 \\
Llama-3.1-3B & Llama-3.1-3B & Both: Det. \& Exp. & 67.73 & \textcolor{darkgreen}{2.69 $\uparrow$} & 0.47 & 0.45 & 105 & 113 & 109 \\
\midrule
Mistral-7B & Mistral-7B & Both: Default & 66.34 & \textcolor{darkgreen}{1.79 $\uparrow$} & 0.30 & 0.22 & 57 & 64 & 57 \\
Mistral-7B & Mistral-7B & Both: Deterministic & 66.99 & \textcolor{darkgreen}{2.44 $\uparrow$} & 0 & 0.00 & 0 & 0 & 0 \\
Mistral-7B & Mistral-7B & Both: Exploratory & 65.11 & \textcolor{darkgreen}{0.56 $\uparrow$} & 0.38 & 0.30 & 81 & 85 & 80 \\
Mistral-7B & Mistral-7B & Both: Det. \& Exp. & 66.42 & \textcolor{darkgreen}{1.87 $\uparrow$} & 0.34 & 0.25 & 62 & 89 & 81 \\
\midrule
Llama-3.1-8B & Llama-3.1-8B & Both: Default & 74.28 & \textcolor{darkgreen}{1.26 $\uparrow$} & 0.41 & 0.47 & 79 & 114 & 106 \\
Llama-3.1-8B & Llama-3.1-8B & Both: Deterministic & 75.43 & \textcolor{darkgreen}{2.41 $\uparrow$} & 0 & 0.00 & 0 & 2 & 1 \\
Llama-3.1-8B & Llama-3.1-8B & Both: Exploratory & 74.86 & \textcolor{darkgreen}{1.84 $\uparrow$} & 0.46 & 0.54 & 95 & 139 & 130 \\
Llama-3.1-8B & Llama-3.1-8B & Both: Det. \& Exp. & 74.45 & \textcolor{darkgreen}{1.43 $\uparrow$} & 0.41 & 0.48 & 99 & 112 & 102 \\
\bottomrule
\end{tabular}
\end{adjustbox}
\caption{Comparative Analysis of Language Model Performance in Multi-Agent Debate Settings on the \textbf{CommonsenseQA} Dataset. This table showcases the impact of different \textbf{Agent Settings} (controlling temperature and top\_p parameters like Default, Deterministic, Exploratory, and a combination) on the \textbf{Accuracy} of various language models. The \textbf{$\Delta$} column quantifies the \textbf{improvement (or decline) over the single base model performance}. Further metrics include average \textbf{Debate Rounds}, normalized \textbf{Sycophancy} (per 1221 data points), and transitions between correct (C) and incorrect (I) states (C$\rightarrow$I, I$\rightarrow$C), highlighting the nuanced effects of debate dynamics.}
\label{tab:debate_performance_commonsenseqa1}
\end{table*}

\begin{table*}[htbp]
\centering
\begin{adjustbox}{width=\textwidth,center} 
\sisetup{round-mode=places,round-precision=2} 
\begin{tabular}{@{}ccccS[table-format=2.2]S[table-format=2.2]S[table-format=3.2]S[table-format=1.2]S[table-format=4.2]cccS[table-format=1.2]@{}} 
\toprule
\textbf{Agent 1} & \textbf{Agent 2} & \textbf{Agent Settings} & \textbf{Accuracy} & \multicolumn{1}{c}{\textbf{$\Delta_1$}} & \multicolumn{1}{c}{\textbf{$\Delta_2$}} & \textbf{Debate} & \multicolumn{1}{c}{\textbf{Sycophancy}} & \textbf{C$\rightarrow$I} & \textbf{I$\rightarrow$C} & \textbf{Debate} \\
 & & & & & & \textbf{Rounds} & \multicolumn{1}{c}{\textbf{(Avg / 1221)}} & & & \textbf{Helped} \\
 & & & & & & \textbf{(Avg)} & & & & \textbf{(Overall)} \\
\midrule
Qwen-2.5-0.5B & Qwen-2.5-1.5B & Both: Default & 56.92 & \textcolor{darkgreen}{20.43 $\uparrow$} & \textcolor{darkred}{9.60 $\downarrow$} & 1.34 & 0.84 & 237 & 370 & 345 \\
Qwen-2.5-0.5B & Qwen-2.5-1.5B & Both: Deterministic & 58.39 & \textcolor{darkgreen}{21.90 $\uparrow$} & \textcolor{darkred}{8.13 $\downarrow$} & 1.26 & 0.63 & 148 & 326 & 295 \\
Qwen-2.5-0.5B & Qwen-2.5-1.5B & Both: Exploratory & 56.91 & \textcolor{darkgreen}{20.42 $\uparrow$} & \textcolor{darkred}{9.61 $\downarrow$} & 1.63 & 0.99 & 216 & 430 & 377 \\
Qwen-2.5-0.5B & Qwen-2.5-1.5B & Both: Det. \& Exp. & 57.08 & \textcolor{darkgreen}{20.59 $\uparrow$} & \textcolor{darkred}{9.44 $\downarrow$} & 1.28 & 0.82 & 177 & 371 & 332 \\
Qwen-2.5-0.5B & Qwen-2.5-1.5B & Both: Exp. \& Det. & 57.49 & \textcolor{darkgreen}{21.00 $\uparrow$} & \textcolor{darkred}{9.03 $\downarrow$} & 1.51 & 0.87 & 206 & 407 & 379 \\
\midrule
Qwen-2.5-1.5B & Llama-3.1-3B & Both: Default & 66.83 & \textcolor{darkgreen}{0.31 $\uparrow$} & \textcolor{darkgreen}{1.79 $\uparrow$} & 0.59 & 0.63 & 168 & 170 & 165 \\
Qwen-2.5-1.5B & Llama-3.1-3B & Both: Deterministic & 68.63 & \textcolor{darkgreen}{2.11 $\uparrow$} & \textcolor{darkgreen}{3.59 $\uparrow$} & 0.66 & 0.80 & 160 & 198 & 184 \\
Qwen-2.5-1.5B & Llama-3.1-3B & Both: Exploratory & 67.08 & \textcolor{darkgreen}{0.56 $\uparrow$} & \textcolor{darkgreen}{2.04 $\uparrow$} & 0.82 & 0.90 & 164 & 237 & 223 \\
Qwen-2.5-1.5B & Llama-3.1-3B & Both: Det. \& Exp. & 69.78 & \textcolor{darkgreen}{3.26 $\uparrow$} & \textcolor{darkgreen}{4.74 $\uparrow$} & 0.61 & 0.69 & 140 & 203 & 193 \\
Qwen-2.5-1.5B & Llama-3.1-3B & Both: Exp. \& Det. & 67.73 & \textcolor{darkgreen}{1.21 $\uparrow$} & \textcolor{darkgreen}{2.69 $\uparrow$} & 0.66 & 0.72 & 160 & 219 & 200 \\
\midrule
Qwen-2.5-3B & Phi-mini-3.8B & Both: Default & 75.02 & \textcolor{darkgreen}{2.36 $\uparrow$} & \textcolor{darkgreen}{5.14 $\uparrow$} & 0.44 & 0.39 & 100 & 158 & 150 \\
Qwen-2.5-3B & Phi-mini-3.8B & Both: Deterministic & 76.09 & \textcolor{darkgreen}{3.43 $\uparrow$} & \textcolor{darkgreen}{6.21 $\uparrow$} & 0.50 & 0.37 & 104 & 161 & 154 \\
Qwen-2.5-3B & Phi-mini-3.8B & Both: Exploratory & 74.69 & \textcolor{darkgreen}{2.03 $\uparrow$} & \textcolor{darkgreen}{4.81 $\uparrow$} & 0.50 & 0.52 & 85 & 177 & 167 \\
Qwen-2.5-3B & Phi-mini-3.8B & Both: Det. \& Exp. & 75.76 & \textcolor{darkgreen}{3.10 $\uparrow$} & \textcolor{darkgreen}{5.88 $\uparrow$} & 0.52 & 0.40 & 114 & 191 & 179 \\
Qwen-2.5-3B & Phi-mini-3.8B & Both: Exp. \& Det. & 75.10 & \textcolor{darkgreen}{2.44 $\uparrow$} & \textcolor{darkgreen}{5.22 $\uparrow$} & 0.49 & 0.49 & 106 & 162 & 156 \\
\midrule
Qwen-2.5-1.5B & Qwen-2.5-3B & Both: Default & 73.87 & \textcolor{darkgreen}{7.35 $\uparrow$} & \textcolor{darkgreen}{1.21 $\uparrow$} & 0.51 & 0.47 & 100 & 225 & 217 \\
Qwen-2.5-1.5B & Qwen-2.5-3B & Both: Deterministic & 74.94 & \textcolor{darkgreen}{8.42 $\uparrow$} & \textcolor{darkgreen}{2.28 $\uparrow$} & 0.48 & 0.40 & 108 & 191 & 187 \\
Qwen-2.5-1.5B & Qwen-2.5-3B & Both: Exploratory & 74.12 & \textcolor{darkgreen}{7.60 $\uparrow$} & \textcolor{darkgreen}{1.46 $\uparrow$} & 0.60 & 0.55 & 115 & 279 & 264 \\
Qwen-2.5-1.5B & Qwen-2.5-3B & Both: Det. \& Exp. & 74.04 & \textcolor{darkgreen}{7.52 $\uparrow$} & \textcolor{darkgreen}{1.38 $\uparrow$} & 0.51 & 0.52 & 106 & 208 & 204 \\
Qwen-2.5-1.5B & Qwen-2.5-3B & Both: Exp. \& Det. & 74.94 & \textcolor{darkgreen}{8.42 $\uparrow$} & \textcolor{darkgreen}{2.28 $\uparrow$} & 0.57 & 0.42 & 108 & 251 & 246 \\
\midrule
Llama-3.1-3B & Llama-3.1-8B & Both: Default & 72.24 & \textcolor{darkgreen}{7.20 $\uparrow$} & \textcolor{darkred}{0.78 $\downarrow$} & 0.54 & 0.52 & 119 & 165 & 153 \\
Llama-3.1-3B & Llama-3.1-8B & Both: Deterministic & 73.79 & \textcolor{darkgreen}{8.75 $\uparrow$} & \textcolor{darkgreen}{0.77 $\uparrow$} & 0.57 & 0.57 & 118 & 190 & 183 \\
Llama-3.1-3B & Llama-3.1-8B & Both: Exploratory & 72.15 & \textcolor{darkgreen}{7.11 $\uparrow$} & \textcolor{darkred}{0.87 $\downarrow$} & 0.59 & 0.58 & 112 & 167 & 157 \\
Llama-3.1-3B & Llama-3.1-8B & Both: Det. \& Exp. & 70.68 & \textcolor{darkgreen}{5.64 $\uparrow$} & \textcolor{darkred}{2.34 $\downarrow$} & 0.60 & 0.58 & 131 & 162 & 154 \\
Llama-3.1-3B & Llama-3.1-8B & Both: Exp. \& Det. & 73.96 & \textcolor{darkgreen}{8.92 $\uparrow$} & \textcolor{darkgreen}{0.94 $\uparrow$} & 0.60 & 0.61 & 120 & 200 & 193 \\
\midrule
Qwen-2.5-7B & Qwen-2.5-14B & Both: Default & 83.37 & \textcolor{darkgreen}{3.81 $\uparrow$} & \textcolor{darkgreen}{1.00 $\uparrow$} & 0.28 & 0.26 & 62 & 98 & 96 \\
Qwen-2.5-7B & Qwen-2.5-14B & Both: Deterministic & 83.78 & \textcolor{darkgreen}{4.22 $\uparrow$} & \textcolor{darkgreen}{1.41 $\uparrow$} & 0.33 & 0.21 & 71 & 101 & 95 \\
Qwen-2.5-7B & Qwen-2.5-14B & Both: Exploratory & 84.19 & \textcolor{darkgreen}{4.63 $\uparrow$} & \textcolor{darkgreen}{1.82 $\uparrow$} & 0.28 & 0.27 & 60 & 112 & 110 \\
Qwen-2.5-7B & Qwen-2.5-14B & Both: Det. \& Exp. & 83.37 & \textcolor{darkgreen}{3.81 $\uparrow$} & \textcolor{darkgreen}{1.00 $\uparrow$} & 0.29 & 0.24 & 66 & 103 & 99 \\
Qwen-2.5-7B & Qwen-2.5-14B & Both: Exp. \& Det. & 83.29 & \textcolor{darkgreen}{3.73 $\uparrow$} & \textcolor{darkgreen}{0.92 $\uparrow$} & 0.28 & 0.21 & 66 & 95 & 93 \\
\bottomrule
\end{tabular}
\end{adjustbox}
\caption{Comparative Analysis of Mixed Language Model Performance in Multi-Agent Debate Settings on the \textbf{CommonsenseQA} Dataset. This table showcases the impact of different \textbf{Agent Settings} (controlling temperature and top\_p parameters) on the \textbf{Accuracy} when pairing different language models. The \textbf{$\Delta_1$} column shows the improvement over the weaker model's performance, while \textbf{$\Delta_2$} shows comparison to the stronger model. This highlights whether mixed-agent debates benefit from model complementarity or are constrained by the weaker model's capabilities. Further metrics include average \textbf{Debate Rounds}, normalized \textbf{Sycophancy} (per 1221 data points), and transitions between correct (C) and incorrect (I) states.}
\label{tab:mixed_debate_performance_commonsenseqa2}
\end{table*}

\begin{table*}[htbp]
\centering
\begin{adjustbox}{width=\textwidth,center} 
\sisetup{round-mode=places,round-precision=2} 
\begin{tabular}{@{}ccccS[table-format=2.2]S[table-format=3.2]S[table-format=1.2]S[table-format=4.2]cccS[table-format=1.2]@{}} 
\toprule
\textbf{Agent 1} & \textbf{Agent 2} & \textbf{Agent 3} & \textbf{Agent Settings} & \multicolumn{1}{c}{\textbf{Accuracy}} & \multicolumn{1}{c}{\textbf{$\Delta$}} & \textbf{Debate} & \multicolumn{1}{c}{\textbf{Sycophancy}} & \textbf{C$\rightarrow$I} & \textbf{I$\rightarrow$C} & \textbf{Debate} \\
 & & & & & & \textbf{Rounds} & \multicolumn{1}{c}{\textbf{(Avg / 1221)}} & & & \textbf{Helped} \\
 & & & & & & \textbf{(Avg)} & & & & \textbf{(Overall)} \\
\midrule
Qwen-2.5-0.5B & Qwen-2.5-0.5B & Qwen-2.5-0.5B & Default & 37.76 & \textcolor{darkgreen}{1.27 $\uparrow$} & 2.69 & 3.02 & 545 & 538 & 327 \\
Qwen-2.5-0.5B & Qwen-2.5-0.5B & Qwen-2.5-0.5B & Deterministic & 39.80 & \textcolor{darkgreen}{3.31 $\uparrow$} & 0 & 0.00 & 0 & 0 & 0 \\
Qwen-2.5-0.5B & Qwen-2.5-0.5B & Qwen-2.5-0.5B & Exploratory & 32.60 & \textcolor{darkred}{3.89 $\downarrow$} & 3.45 & 3.66 & 580 & 604 & 336 \\
Qwen-2.5-0.5B & Qwen-2.5-0.5B & Qwen-2.5-0.5B & 1 Det. \& 2 Exp. & 36.77 & \textcolor{darkgreen}{0.28 $\uparrow$} & 3.05 & 3.11 & 569 & 558 & 317 \\
Qwen-2.5-0.5B & Qwen-2.5-0.5B & Qwen-2.5-0.5B & 2 Det. \& 1 Exp. & 37.51 & \textcolor{darkgreen}{1.02 $\uparrow$} & 1.76 & 1.84 & 433 & 420 & 237 \\
\midrule
Qwen-2.5-1.5B & Qwen-2.5-1.5B & Qwen-2.5-1.5B & Default & 68.80 & \textcolor{darkgreen}{2.28 $\uparrow$} & 0.77 & 0.83 & 193 & 333 & 264 \\
Qwen-2.5-1.5B & Qwen-2.5-1.5B & Qwen-2.5-1.5B & Deterministic & 67.90 & \textcolor{darkgreen}{1.38 $\uparrow$} & 0 & 0.00 & 0 & 3 & 1 \\
Qwen-2.5-1.5B & Qwen-2.5-1.5B & Qwen-2.5-1.5B & Exploratory & 67.57 & \textcolor{darkgreen}{1.05 $\uparrow$} & 1.14 & 1.34 & 256 & 429 & 315 \\
Qwen-2.5-1.5B & Qwen-2.5-1.5B & Qwen-2.5-1.5B & 1 Det. \& 2 Exp. & 68.55 & \textcolor{darkgreen}{2.03 $\uparrow$} & 0.92 & 1.01 & 211 & 346 & 270 \\
Qwen-2.5-1.5B & Qwen-2.5-1.5B & Qwen-2.5-1.5B & 2 Det. \& 1 Exp. & 68.55 & \textcolor{darkgreen}{2.03 $\uparrow$} & 0.57 & 0.57 & 172 & 244 & 179 \\
\midrule
Qwen-2.5-3B & Qwen-2.5-3B & Qwen-2.5-3B & Default & 75.18 & \textcolor{darkgreen}{2.52 $\uparrow$} & 0.63 & 0.68 & 147 & 225 & 180 \\
Qwen-2.5-3B & Qwen-2.5-3B & Qwen-2.5-3B & Deterministic & 74.28 & \textcolor{darkgreen}{1.62 $\uparrow$} & 0 & 0.00 & 0 & 0 & 0 \\
Qwen-2.5-3B & Qwen-2.5-3B & Qwen-2.5-3B & Exploratory & 74.37 & \textcolor{darkgreen}{1.71 $\uparrow$} & 0.66 & 0.82 & 164 & 248 & 196 \\
Qwen-2.5-3B & Qwen-2.5-3B & Qwen-2.5-3B & 1 Det. \& 2 Exp. & 75.02 & \textcolor{darkgreen}{2.36 $\uparrow$} & 0.67 & 0.66 & 166 & 211 & 163 \\
Qwen-2.5-3B & Qwen-2.5-3B & Qwen-2.5-3B & 2 Det. \& 1 Exp. & 75.76 & \textcolor{darkgreen}{3.10 $\uparrow$} & 0.45 & 0.44 & 116 & 163 & 115 \\
\midrule
Qwen-2.5-7B & Qwen-2.5-7B & Qwen-2.5-7B & Default & 81.90 & \textcolor{darkgreen}{2.34 $\uparrow$} & 0.31 & 0.38 & 85 & 122 & 96 \\
Qwen-2.5-7B & Qwen-2.5-7B & Qwen-2.5-7B & Deterministic & 81.57 & \textcolor{darkgreen}{2.01 $\uparrow$} & 0 & 0.00 & 0 & 0 & 0 \\
Qwen-2.5-7B & Qwen-2.5-7B & Qwen-2.5-7B & Exploratory & 81.98 & \textcolor{darkgreen}{2.42 $\uparrow$} & 0.38 & 0.47 & 99 & 147 & 117 \\
Qwen-2.5-7B & Qwen-2.5-7B & Qwen-2.5-7B & 1 Det. \& 2 Exp. & 81.41 & \textcolor{darkgreen}{1.85 $\uparrow$} & 0.32 & 0.38 & 98 & 124 & 99 \\
Qwen-2.5-7B & Qwen-2.5-7B & Qwen-2.5-7B & 2 Det. \& 1 Exp. & 81.74 & \textcolor{darkgreen}{2.18 $\uparrow$} & 0.25 & 0.26 & 84 & 89 & 65 \\
\midrule
Qwen-2.5-14B & Qwen-2.5-14B & Qwen-2.5-14B & Default & 83.05 & \textcolor{darkgreen}{0.68 $\uparrow$} & 0.27 & 0.28 & 84 & 85 & 69 \\
Qwen-2.5-14B & Qwen-2.5-14B & Qwen-2.5-14B & Deterministic & 83.87 & \textcolor{darkgreen}{1.50 $\uparrow$} & 0 & 0.00 & 0 & 0 & 0 \\
Qwen-2.5-14B & Qwen-2.5-14B & Qwen-2.5-14B & Exploratory & 83.13 & \textcolor{darkgreen}{0.76 $\uparrow$} & 0.28 & 0.33 & 76 & 100 & 75 \\
Qwen-2.5-14B & Qwen-2.5-14B & Qwen-2.5-14B & 1 Det. \& 2 Exp. & 83.54 & \textcolor{darkgreen}{1.17 $\uparrow$} & 0.25 & 0.25 & 74 & 93 & 77 \\
Qwen-2.5-14B & Qwen-2.5-14B & Qwen-2.5-14B & 2 Det. \& 1 Exp. & 83.95 & \textcolor{darkgreen}{1.58 $\uparrow$} & 0.14 & 0.12 & 45 & 56 & 46 \\
\midrule
Qwen-2.5-32B & Qwen-2.5-32B & Qwen-2.5-32B & Default & 86.00 & \textcolor{darkgreen}{0.24 $\uparrow$} & 0.18 & 0.26 & 61 & 80 & 67 \\
Qwen-2.5-32B & Qwen-2.5-32B & Qwen-2.5-32B & Deterministic & 85.75 & \textcolor{darkred}{0.01 $\downarrow$} & 0 & 0.00 & 0 & 0 & 0 \\
Qwen-2.5-32B & Qwen-2.5-32B & Qwen-2.5-32B & Exploratory & 86.57 & \textcolor{darkgreen}{0.81 $\uparrow$} & 0.18 & 0.25 & 56 & 87 & 74 \\
Qwen-2.5-32B & Qwen-2.5-32B & Qwen-2.5-32B & 1 Det. \& 2 Exp. & 86.00 & \textcolor{darkgreen}{0.24 $\uparrow$} & 0.16 & 0.21 & 61 & 71 & 57 \\
Qwen-2.5-32B & Qwen-2.5-32B & Qwen-2.5-32B & 2 Det. \& 1 Exp. & 86.08 & \textcolor{darkgreen}{0.32 $\uparrow$} & 0.11 & 0.14 & 35 & 50 & 41 \\
\midrule
Phi-mini-3.8B & Phi-mini-3.8B & Phi-mini-3.8B & Default & 73.22 & \textcolor{darkgreen}{3.34 $\uparrow$} & 0.62 & 1.12 & 170 & 171 & 121 \\
Phi-mini-3.8B & Phi-mini-3.8B & Phi-mini-3.8B & Deterministic & 73.71 & \textcolor{darkgreen}{3.83 $\uparrow$} & 0 & 0.00 & 0 & 0 & 0 \\
Phi-mini-3.8B & Phi-mini-3.8B & Phi-mini-3.8B & Exploratory & 73.96 & \textcolor{darkgreen}{4.08 $\uparrow$} & 0.74 & 1.24 & 161 & 231 & 170 \\
Phi-mini-3.8B & Phi-mini-3.8B & Phi-mini-3.8B & 1 Det. \& 2 Exp. & 75.18 & \textcolor{darkgreen}{5.30 $\uparrow$} & 0.69 & 1.21 & 134 & 217 & 159 \\
Phi-mini-3.8B & Phi-mini-3.8B & Phi-mini-3.8B & 2 Det. \& 1 Exp. & 73.71 & \textcolor{darkgreen}{3.83 $\uparrow$} & 0.47 & 0.86 & 107 & 137 & 97 \\
\midrule
Llama-3.1-3B & Llama-3.1-3B & Llama-3.1-3B & Default & 68.39 & \textcolor{darkgreen}{3.35 $\uparrow$} & 0.87 & 0.92 & 210 & 237 & 169 \\
Llama-3.1-3B & Llama-3.1-3B & Llama-3.1-3B & Deterministic & 68.06 & \textcolor{darkgreen}{3.02 $\uparrow$} & 0 & 0.00 & 0 & 0 & 0 \\
Llama-3.1-3B & Llama-3.1-3B & Llama-3.1-3B & Exploratory & 67.65 & \textcolor{darkgreen}{2.61 $\uparrow$} & 1.04 & 1.16 & 250 & 261 & 190 \\
Llama-3.1-3B & Llama-3.1-3B & Llama-3.1-3B & 1 Det. \& 2 Exp. & 67.08 & \textcolor{darkgreen}{2.04 $\uparrow$} & 0.89 & 0.95 & 213 & 225 & 165 \\
Llama-3.1-3B & Llama-3.1-3B & Llama-3.1-3B & 2 Det. \& 1 Exp. & 67.73 & \textcolor{darkgreen}{2.69 $\uparrow$} & 0.58 & 0.58 & 132 & 149 & 105 \\
\midrule
Mistral-7B & Mistral-7B & Mistral-7B & Default & 66.83 & \textcolor{darkgreen}{2.28 $\uparrow$} & 0.53 & 0.57 & 121 & 137 & 99 \\
Mistral-7B & Mistral-7B & Mistral-7B & Deterministic & 66.75 & \textcolor{darkgreen}{2.20 $\uparrow$} & 0 & 0.00 & 0 & 0 & 0 \\
Mistral-7B & Mistral-7B & Mistral-7B & Exploratory & 65.60 & \textcolor{darkgreen}{1.05 $\uparrow$} & 0.79 & 0.83 & 179 & 167 & 119 \\
Mistral-7B & Mistral-7B & Mistral-7B & 1 Det. \& 2 Exp. & 65.44 & \textcolor{darkgreen}{0.89 $\uparrow$} & 0.64 & 0.70 & 157 & 144 & 97 \\
Mistral-7B & Mistral-7B & Mistral-7B & 2 Det. \& 1 Exp. & 66.75 & \textcolor{darkgreen}{2.20 $\uparrow$} & 0.32 & 0.35 & 81 & 98 & 68 \\
\midrule
Llama-3.1-8B & Llama-3.1-8B & Llama-3.1-8B & Default & 75.92 & \textcolor{darkgreen}{2.90 $\uparrow$} & 0.62 & 0.83 & 147 & 211 & 148 \\
Llama-3.1-8B & Llama-3.1-8B & Llama-3.1-8B & Deterministic & 75.84 & \textcolor{darkgreen}{2.82 $\uparrow$} & 0.00 & 0.00 & 0 & 9 & 3 \\
Llama-3.1-8B & Llama-3.1-8B & Llama-3.1-8B & Exploratory & 74.12 & \textcolor{darkgreen}{1.10 $\uparrow$} & 0.79 & 1.13 & 203 & 246 & 168 \\
Llama-3.1-8B & Llama-3.1-8B & Llama-3.1-8B & 1 Det. \& 2 Exp. & 75.51 & \textcolor{darkgreen}{2.49 $\uparrow$} & 0.71 & 0.94 & 173 & 233 & 161 \\
Llama-3.1-8B & Llama-3.1-8B & Llama-3.1-8B & 2 Det. \& 1 Exp. & 75.51 & \textcolor{darkgreen}{2.49 $\uparrow$} & 0.44 & 0.60 & 118 & 150 & 92 \\
\bottomrule
\end{tabular}
\end{adjustbox}
\caption{Comparative Analysis of Language Model Performance in Multi-Agent Debate Settings on the \textbf{CommonsenseQA} Dataset. This table showcases the impact of different \textbf{Agent Settings} (controlling temperature and top\_p parameters like Default, Deterministic, Exploratory, and combinations) on the \textbf{Accuracy} of various language models. The \textbf{$\Delta$} column quantifies the \textbf{improvement (or decline) over the single base model performance}. Further metrics include average \textbf{Debate Rounds}, normalized \textbf{Sycophancy} (per 1221 data points), and transitions between correct (C) and incorrect (I) states (C$\rightarrow$I, I$\rightarrow$C), highlighting the nuanced effects of debate dynamics.}
\label{tab:debate_performance_commonsenseqa3}
\end{table*}

\begin{table*}[htbp]
\centering
\begin{adjustbox}{width=\textwidth,center} 
\sisetup{round-mode=places,round-precision=2} 
\begin{tabular}{@{}cccS[table-format=2.2]S[table-format=3.2]S[table-format=1.2]S[table-format=4.2]cccS[table-format=1.2]@{}} 
\toprule
\textbf{Agent 1} & \textbf{Agent 2} & \textbf{Agent 3} & \multicolumn{1}{c}{\textbf{Accuracy}} & \multicolumn{1}{c}{\textbf{$\Delta$}} & \textbf{Debate} & \multicolumn{1}{c}{\textbf{Sycophancy}} & \textbf{C$\rightarrow$I} & \textbf{I$\rightarrow$C} & \textbf{Debate} \\
 & & & & & \textbf{Rounds} & \multicolumn{1}{c}{\textbf{(Avg / 1221)}} & & & \textbf{Helped} \\
 & & & & & \textbf{(Avg)} & & & & \textbf{(Overall)} \\
\midrule
Qwen-2.5-0.5B & Qwen-2.5-1.5B & Qwen-2.5-3B & 72.48 & \textcolor{darkgreen}{35.99 $\uparrow$} & 1.64 & 1.51 & 228 & 748 & 563 \\
Qwen-2.5-0.5B & Qwen-2.5-1.5B & Llama-3.1-3B & 65.03 & \textcolor{darkgreen}{28.54 $\uparrow$} & 1.81 & 1.89 & 343 & 622 & 480 \\
Qwen-2.5-0.5B & Qwen-2.5-1.5B & Phi-mini-3.8B & 70.60 & \textcolor{darkgreen}{34.11 $\uparrow$} & 1.68 & 1.73 & 246 & 691 & 537 \\
Qwen-2.5-0.5B & Qwen-2.5-3B & Llama-3.1-3B & 72.56 & \textcolor{darkgreen}{36.07 $\uparrow$} & 1.81 & 1.59 & 234 & 697 & 544 \\
Qwen-2.5-0.5B & Qwen-2.5-3B & Phi-mini-3.8B & 72.15 & \textcolor{darkgreen}{35.66 $\uparrow$} & 1.66 & 1.59 & 243 & 629 & 517 \\
Qwen-2.5-0.5B & Llama-3.1-3B & Phi-mini-3.8B & 69.12 & \textcolor{darkgreen}{32.63 $\uparrow$} & 1.76 & 1.91 & 298 & 617 & 483 \\
Qwen-2.5-1.5B & Qwen-2.5-3B & Llama-3.1-3B & 73.38 & \textcolor{darkgreen}{6.86 $\uparrow$} & 1.08 & 1.22 & 230 & 399 & 305 \\
Qwen-2.5-1.5B & Qwen-2.5-3B & Phi-mini-3.8B & 75.68 & \textcolor{darkgreen}{9.16 $\uparrow$} & 0.95 & 1.17 & 202 & 382 & 303 \\
Qwen-2.5-1.5B & Llama-3.1-3B & Phi-mini-3.8B & 71.09 & \textcolor{darkgreen}{4.57 $\uparrow$} & 1.04 & 1.42 & 260 & 347 & 273 \\
Qwen-2.5-3B & Phi-mini-3.8B & Llama-3.1-3B & 74.20 & \textcolor{darkgreen}{1.54 $\uparrow$} & 1.00 & 1.15 & 222 & 334 & 253 \\
\midrule
Qwen-2.5-3B & Qwen-2.5-3B & Phi-mini-3.8B & 74.77 & \textcolor{darkgreen}{2.11 $\uparrow$} & 0.73 & 0.84 & 200 & 256 & 193 \\
Qwen-2.5-3B & Phi-mini-3.8B & Phi-mini-3.8B & 76.09 & \textcolor{darkgreen}{3.43 $\uparrow$} & 0.85 & 1.18 & 183 & 258 & 186 \\
Qwen-2.5-0.5B & Qwen-2.5-1.5B & Qwen-2.5-1.5B & 64.86 & \textcolor{darkgreen}{28.37 $\uparrow$} & 1.86 & 1.50 & 267 & 576 & 447 \\
Qwen-2.5-0.5B & Qwen-2.5-0.5B & Qwen-2.5-1.5B & 55.12 & \textcolor{darkgreen}{18.63 $\uparrow$} & 2.41 & 2.44 & 384 & 651 & 438 \\
\bottomrule
\end{tabular}
\end{adjustbox}
\caption{Comparative Analysis of Mixed Language Model Performance in Multi-Agent Debate Settings on the \textbf{CommonsenseQA} Dataset. This table presents results for heterogeneous combinations of language models in debate settings. The \textbf{$\Delta$} column quantifies the improvement over the performance of the weakest model in each combination (for combinations with Qwen-2.5-0.5B, the baseline is 36.49\%; for others, the baseline corresponds to the lowest-performing model). All experiments use the default debate setting. The table shows that combining models of different capacities can lead to significant performance gains, especially when smaller models are paired with larger ones.}
\label{tab:mixed_debate_performance_commonsenseqa4}
\end{table*}

\clearpage

\section{Additional Results}

\subsection{Original MAD Results}

We also report our experiments with the original Multi-Agent Debate (MAD) framework across various model sizes and architectures. Table~\ref{tab:original_mad} presents the results on three challenging reasoning benchmarks: GSM-Plus, GSM8K, and ARC-Challenge.

\subsection{Majority Vote@3 Results}

To further investigate the impact of stochastic diversity on model performance, we report results on a Majority Vote@3 approach where we sample three independent responses from each model and take a majority vote to determine the final answer. Table~\ref{tab:majority_vote} presents these results across five benchmarks: GSM8K, GSM-Plus, ARC-Easy, ARC-Challenge, and CommonsenseQA.

The results demonstrate that simple ensemble-based approaches can significantly boost performance without requiring multi-agent debate or model fine-tuning. Across all model sizes and architectures, Majority Vote@3 consistently outperforms single-sample inference. The relative improvements are most pronounced for smaller models, with Qwen-2.5-0.5B gaining up to 4.27 percentage points on ARC-Challenge and Qwen-2.5-1.5B showing similar substantial improvements across benchmarks.

Interestingly, this pattern holds across model families. Llama-3.1-3B, Phi-3.5-mini, and Mistral-7B all exhibit significant gains when using majority voting, suggesting that the benefits of ensemble diversity transcend specific model architectures. The results also indicate diminishing returns for larger models—Qwen-2.5-14B shows more modest improvements compared to its smaller counterparts, likely because these larger models already produce more consistent answers across samples.

These findings highlight an important baseline for our research: simple ensemble methods provide strong performance improvements with minimal computational overhead during inference. However, they still require multiple forward passes for each query, motivating our DTE approach that aims to distill these benefits into a single model through training on debate traces.

\subsection{Scaling Results for Multiple Agents}

We investigated how performance scales with increasing numbers of debating agents (1-7) across different model sizes and reasoning benchmarks. Table~\ref{tab:scaling_results} presents these results, revealing several important trends in multi-agent scaling behavior.

First, we observe that performance generally improves as we add more agents to the debate, but with diminishing returns. The most significant gains occur when moving from a single agent (equivalent to standard inference) to two agents, with more modest improvements as additional agents join the debate. For example, on GSM8K, Qwen-2.5-1.5B shows a substantial jump from 62.77\% (1 agent) to 71.57\% (2 agents), but only incremental improvements thereafter.

Second, the benefits of additional agents vary across tasks. On more complex tasks like GSM-Plus, we see continued performance improvements even with 7 agents, particularly for larger models. Qwen-2.5-14B shows its peak GSM-Plus performance with 7 agents (78.08\%), suggesting that more difficult problems benefit from extended multi-agent collaboration. In contrast, on simpler tasks like ARC-Easy, performance plateaus more quickly.

Third, we find that model size influences scaling behavior. Smaller models like Qwen-2.5-1.5B show more variability in performance as agents are added, with occasional performance drops when moving from 3 to 4 agents. Larger models exhibit more stable scaling patterns, suggesting that they can more consistently integrate insights from multiple debate participants.

These results have important implications for our DTE framework. They demonstrate that while adding more agents generally improves performance, the computational costs may outweigh the benefits beyond 3-5 agents for most applications. This insight helped inform our design choices in balancing performance gains against computational efficiency in our final framework.

\begin{table*}[t]
\centering

\resizebox{\textwidth}{!}{%
\begin{tabular}{lllccccccc}
\toprule
\multicolumn{3}{c}{\textbf{Model Configuration}} & \multicolumn{7}{c}{\textbf{Debate Performance Metrics}} \\
\cmidrule(lr){1-3} \cmidrule(lr){4-10}
\textbf{Agent 1} & \textbf{Agent 2} & \textbf{Debate Setting} & \textbf{Accuracy} & \textbf{Delta} & \textbf{Debate Rounds} & \textbf{Sycophancy} & \textbf{Correct→Incorrect} & \textbf{Incorrect→Correct} & \textbf{Net Benefit} \\
\midrule
\multicolumn{10}{c}{\textit{GSM-Plus}} \\
\midrule
Qwen-2.5-0.5B & Qwen-2.5-0.5B & exploratory & 28.12\% & \textcolor{darkgreen}{3.33 $\uparrow$} & 3.48 & 6906 & 261 & 575 & 432 \\
Qwen-2.5-1.5B & Qwen-2.5-1.5B & exploratory & 46.50\% & \textcolor{darkgreen}{4.50 $\uparrow$} & 2.33 & 5642 & 194 & 861 & 670 \\
Qwen-2.5-3B & Qwen-2.5-3B & exploratory & 66.79\% & \textcolor{darkgreen}{5.04 $\uparrow$} & 1.34 & 5315 & 231 & 373 & 187 \\
Qwen-2.5-7B & Qwen-2.5-7B & exploratory & 69.71\% & \textcolor{darkgreen}{1.09 $\uparrow$} & 0.76 & 2967 & 102 & 200 & 121 \\
Qwen-2.5-14B & Qwen-2.5-14B & exploratory & 76.92\% & \textcolor{darkgreen}{5.13 $\uparrow$} & 0.61 & 2722 & 119 & 151 & 47 \\
Phi-mini-3.8B & Phi-mini-3.8B & exploratory & 65.79\% & \textcolor{darkgreen}{2.37 $\uparrow$} & 1.07 & 3620 & 180 & 272 & 136 \\
Llama-3.1-3B & Llama-3.1-3B & exploratory & 42.42\% & \textcolor{darkred}{3.25 $\downarrow$} & 2.07 & 5507 & 379 & 369 & 238 \\
Mistral-7B & Mistral-7B & exploratory & 26.35\% & \textcolor{darkgreen}{11.31 $\uparrow$} & 1.85 & 4500 & 210 & 290 & 115 \\
Llama-3.1-8B & Llama-3.1-8B & exploratory & 57.63\% & \textcolor{darkgreen}{2.01 $\uparrow$} & 1.75 & 5667 & 273 & 585 & 351 \\
\midrule
\multicolumn{10}{c}{\textit{GSM8K}} \\
\midrule
Qwen-2.5-0.5B & Qwen-2.5-0.5B & exploratory & 45.56\% & \textcolor{darkgreen}{3.56 $\uparrow$} & 2.85 & 3469 & 175 & 427 & 328 \\
Qwen-2.5-1.5B & Qwen-2.5-1.5B & exploratory & 65.81\% & \textcolor{darkgreen}{3.04 $\uparrow$} & 1.99 & 3471 & 144 & 650 & 489 \\
Qwen-2.5-3B & Qwen-2.5-3B & exploratory & 86.96\% & \textcolor{darkgreen}{1.82 $\uparrow$} & 0.63 & 1390 & 82 & 165 & 97 \\
Qwen-2.5-7B & Qwen-2.5-7B & exploratory & 91.74\% & \textcolor{darkgreen}{1.07 $\uparrow$} & 0.38 & 930 & 64 & 93 & 33 \\
Qwen-2.5-14B & Qwen-2.5-14B & exploratory & 94.39\% & \textcolor{darkgreen}{1.59 $\uparrow$} & 0.18 & 448 & 30 & 48 & 18 \\
Phi-mini-3.8B & Phi-mini-3.8B & exploratory & 88.17\% & \textcolor{darkgreen}{1.29 $\uparrow$} & 0.45 & 1050 & 65 & 120 & 65 \\
Llama-3.1-3B & Llama-3.1-3B & exploratory & 67.63\% & \textcolor{darkred}{4.92 $\downarrow$} & 1.51 & 2418 & 238 & 215 & 127 \\
Mistral-7B & Mistral-7B & exploratory & 43.44\% & \textcolor{darkgreen}{22.06 $\uparrow$} & 1.65 & 2100 & 175 & 235 & 95 \\
Llama-3.1-8B & Llama-3.1-8B & exploratory & 83.02\% & \textcolor{darkgreen}{1.29 $\uparrow$} & 0.94 & 1587 & 93 & 308 & 236 \\
\midrule
\multicolumn{10}{c}{\textit{ARC-Challenge}} \\
\midrule
Qwen-2.5-0.5B & Qwen-2.5-0.5B & exploratory & 38.65\% & \textcolor{darkgreen}{0.68 $\uparrow$} & 1.88 & 2728 & 272 & 308 & 232 \\
Qwen-2.5-1.5B & Qwen-2.5-1.5B & exploratory & 74.15\% & \textcolor{darkgreen}{0.94 $\uparrow$} & 0.85 & 1671 & 121 & 231 & 156 \\
Qwen-2.5-3B & Qwen-2.5-3B & exploratory & 85.41\% & \textcolor{darkgreen}{1.88 $\uparrow$} & 0.57 & 1227 & 94 & 135 & 57 \\
Qwen-2.5-7B & Qwen-2.5-7B & exploratory & 91.47\% & \textcolor{darkgreen}{6.25 $\uparrow$} & 0.23 & 501 & 41 & 49 & 13 \\
Qwen-2.5-14B & Qwen-2.5-14B & exploratory & 94.54\% & \textcolor{darkgreen}{4.27 $\uparrow$} & 0.15 & 326 & 31 & 37 & 9 \\
Phi-mini-3.8B & Phi-mini-3.8B & exploratory & 87.46\% & \textcolor{darkgreen}{2.73 $\uparrow$} & 0.15 & 313 & 24 & 47 & 25 \\
Llama-3.1-3B & Llama-3.1-3B & exploratory & 76.37\% & \textcolor{darkgreen}{3.25 $\uparrow$} & 0.73 & 1525 & 111 & 155 & 69 \\
Mistral-7B & Mistral-7B & exploratory & 73.29\% & \textcolor{darkgreen}{4.52 $\uparrow$} & 0.40 & 795 & 63 & 114 & 73 \\
Llama-3.1-8B & Llama-3.1-8B & exploratory & 86.09\% & \textcolor{darkgreen}{8.44 $\uparrow$} & 0.27 & 514 & 31 & 84 & 58 \\
\bottomrule
\end{tabular}%
}
\caption{Performance of the original Multi-Agent Debate (MAD) framework across different model sizes and reasoning benchmarks. Results show accuracy, improvement over single-agent baseline (Delta), average debate rounds, and debate transition statistics. The Delta column highlights performance changes compared to individual model accuracy, with green indicating improvement and red indicating decline.}
\label{tab:original_mad}
\end{table*}

\begin{table*}[t]
\centering
\begin{tabular}{lccccc}
\toprule
\multirow{2}{*}{\textbf{Model}} & \multicolumn{5}{c}{\textbf{Accuracy (\%) on Benchmarks}} \\
\cmidrule(lr){2-6}
& \textbf{GSM8K} & \textbf{GSM-Plus} & \textbf{ARC-E} & \textbf{ARC-C} & \textbf{CQA} \\
\midrule
Qwen-2.5-0.5B & 49.73 & 30.54 & 58.71 & 42.92 & 42.51 \\
Qwen-2.5-1.5B & 75.82 & 52.08 & 87.12 & 73.55 & 69.62 \\
Qwen-2.5-3B & 86.28 & 64.08 & 94.19 & 84.13 & 76.90 \\
Qwen-2.5-7B & 92.19 & 70.46 & 96.46 & 91.21 & 82.88 \\
Qwen-2.5-14B & 94.09 & 72.54 & 98.44 & 94.20 & 82.15 \\
\midrule
Llama-3.1-3B & 77.03 & 52.79 & 88.51 & 75.00 & 69.94 \\
Llama-3.1-8B & 85.82 & 60.88 & 93.56 & 83.11 & 74.86 \\
Phi-3.5-mini & 87.87 & 65.79 & 96.00 & 86.95 & 75.10 \\
Mistral-7B & 56.86 & 36.88 & 87.58 & 75.68 & 69.04 \\
\bottomrule
\end{tabular}
\caption{Performance comparison using Majority Vote@3 approach across different benchmarks. For each model, we sample three independent responses and determine the final answer through majority voting.}
\label{tab:majority_vote}
\end{table*}

\begin{table*}[t]
\centering

\caption{Performance scaling with increasing numbers of debating agents (1-7) across different model sizes and reasoning benchmarks. Results show accuracy percentages for each configuration.}
\label{tab:scaling_results}
\begin{tabular}{lccccccc}
\toprule
\multirow{2}{*}{\textbf{Model}} & \multicolumn{7}{c}{\textbf{Number of Agents}} \\
\cmidrule(lr){2-8}
& \textbf{1} & \textbf{2} & \textbf{3} & \textbf{4} & \textbf{5} & \textbf{6} & \textbf{7} \\
\midrule
\multicolumn{8}{c}{\textit{GSM8K Accuracy (\%)}} \\
\midrule
Qwen-2.5-1.5B & 62.77 & 71.57 & 75.13 & 75.89 & 75.13 & 74.98 & 76.50 \\
Qwen-2.5-3B & 85.14 & 85.52 & 87.64 & 87.11 & 87.04 & 86.66 & 87.11 \\
Qwen-2.5-7B & 90.67 & 91.21 & 92.42 & 92.49 & 92.57 & 92.34 & 92.72 \\
Qwen-2.5-14B & 92.80 & 93.33 & 94.84 & 94.31 & 94.69 & 94.62 & 94.24 \\
\midrule
\multicolumn{8}{c}{\textit{GSM-Plus Accuracy (\%)}} \\
\midrule
Qwen-2.5-1.5B & 42.00 & 51.62 & 53.33 & 50.62 & 54.21 & 51.50 & 52.67 \\
Qwen-2.5-3B & 61.75 & 67.79 & 68.00 & 64.21 & 69.71 & 64.88 & 68.54 \\
Qwen-2.5-7B & 68.62 & 74.17 & 74.96 & 70.88 & 71.08 & 71.38 & 76.00 \\
Qwen-2.5-14B & 71.79 & 77.25 & 72.29 & 72.83 & 73.29 & 73.38 & 78.08 \\
\midrule
\multicolumn{8}{c}{\textit{ARC-Challenge Accuracy (\%)}} \\
\midrule
Qwen-2.5-1.5B & 69.21 & 68.52 & 72.10 & 71.50 & 72.53 & 71.50 & 72.10 \\
Qwen-2.5-3B & 82.53 & 84.64 & 86.26 & 85.75 & 86.26 & 86.95 & 87.03 \\
Qwen-2.5-7B & 87.22 & 91.64 & 91.72 & 91.47 & 92.06 & 91.38 & 92.32 \\
Qwen-2.5-14B & 90.27 & 93.77 & 94.80 & 95.14 & 94.20 & 94.62 & 94.28 \\
\midrule
\multicolumn{8}{c}{\textit{ARC-Easy Accuracy (\%)}} \\
\midrule
Qwen-2.5-1.5B & 86.62 & 83.42 & 85.61 & 86.32 & 87.46 & 86.57 & 87.16 \\
Qwen-2.5-3B & 93.06 & 94.15 & 94.28 & 94.32 & 94.82 & 94.91 & 94.99 \\
Qwen-2.5-7B & 94.69 & 96.93 & 96.55 & 96.34 & 96.42 & 96.25 & 96.59 \\
Qwen-2.5-14B & 95.66 & 98.15 & 98.19 & 98.23 & 98.15 & 98.19 & 98.23 \\
\bottomrule
\end{tabular}
\end{table*}

\end{document}